\newcommand{\SparvAuthors}{%
  \author{%
    Jianheng~Liu, Yunfei Wan, Fangming Cheng, Xiangyu Hong, Zuhao Zou, Haotian Li, \\Longji Yin, Chunran Zheng, Jiarong Lin, and~Fu~Zhang%
    \thanks{The authors are with the Department of Mechanical Engineering, The University of Hong Kong. Corresponding email: jianheng@connect.hku.hk, fuzhang@hku.hk}%
  }%
}
\newcommand{\SparvCodeStatement}{Our code will be released at \url{https://github.com/hku-mars/SpaRV}.}

\documentclass[journal]{IEEEtran}

\IEEEoverridecommandlockouts

\usepackage{cite}
\usepackage{amsmath}
\usepackage{float}
\usepackage{subfigure}
\usepackage{overpic}
\usepackage{wrapfig}
\usepackage{balance}
\usepackage{ulem}
\usepackage{color}
\usepackage[colorlinks,linkcolor=red]{hyperref}
\usepackage{mathtools}
\usepackage{booktabs}
\usepackage{newtxmath}
\usepackage{capt-of}
\usepackage[linesnumbered, ruled]{algorithm2e}

\graphicspath{{figures/}}
\usepackage{multirow}
\usepackage{graphicx}
\usepackage{dsfont}
\usepackage{physics}
\usepackage{tikz}
\usetikzlibrary{arrows.meta,positioning,calc,shapes.geometric,decorations.markings,decorations.pathreplacing,fit,backgrounds,patterns}
\usepackage{makecell}

\definecolor{gold}{rgb}{1.0, 0.874, 0}
\definecolor{silver}{rgb}{0.75, 0.75, 0.75}
\definecolor{brown}{rgb}{0.8, 0.678, 0.4}
\definecolor{gray}{rgb}{0.5, 0.5, 0.5}

\makeatletter
\DeclareRobustCommand\onedot{\futurelet\@let@token\@onedot}
\def\@onedot{\ifx\@let@token.\else.\null\fi\xspace}

\usepackage{enumitem}
\setlist[itemize]{parsep=0pt,partopsep=0pt,leftmargin=*,itemsep=5pt}
\setlist[enumerate]{parsep=0pt,partopsep=0pt,leftmargin=*,itemsep=5pt}
\makeatletter
\renewcommand\paragraph{\@startsection{paragraph}{4}{\z@}%
    {.5em}%
    {-1em}%
    {\normalfont\normalsize\bfseries}}
\makeatother
\expandafter\def\expandafter\normalsize\expandafter{%
    \normalsize%
    \setlength\abovedisplayskip{1ex}%
    \setlength\belowdisplayskip{1ex}%
    \setlength\abovedisplayshortskip{1ex}%
    \setlength\belowdisplayshortskip{1ex}%
}

\newcommand{\Real}{\mathbb{R}}

\newcommand{\maxlv}{L}

\newcommand{\nsamp}{K}

\newcommand{\ray}{\mathbf{r}}

\newcommand{\rayq}{\mathbf{q}}
\newcommand{\pt}{\mathbf{p}}

\newcommand{\vox}{\mathbf{v}}

\newcommand{\voxcen}{\vox_{\mathrm{c}}}
\newcommand{\voxsiz}{v_{\mathrm{s}}}
\newcommand{\voxgeo}{\vox_{\mathrm{geo}}}
\newcommand{\voxidx}{\vox_{\mathrm{idx}}}

\newcommand{\voxpriority}{v_{\mathrm{priority}}}

\newcommand{\contract}{\operatorname{Contr}}
\newcommand{\pixrgb}{\mathbf{C}}
\newcommand{\pixnormal}{\mathbf{N}}
\newcommand{\pixnormalmean}{\pixnormal_{\mathrm{mean}}}
\newcommand{\pixnormalmed}{\pixnormal_{\mathrm{med}}}
\newcommand{\pixnormalsigma}{\pixnormal_{\sigma}}
\newcommand{\loss}{\mathcal{L}}

\newcommand{\explin}{\mathrm{explin}}

\newcommand{\interp}{\mathrm{interp}}

\newcommand{\world}{\mathbf{w}}
\newcommand{\worldcen}{\world_{\mathrm{c}}}
\newcommand{\worldsiz}{w_{\mathrm{s}}}

\newcommand{\lamdepth}{\lambda_{\mathrm{d}}}
\newcommand{\lampocc}{\lambda_{\mathrm{occ}}}
\newcommand{\lamnormal}{\lambda_{\mathrm{n}}}
\newcommand{\lamsmooth}{\lambda_{\mathrm{s}}}
\newcommand{\lamssim}{\lambda_{\mathrm{ssim}}}

\definecolor{zoomedge}{rgb}{0.745,0.071,0.235}  %
\definecolor{oursedge}{rgb}{0.114,0.306,0.847}  %
\definecolor{zoomhair}{rgb}{0.820,0.835,0.859}  %

\newcommand{\zcropx}{1.024}
\newcommand{\zcropy}{0.055}
\newcommand{\zcropw}{1.691}
\newcommand{\zcroph}{0.440}
\newcommand{\setzoomcrop}[4]{%
  \renewcommand{\zcropx}{#1}\renewcommand{\zcropy}{#2}%
  \renewcommand{\zcropw}{#3}\renewcommand{\zcroph}{#4}}

\newlength{\zoomcellwd}
\newsavebox{\zoomcellbx}
\newcommand{\zoomcellgap}{2}
\newsavebox{\qualblk}%
\newcommand{\zoomcell}[3][zoomhair]{%
  \def\zoomcellframe{#1}%
  \setlength{\zoomcellwd}{#2}%
  \sbox{\zoomcellbx}{\includegraphics[height=100pt]{#3}}%
  \edef\zoomcellnat{\the\wd\zoomcellbx}%
  \begin{tikzpicture}[x=1pt,y=1pt]
    \pgfmathsetmacro{\zR}{\zoomcellnat/100}%
    \pgfmathsetmacro{\zW}{\zoomcellwd}%
    \pgfmathsetmacro{\zMH}{\zW/\zR}%
    \pgfmathsetmacro{\zfx}{\zcropx/\zR}%
    \pgfmathsetmacro{\zfw}{\zcropw/\zR}%
    \pgfmathsetmacro{\zS}{\zW/\zfw}%
    \pgfmathsetmacro{\zZH}{\zS*\zcroph/\zR}%
    \pgfmathsetmacro{\zYb}{\zZH+\zoomcellgap}%
    \begin{scope}
      \clip (0,0) rectangle (\zW,\zZH);
      \node[anchor=south west,inner sep=0]
        at ({-\zfx*\zS},{-(1-\zcropy-\zcroph)*\zS/\zR})
        {\includegraphics[width=\zS pt]{#3}};
    \end{scope}
    \node[anchor=south west,inner sep=0] at (0,\zYb)
      {\includegraphics[width=\zW pt]{#3}};
    \draw[zoomedge,line width=0.6pt,dash pattern=on 2pt off 1.3pt]
      ({\zfx*\zW},{\zYb+(1-\zcropy-\zcroph)*\zMH})
      rectangle ({(\zfx+\zfw)*\zW},{\zYb+(1-\zcropy)*\zMH});
    \ifx\zoomcellframe\empty\else
      \draw[#1,line width=0.8pt] (0,\zYb) rectangle (\zW,{\zYb+\zMH});
    \fi
    \draw[zoomedge,line width=0.8pt] (0,0) rectangle (\zW,\zZH);
    \ifx\zoomcelllabel\empty\else
      \node[anchor=north west,inner sep=1.2pt,fill=white,fill opacity=0.8,
        text opacity=1,font=\zoomtagfont] at (1,{\zYb+\zMH-1}) {\zoomcelllabel};
    \fi
  \end{tikzpicture}%
}
\def\zoomcelllabel{}
\newcommand{\zoomtagfont}{\scriptsize}
\newcommand{\zoomcelltag}[3]{%
  \def\zoomcelllabel{#1}\zoomcell[]{#2}{#3}\def\zoomcelllabel{}}

\newcommand{\zcropax}{0.191}
\newcommand{\zcropay}{0.341}
\newcommand{\zcropaw}{0.485}
\newcommand{\zcropah}{0.301}
\newcommand{\zcropbx}{1.379}
\newcommand{\zcropby}{0.206}
\newcommand{\zcropbw}{0.379}
\newcommand{\zcropbh}{0.235}
\newcommand{\setzoomcropa}[4]{%
  \renewcommand{\zcropax}{#1}\renewcommand{\zcropay}{#2}%
  \renewcommand{\zcropaw}{#3}\renewcommand{\zcropah}{#4}}
\newcommand{\setzoomcropb}[4]{%
  \renewcommand{\zcropbx}{#1}\renewcommand{\zcropby}{#2}%
  \renewcommand{\zcropbw}{#3}\renewcommand{\zcropbh}{#4}}

\newcommand{\zoomcelltwo}[3][zoomhair]{%
  \def\zoomcellframe{#1}%
  \setlength{\zoomcellwd}{#2}%
  \sbox{\zoomcellbx}{\includegraphics[height=100pt]{#3}}%
  \edef\zoomcellnat{\the\wd\zoomcellbx}%
  \begin{tikzpicture}[x=1pt,y=1pt]
    \pgfmathsetmacro{\zR}{\zoomcellnat/100}%
    \pgfmathsetmacro{\zW}{\zoomcellwd}%
    \pgfmathsetmacro{\zG}{2}%
    \pgfmathsetmacro{\zMH}{\zW/\zR}%
    \pgfmathsetmacro{\zWt}{(\zW-\zG)/2}%
    \pgfmathsetmacro{\zIa}{\zcropah/\zcropaw}%
    \pgfmathsetmacro{\zIb}{\zcropbh/\zcropbw}%
    \pgfmathsetmacro{\zZHa}{\zWt*\zIa}%
    \pgfmathsetmacro{\zZHb}{\zWt*\zIb}%
    \pgfmathsetmacro{\zZH}{max(\zZHa,\zZHb)}%
    \pgfmathsetmacro{\zYb}{\zZH+\zG}%
    \pgfmathsetmacro{\zfxa}{\zcropax/\zR}%
    \pgfmathsetmacro{\zfwa}{\zcropaw/\zR}%
    \pgfmathsetmacro{\zSa}{\zWt/\zfwa}%
    \pgfmathsetmacro{\zfxb}{\zcropbx/\zR}%
    \pgfmathsetmacro{\zfwb}{\zcropbw/\zR}%
    \pgfmathsetmacro{\zSb}{\zWt/\zfwb}%
    \pgfmathsetmacro{\zXb}{\zWt+\zG}%
    \begin{scope}
      \clip (0,0) rectangle (\zWt,\zZHa);
      \node[anchor=south west,inner sep=0]
        at ({-\zfxa*\zSa},{-(1-\zcropay-\zcropah)*\zSa/\zR})
        {\includegraphics[width=\zSa pt]{#3}};
    \end{scope}
    \begin{scope}
      \clip (\zXb,0) rectangle (\zW,\zZHb);
      \node[anchor=south west,inner sep=0]
        at ({\zXb-\zfxb*\zSb},{-(1-\zcropby-\zcropbh)*\zSb/\zR})
        {\includegraphics[width=\zSb pt]{#3}};
    \end{scope}
    \node[anchor=south west,inner sep=0] at (0,\zYb)
      {\includegraphics[width=\zW pt]{#3}};
    \draw[zoomedge,line width=0.6pt,dash pattern=on 2pt off 1.3pt]
      ({\zfxa*\zW},{\zYb+(1-\zcropay-\zcropah)*\zMH})
      rectangle ({(\zfxa+\zfwa)*\zW},{\zYb+(1-\zcropay)*\zMH});
    \draw[zoomedge,line width=0.6pt,dash pattern=on 2pt off 1.3pt]
      ({\zfxb*\zW},{\zYb+(1-\zcropby-\zcropbh)*\zMH})
      rectangle ({(\zfxb+\zfwb)*\zW},{\zYb+(1-\zcropby)*\zMH});
    \ifx\zoomcellframe\empty\else
      \draw[#1,line width=0.8pt] (0,\zYb) rectangle (\zW,{\zYb+\zMH});
    \fi
    \draw[zoomedge,line width=0.8pt] (0,0) rectangle (\zWt,\zZHa);
    \draw[zoomedge,line width=0.8pt] (\zXb,0) rectangle (\zW,\zZHb);
  \end{tikzpicture}%
}

\newcommand{\cropboxcell}[3][zoomhair]{%
  \def\zoomcellframe{#1}%
  \setlength{\zoomcellwd}{#2}%
  \sbox{\zoomcellbx}{\includegraphics[height=100pt]{#3}}%
  \edef\zoomcellnat{\the\wd\zoomcellbx}%
  \begin{tikzpicture}[x=1pt,y=1pt]
    \pgfmathsetmacro{\zR}{\zoomcellnat/100}%
    \pgfmathsetmacro{\zW}{\zoomcellwd}%
    \pgfmathsetmacro{\zMH}{\zW/\zR}%
    \pgfmathsetmacro{\zfxa}{\zcropax/\zR}%
    \pgfmathsetmacro{\zfwa}{\zcropaw/\zR}%
    \pgfmathsetmacro{\zfxb}{\zcropbx/\zR}%
    \pgfmathsetmacro{\zfwb}{\zcropbw/\zR}%
    \node[anchor=south west,inner sep=0] at (0,0)
      {\includegraphics[width=\zW pt]{#3}};
    \draw[zoomedge,line width=0.6pt,dash pattern=on 2pt off 1.3pt]
      ({\zfxa*\zW},{(1-\zcropay-\zcropah)*\zMH})
      rectangle ({(\zfxa+\zfwa)*\zW},{(1-\zcropay)*\zMH});
    \draw[zoomedge,line width=0.6pt,dash pattern=on 2pt off 1.3pt]
      ({\zfxb*\zW},{(1-\zcropby-\zcropbh)*\zMH})
      rectangle ({(\zfxb+\zfwb)*\zW},{(1-\zcropby)*\zMH});
    \ifx\zoomcellframe\empty\else
      \draw[#1,line width=0.8pt] (0,0) rectangle (\zW,\zMH);
    \fi
    \ifx\zoomcelllabel\empty\else
      \node[anchor=north west,inner sep=1.2pt,fill=white,fill opacity=0.8,
        text opacity=1,font=\zoomtagfont] at (1,{\zMH-1}) {\zoomcelllabel};
    \fi
  \end{tikzpicture}%
}
\newcommand{\cropboxcelltag}[3]{%
  \def\zoomcelllabel{#1}\cropboxcell[]{#2}{#3}\def\zoomcelllabel{}}

\newcommand{\imagetag}[3]{%
  \begin{tikzpicture}[x=1pt,y=1pt]
    \node[anchor=north west,inner sep=0] at (0,0)
      {\includegraphics[width=#2]{#3}};
    \node[anchor=north west,inner sep=1.2pt,fill=white,fill opacity=0.8,
      text opacity=1,font=\zoomtagfont] at (1,-1) {#1};
  \end{tikzpicture}%
}

\providecommand{\SparvAuthors}{}
\providecommand{\SparvCodeStatement}{We will open-source our implementation.}

\title{\LARGE \bf
  SpaRV: Sparse Radiance Voxels for Scalable Geometrically Consistent LiDAR-Visual Mapping and Rendering
}

\IEEEaftertitletext{%
  \begin{center}
    \includegraphics[width=\textwidth]{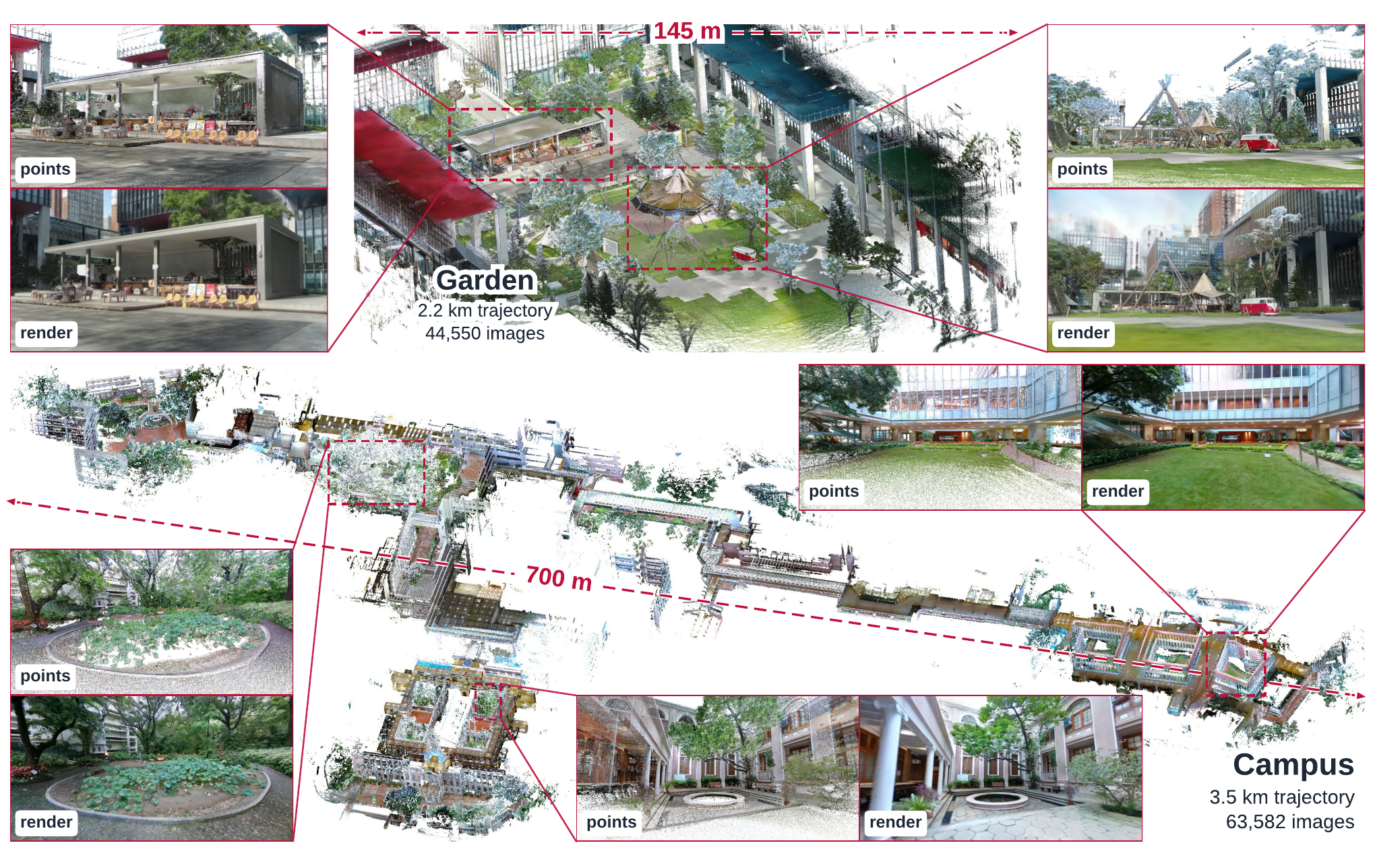}
    \vspace{-25pt}
    \captionof{figure}{The two self-collected scenes, \textit{Garden} (top) and \textit{Campus} (bottom), reconstructed by SpaRV on a single GPU. Each scene is drawn as the accumulated LiDAR point cloud of the capture, viewed from above. The red dashed boxes mark close-up regions; beside each box, \textit{points} is the point cloud of that region seen from a free viewpoint and \textit{render} is the SpaRV rendering from the same viewpoint. Garden has 44,550 images and 146.7\,M LiDAR points along a 2,239\,m trajectory, with a 45,199\,m$^2$ footprint; Campus has 63,582 images and 151.6\,M LiDAR points along a 3,534\,m trajectory, with a 234,035\,m$^2$ footprint.}
    \label{fig:manifold_overview}
  \end{center}%
}

\SparvAuthors

\begin{document}

\maketitle
\thispagestyle{empty}
\pagestyle{empty}

\begin{abstract}
Robotic simulation, navigation, and inspection need photorealistic and geometrically accurate maps.
Radiance fields achieve photorealism, but building them from robot data raises three challenges.
A robot's trajectory serves its task, not reconstruction, so images from few viewpoints leave the geometry under-constrained.
Residual errors in camera poses and intrinsics blur the map.
The map's GPU memory grows with scene size, which limits its level of detail in large scenes.
We present \textbf{SpaRV}, a \textbf{Spa}rse \textbf{R}adiance \textbf{V}oxel map rendered by differentiable ray marching, which lets irregular LiDAR scans supervise it alongside images.
LiDAR depth and occupancy anchor the under-constrained geometry, and normal and smoothness regularizers keep it smooth between LiDAR points.
To correct camera errors, SpaRV alternates map training with photometric optimization of the camera poses and focal lengths.
To bound GPU memory in training and rendering, SpaRV holds one submap at a time at full resolution and the rest at coarser octree levels.
Experiments show that SpaRV keeps the geometry consistent away from the captured trajectory and sharpens textures by optimizing the camera parameters.
On one 24\,GB GPU, SpaRV reconstructs a university campus captured in 63{,}582 images and simulates other cameras and LiDARs.
\SparvCodeStatement
\end{abstract}

\section{Introduction}
Digital twins of real environments support robot simulation, navigation, and inspection~\cite{mazumder2023towards}, and simulators train robot policies before deployment~\cite{makoviychuk2021isaac}.
Such applications need one map that supports novel view synthesis, surface extraction, and sensor simulation.
It must therefore be both photorealistic and geometrically accurate.
Cameras and LiDAR supply both properties: images record appearance, and LiDAR measures metric range.
LiDAR-visual SLAM~\cite{zheng2024fast,zhou2025fast} already fuses them into accurate colored point clouds, but point clouds do not render photorealistic novel views.

\begin{figure}[t]
  \centering
  \setzoomcrop{0.870}{0.090}{2.020}{0.430}
  \renewcommand{\zoomcellgap}{0}%
  \zoomcelltag{InstantNGP}{0.494\columnwidth}{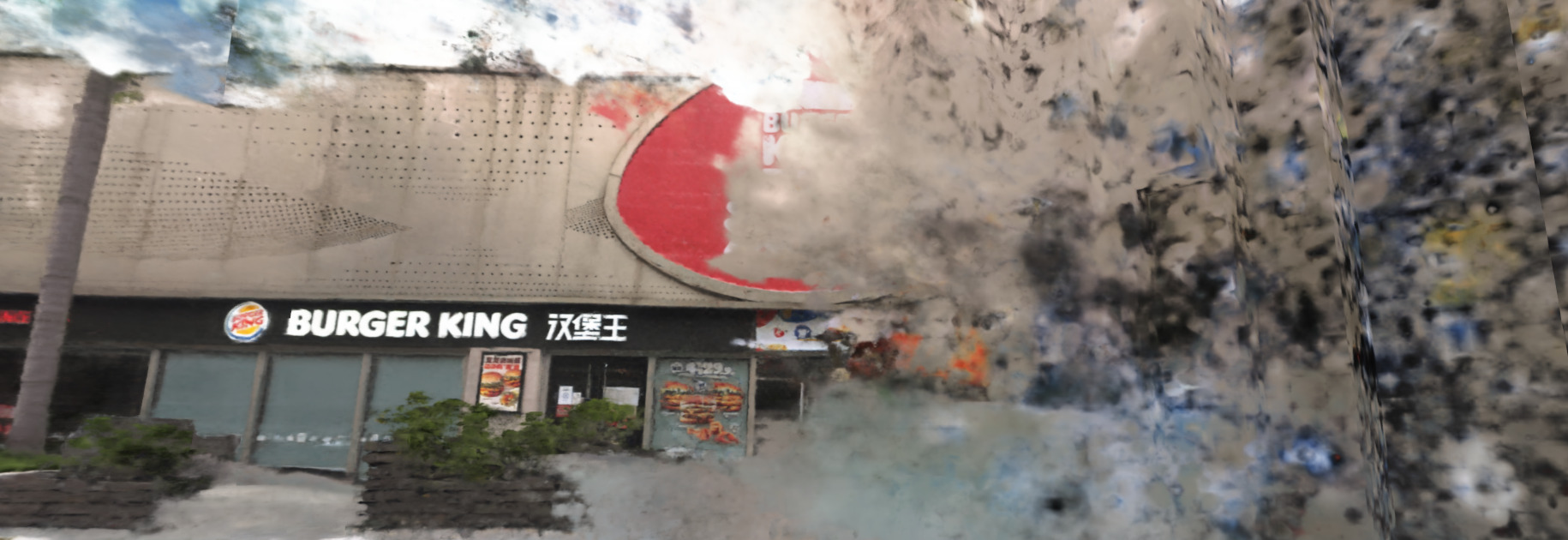}\hfill
  \zoomcelltag{3DGS}{0.494\columnwidth}{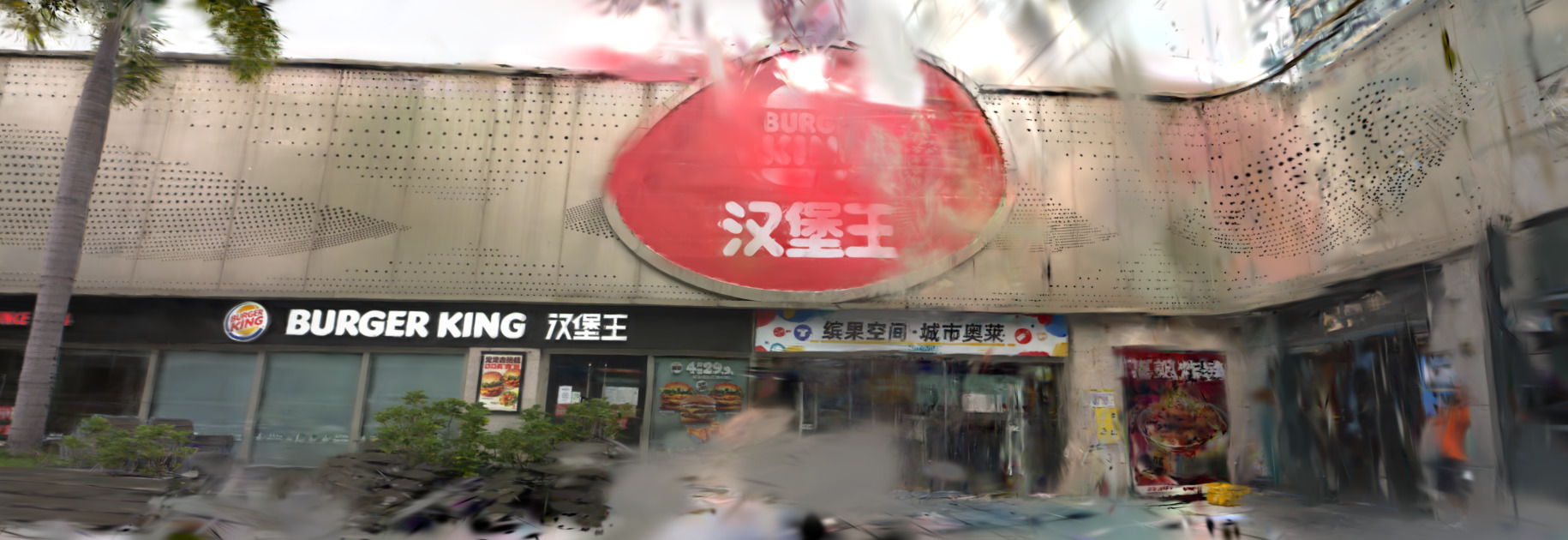}\par\vspace{3pt}
  \zoomcelltag{SVRaster}{0.494\columnwidth}{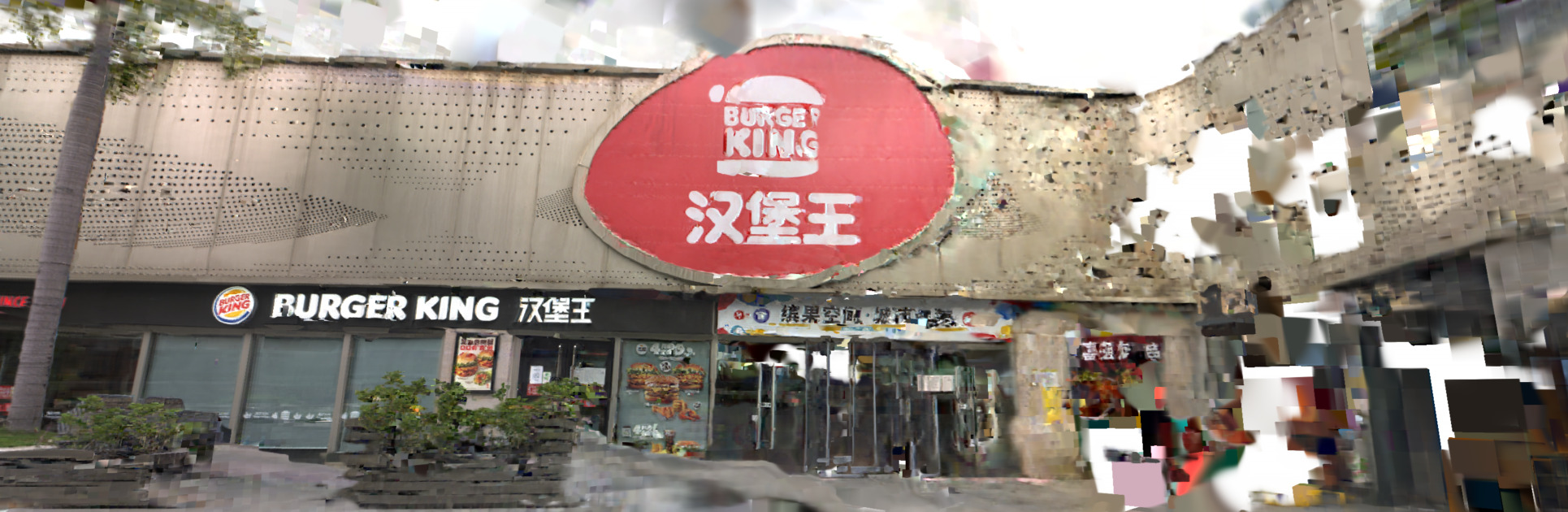}\hfill
  \zoomcelltag{LetsGo}{0.494\columnwidth}{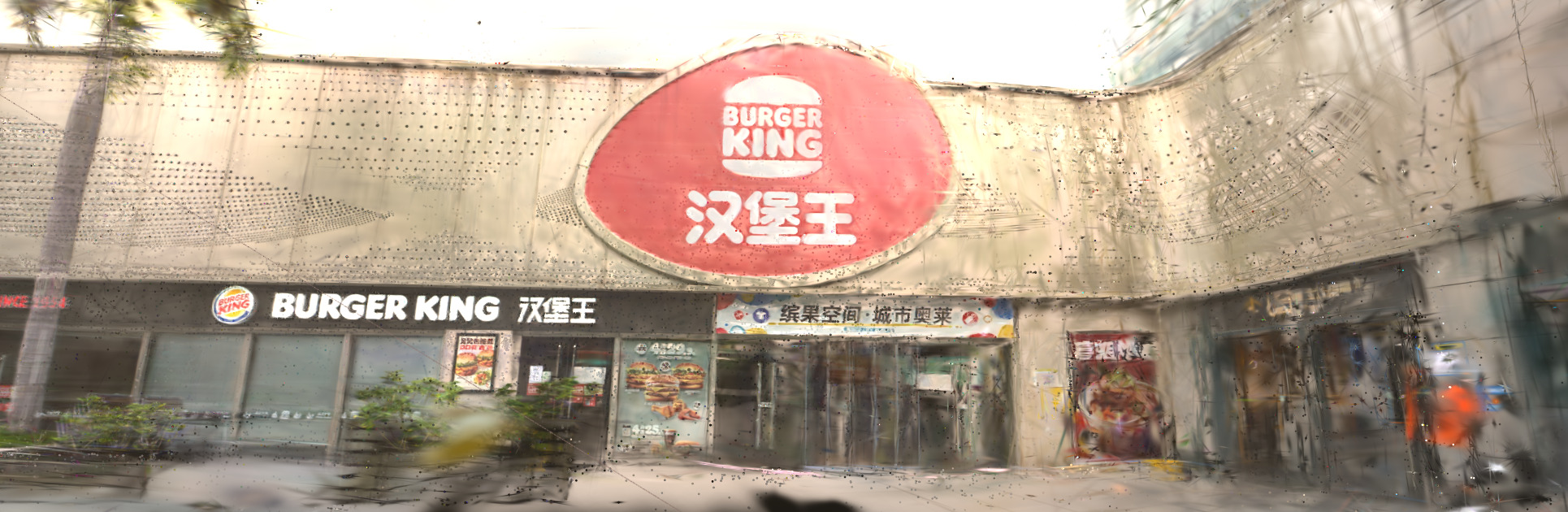}\par\vspace{3pt}
  \zoomcelltag{GS-SDF}{0.494\columnwidth}{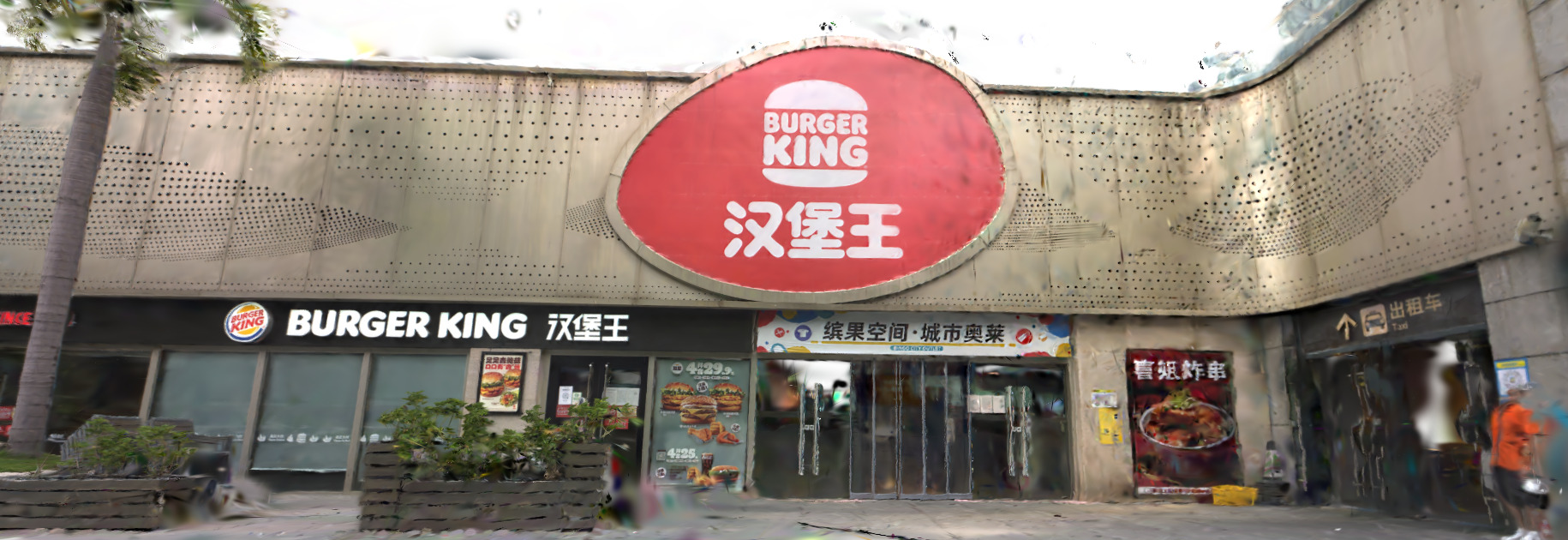}\hfill
  \zoomcelltag{\textbf{SpaRV}}{0.494\columnwidth}{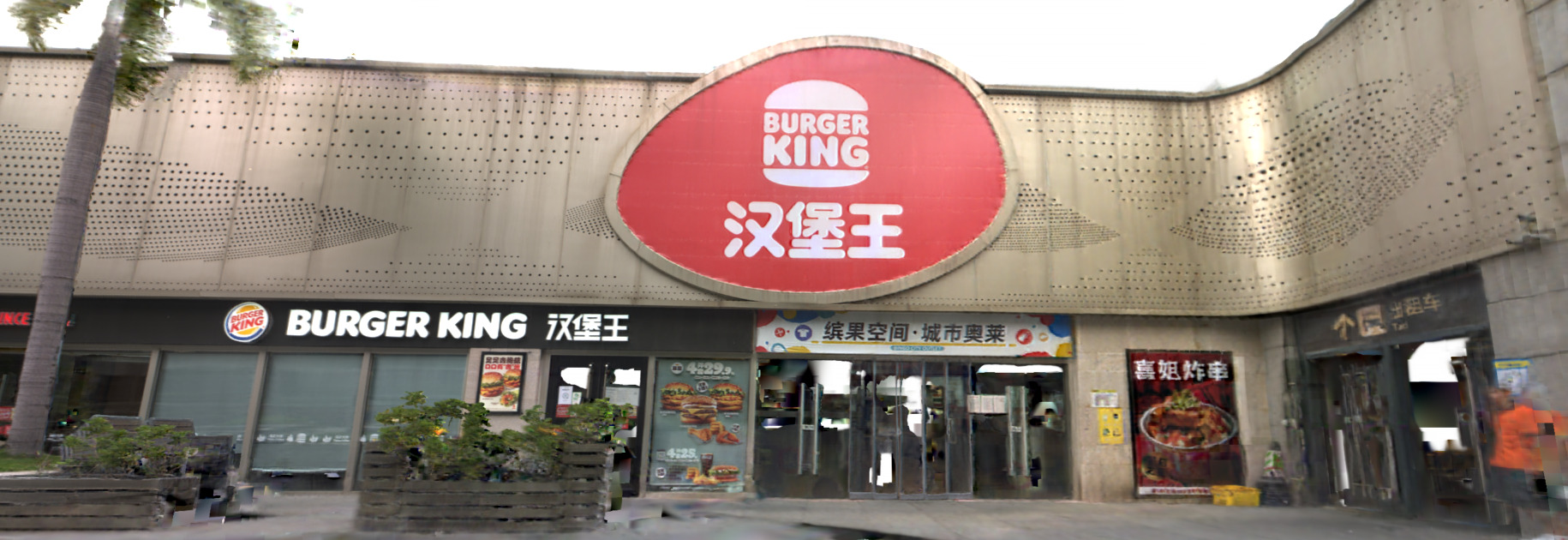}
  \vspace{-15pt}
  \caption{Rendering results on the FAST-LIVO2 Street scene at an extrapolated view away from the captured trajectory, with zoom crops.}
  \vspace{-10pt}
  \label{fig:fast_livo_qual_extrap}
\end{figure}

Radiance fields such as Neural Radiance Fields (NeRF)~\cite{mildenhall2021nerf} and 3D Gaussian Splatting (3DGS)~\cite{kerbl20233d} render photorealistic novel views by fitting a scene representation to posed images.
Fitted to robot data, however, a radiance field can reproduce the captured images and still fail once the viewpoint leaves the captured trajectory.
At an extrapolated view, the image-only methods break into noise and holes (Fig.~\ref{fig:fast_livo_qual_extrap}).
This paper asks how to build a radiance map from robot data that stays reliable at viewpoints away from the captured trajectory.
Radiance fields are usually built from curated captures with dense views, accurate camera parameters, and a scene of limited extent.
Robot data breaks each of these conditions, which raises three challenges to reliability: under-constrained geometry, inaccurate camera parameters, and limited GPU memory.

The first challenge is under-constrained geometry.
A robot's trajectory follows its task rather than the needs of reconstruction, so each region is observed from few and uneven viewpoints.
Photometry constrains depth only through the parallax between views, and a short baseline gives little parallax.
A radiance field then overfits the captured images by placing geometry at the wrong depth, and the misplaced geometry appears as floating artifacts away from the trajectory.
LiDAR measures depth directly, and prior work supervises the rendered depth with LiDAR~\cite{rematas2022urban,jiang2024li} or builds a geometric prior from it~\cite{cui2024letsgo,liu2025gs}.
Depth supervision, however, constrains only the rays that LiDAR measures, and camera rays between the sparse LiDAR points can still terminate at floating artifacts.
This motivates us to ground the geometry along the LiDAR rays and also on the camera rays that LiDAR does not measure.

Correct geometry must also render consistently as the viewpoint moves, which depends on the representation.
3DGS sorts its Gaussians by their centers for each view, and a change of viewpoint can invert this order, so the rendered surface pops~\cite{radl2024stopthepop}.
Popping persists even when LiDAR places the Gaussians on the true surface, as for LetsGo and GS-SDF in Fig.~\ref{fig:fast_livo_qual_extrap}.
NeRF avoids popping by compositing samples along each ray in the order the ray meets them, but its capacity is fixed by the network parameters.
Explicit spatial structures~\cite{takikawa2021neural,muller2022instant,fridovich2022plenoxels,sun2022direct} attach parameters to locations, so capacity can grow where the scene needs it.
Among them, SVRaster~\cite{svraster} uses non-overlapping adaptive sparse voxels, which keep the compositing order exact and add detail where required.
SVRaster renders these voxels by rasterization, however, which assumes a pinhole image grid and does not directly apply to LiDAR rays or non-pinhole cameras.
We therefore render the sparse voxels with a differentiable ray marcher that evaluates each ray independently, so the map can be supervised directly along the LiDAR rays.

The second challenge is inaccurate camera parameters.
Camera poses and intrinsics are difficult to obtain exactly in practice, whether they come from calibration~\cite{furgale2013unified}, Structure-from-Motion~\cite{schoenberger2016sfm}, SLAM~\cite{zheng2024fast}, or LiDAR-visual bundle adjustment~\cite{li2025lvba}.
A small error projects the same surface point to different pixels in different images.
A map fitted to these misaligned images tends to have blurred textures, even where LiDAR constrains its geometry well.
Prior work optimizes the cameras jointly with the field from little or no prior~\cite{lin2021barf,wang2021nerfmm}, or holds the trajectory fixed and refines only the LiDAR-camera extrinsics~\cite{herau2023moisst,yang2024unical}.
The cameras of robot data, in contrast, are metrically plausible but carry residual errors in every frame.
This motivates us to optimize the pose and focal length of each camera against the map, in alternation with map training.

The third challenge is limited GPU memory.
The GPU memory of a map grows with the extent of the scene.
Within a fixed GPU memory budget, a large scene can then be trained and rendered only at a coarser level of detail.
A single view, however, needs fine detail only near the camera, since distant regions project to few pixels.
Divide-and-conquer methods~\cite{tancik2022block,tao2025silvr} train parts of the scene separately, and level-of-detail methods~\cite{ren2024octree} select coarser primitives for distant regions at render time.
Hierarchical 3DGS~\cite{kerbl2024hierarchical} combines the two, but it needs extra stages after training: it merges the Gaussians of each part into a hierarchy, optimizes the merged nodes, and consolidates the parts into one hierarchy.
This motivates us to simplify the pipeline: the octree keeps every coarser level that training passes through, and each coarser voxel was already trained at its own level before it was subdivided, so it serves as that level of detail without further optimization.
One cut rule that selects voxels by their projected size then serves training as well as rendering.

To address these challenges, we present \textbf{SpaRV}, a \textbf{Spa}rse \textbf{R}adiance \textbf{V}oxel map for scalable, geometrically consistent LiDAR-visual mapping and rendering.
Our contributions are as follows:
\begin{enumerate}
  \item Ray-marching rendering of sparse radiance voxels: A differentiable ray marcher lets camera pixels, LiDAR rays, and rays of non-pinhole cameras supervise the map through a unified renderer (Secs.~\ref{sec:preliminaries}--\ref{sec:rendering}).
  \item Geometry-grounded map training: LiDAR depth supervision, occupied voxel guidance, and normal and smoothness regularizers keep the map geometrically consistent beyond the captured trajectory (Sec.~\ref{sec:methodology}).
  \item Map-driven camera optimization: SpaRV alternates map training with photometric optimization of the camera poses and focal lengths, which reduces the texture blur caused by residual camera errors (Sec.~\ref{sec:pose_refinement}).
  \item Large-scale reconstruction under bounded GPU memory: One octree and one granularity-driven cut serve training and rendering, holding the submap in focus at the finest available level and the rest of the scene at coarser levels (Sec.~\ref{sec:large_scale}).
\end{enumerate}

We evaluate SpaRV on synthetic indoor scenes, two outdoor LiDAR-visual datasets, an omnidirectional benchmark, and two large scenes we collected.
At viewpoints away from the captured trajectory, SpaRV keeps the geometry consistent, while image-only baselines degrade and Gaussian-based baselines pop.
Camera optimization sharpens textures on real data, and the optimized cameras also improve the baselines retrained on them.
On a single GPU, SpaRV reconstructs and renders a kilometer-scale university campus.
The trained maps also support surface extraction and the simulation of other camera and LiDAR models.

\section{Related Work}
\label{sec:related_work}

\subsection{Radiance Field Representations}

Radiance-field representations store scene parameters either implicitly in network weights or explicitly at spatial locations.
Neural Radiance Fields (NeRF)~\cite{mildenhall2021nerf} established continuous volumetric modeling for photorealistic novel view synthesis.
A multilayer perceptron maps a position and viewing direction to density and color.
Methods that store learned features in hash grids~\cite{muller2022instant} or factorized tensors~\cite{chen2022tensorf} retain a network only as a decoder, which reduces training cost.
A shared decoder, however, entangles those features: an update in one region perturbs weights used throughout the field, so previously reconstructed regions may drift as new observations arrive~\cite{sucar2021imap,zhu2022nice}.
The parameter is also fixed for the whole field, so capacity cannot readily be increased locally where additional detail is required.

3D Gaussian Splatting (3DGS)~\cite{kerbl20233d} makes the parameters explicit and discrete.
The scene becomes a set of anisotropic Gaussian primitives, each with its own position, shape, opacity, and appearance.
A highly efficient differentiable rasterizer sorts the primitives and blends them into the image.
The primitives remain independent, so the field they assemble has no intrinsic compositing order: the per-view sort on their centers can invert under viewpoint change and the rendered surface pops~\cite{radl2024stopthepop}.
Voxel-based radiance fields store parameters explicitly on a voxel grid.
Queries are answered by interpolating vertex values, without a decoder network.
NSVF~\cite{liu2020neural} attached learned features to a sparse voxel octree, DVGO~\cite{sun2022direct} optimized an explicit density grid, and Plenoxels~\cite{fridovich2022plenoxels} removed the network altogether, reducing optimization from hours to minutes.
SVRaster~\cite{svraster} further makes those voxels adaptive: non-overlapping sparse voxels across levels of detail, with efficient rasterization and an exact compositing order that avoids popping.
Its rasterizer, however, assumes a pinhole image grid and does not directly apply to LiDAR rays or non-pinhole cameras.
In contrast, our method shares the adaptive sparse-voxel representation of SVRaster but differs from it in three designs.
A differentiable ray marcher replaces the rasterizer, so camera pixels, LiDAR rays, and rays of non-pinhole cameras supervise the map through the unified renderer (Sec.~\ref{sec:rendering}).
LiDAR measurements ground the geometry (Sec.~\ref{sec:methodology}), and a granularity-driven octree cut keeps GPU memory bounded in large scenes (Sec.~\ref{sec:large_scale}).

\begin{figure*}[!t]
  \centering
  \includegraphics[width=\linewidth]{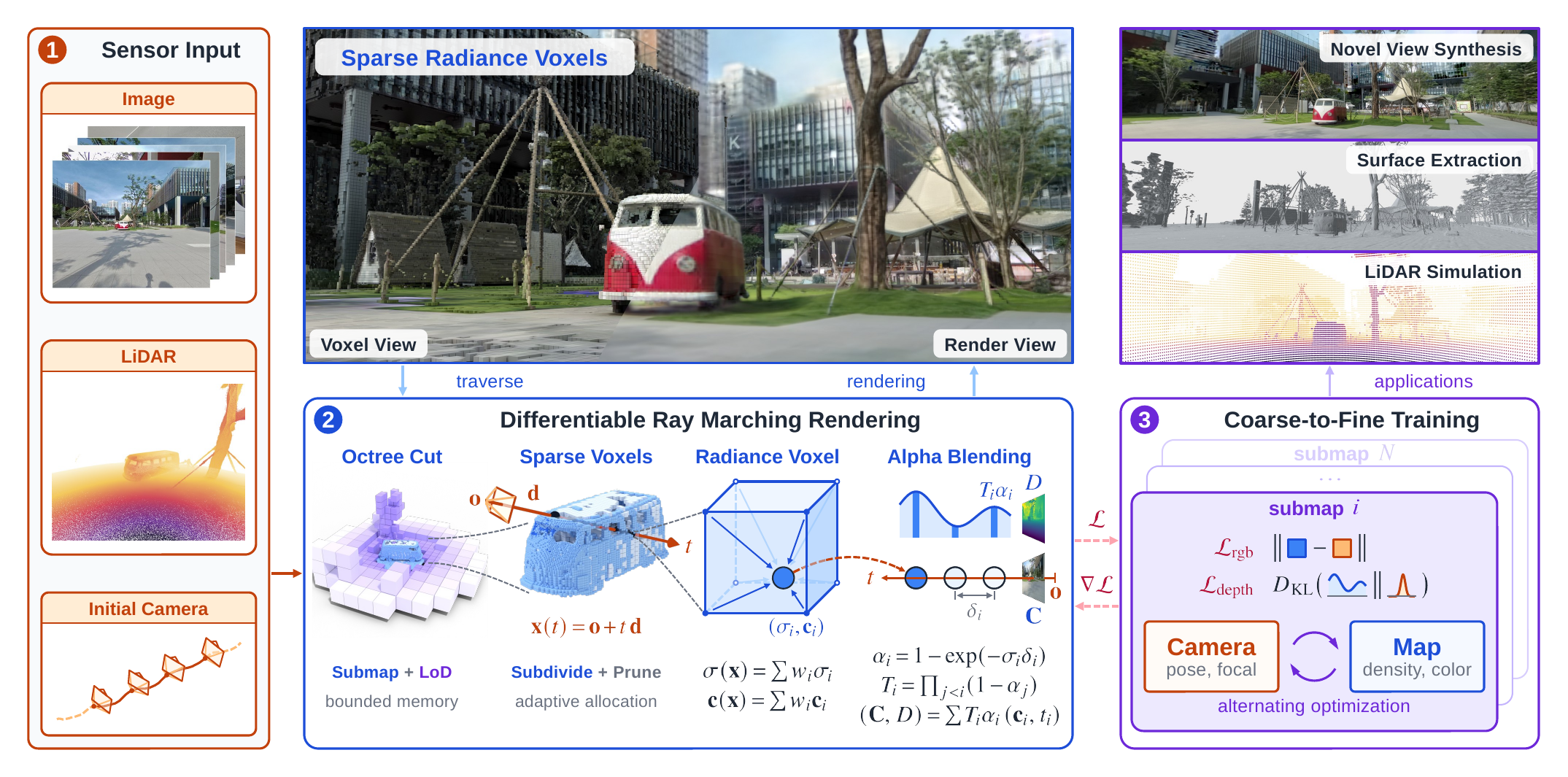}
  \vspace{-25pt}
  \caption{Overview of SpaRV. From RGB images, LiDAR scans, and initial camera parameters~\textbf{(1)}, SpaRV builds a sparse radiance voxel map for geometrically consistent rendering. The top panel shows one frame as its allocated voxels (left) and the view rendered from them (right).
  \textbf{(2)}~A granularity-driven cut selects voxels from the octree, and a differentiable ray marcher renders them into color $\pixrgb$ and depth $D$.
  \textbf{(3)}~Training runs over the submaps coarse-to-fine and alternates map training with camera optimization.
  The trained map is applied to novel view synthesis, surface reconstruction, and LiDAR simulation.}
  \label{fig:overview}
    \vspace{-10pt}
\end{figure*}

\subsection{Geometric Grounding in Radiance Fields}

A photometric loss alone leaves geometry under-constrained.
2DGS~\cite{huang20242d} collapses each Gaussian primitive to an oriented disk and aligns the normals derived from the rendered depth to the disk normals.
PGSR~\cite{chen2024pgsr} further introduces unbiased depth rendering with multi-view photometric regularization to preserve global geometric accuracy.
They require the rendered geometry to be self-consistent, which improves the quality of the surface reconstruction.
DS-NeRF~\cite{deng2022depth} reuses the sparse 3D points that Structure-from-Motion (SfM)~\cite{schoenberger2016sfm} has already produced, which anchors the scene geometry to the observation.
This supervision inherits what SfM has produced, so it thins out exactly where correspondence fails, which is where geometry is hardest to recover.
Monocular depth predictors such as Depth Anything~\cite{yang2024depth} provide dense supervision~\cite{xu2025depthsplat}, but their depth is not guaranteed to be accurate or metrically consistent across images.

Range sensors measure depth directly: each range measurement gives the metric distance to a surface along its own ray, independent of texture.
The most common approach is to penalize the difference between rendered depth and measured depth.
This loss is used wherever such measurements are available, across implicit fields~\cite{rematas2022urban,ziyang2023snerf,tao2025silvr,sucar2021imap,zhu2022nice,jiang2023h} and 3D Gaussian representations~\cite{jiang2024li,lang2025gaussianlic}.
The alternative is to reconstruct an explicit or implicit geometry from the depth measurements and use it to constrain the radiance field.
LetsGo~\cite{cui2024letsgo} uses such a geometry as a starting point.
It meshes the LiDAR point cloud by Poisson reconstruction and initializes Gaussians from that mesh.
A depth term on the measurements then constrains the rendered depth during training.
LIV-GaussMap~\cite{hong2024liv} and LI-GS~\cite{jiang2024li} constrain splat orientation and distribution with LiDAR-derived planar priors and Gaussian mixtures.
M2Mapping~\cite{liu2024neural} and GS-SDF~\cite{liu2025gs} instead build a neural SDF from the range measurements.
GS-SDF further fits the Gaussians to its isosurface to improve the geometric consistency of 3DGS.
In contrast, our method grounds the sparse-voxel field itself: LiDAR depth supervision acts directly along the LiDAR rays, while occupied voxel guidance and the normal and smoothness regularizers constrain the field where LiDAR does not measure.

\vspace{-10pt}
\subsection{Camera Optimization in Radiance Fields}

Radiance-field reconstruction depends on accurate camera parameters.
SfM~\cite{schoenberger2016sfm} is the common source, but it is computationally expensive and brittle in textureless or repetitive scenes.
SLAM systems~\cite{xu2022fast,zheng2024fast} produce poses efficiently in these environments, yet their residual error still exceeds the pixel-level accuracy a radiance field requires.
Joint photometric optimization against the field corrects such errors, and existing methods differ mainly in what they start from.
Some recover the cameras with little or no prior.
Starting from rough or identity poses, they optimize the camera parameters jointly with the field through the photometric loss: the poses in BARF~\cite{lin2021barf,chen2023local}, and also the focal length in NeRF-{}-~\cite{wang2021nerfmm}.
SCNeRF~\cite{jeong2021self} additionally calibrates the intrinsics and the lens distortion, and InstantSplat~\cite{fan2024instantsplat} optimizes the poses jointly with 3D Gaussians from sparse, unposed views.
Others estimate the poses incrementally as the frames arrive.
COLMAP-free 3DGS~\cite{fu2024colmap} holds the camera fixed and instead estimates the rigid transform of the Gaussians between consecutive frames, which is the relative pose.
3DGS SLAM systems~\cite{yan2024gs,GSSLAM2024}, mostly with RGB-D input, track each incoming camera by minimizing the photometric and depth residuals against the map rendered at its pose, and then extend the map with that frame.
A third group refines cameras that are already approximately known, which is the setting closest to ours.
CamP~\cite{park2023camp} refines the poses and intrinsics of SfM cameras jointly with the field, and reparameterizes the cameras to make this optimization more stable.
With photometric supervision alone, however, the joint optimization can still converge to misaligned geometry.
With LiDAR, MOISST~\cite{herau2023moisst} and UniCal~\cite{yang2024unical} optimize a neural scene representation jointly with the extrinsics between the sensors, while holding the sensor trajectory from SLAM or localization fixed.
In contrast, our method refines every camera of the input trajectory, such as one from LiDAR-inertial-visual odometry~\cite{zheng2024fast}, whose poses are metrically plausible but carry residual errors in every frame.
It optimizes the cameras against a map whose geometry is anchored by LiDAR.
The cameras and the map are updated in alternation.
Each update corrects the pose and focal length of a camera by an error-state step around its initialization.

\vspace{-10pt}
\subsection{Large-Scale Radiance Field Reconstruction}

Scaling radiance fields to large scenes raises two distinct issues: representing unbounded outdoor extents, and keeping active GPU memory bounded during training and rendering.
NeRF++~\cite{zhang2020nerf++} and Mip-NeRF 360~\cite{barron2022mip} address unbounded extents by wrapping the whole space with a single non-linear parameterization into a bounded domain, so the field can be queried anywhere.
3DGS does not require this contraction: Gaussian positions live in Euclidean world space, so distant content is represented by placing primitives farther away rather than by warping the domain.
Active GPU memory is kept bounded by divide-and-conquer training and level-of-detail (LoD) rendering.
Divide-and-conquer methods such as Block-NeRF~\cite{tancik2022block} and SiLVR~\cite{tao2025silvr} decompose a large-scale capture into individually trained NeRFs, which decouples rendering time from scene size and lets one block be updated independently.
LoD rendering methods such as Octree-GS~\cite{ren2024octree} organize the Gaussians into multi-resolution anchors and select a level per view, so the number of primitives inside the viewing frustum stays stable as the viewpoint zooms out.
Hierarchical 3DGS (H3DGS)~\cite{kerbl2024hierarchical} combines the two.
It partitions the capture into parts trained independently, merges the Gaussians of each part into coarser intermediate nodes, and consolidates the parts into one hierarchy that is optimized further.
Rendering selects Gaussian nodes through this hierarchy by the projected size of its Gaussians and interpolates between levels for smooth transitions.
In contrast, our method addresses both issues.
For unbounded extents, it follows the NeRF line and queries its bounded voxel domain in contracted space.
For GPU memory, a granularity-driven cut of one octree holds the submap in focus at the finest available level and the rest of the scene at coarser levels, in training as well as in rendering.

\section{Sparse Radiance Voxel Representation}
\label{sec:preliminaries}
\label{sec:overview}

Given RGB images and LiDAR points, together with initial camera parameters, SpaRV reconstructs a sparse radiance voxel map for geometrically consistent rendering (Fig.~\ref{fig:overview}).
A granularity-driven cut selects the resident voxels from the octree by their projected size, so GPU memory stays bounded as the scene grows (Sec.~\ref{sec:large_scale}).
The selected voxels form an adaptive sparse-voxel map, with finer voxels where additional detail is required (Sec.~\ref{sec:sparse_voxel_grid}).
A differentiable ray marcher traverses these resident voxels along each ray independently, and therefore renders camera images and irregular LiDAR scans in their native sampling patterns (Sec.~\ref{sec:rendering}).
Both then supervise the map through the same renderer.
SpaRV alternates between geometry-grounded map training (Sec.~\ref{sec:methodology}) and camera optimization against the map (Sec.~\ref{sec:pose_refinement}).
The trained map supports novel view synthesis, surface extraction, and LiDAR simulation.

\begin{figure}[t]
  \centering
  \includegraphics[width=\linewidth]{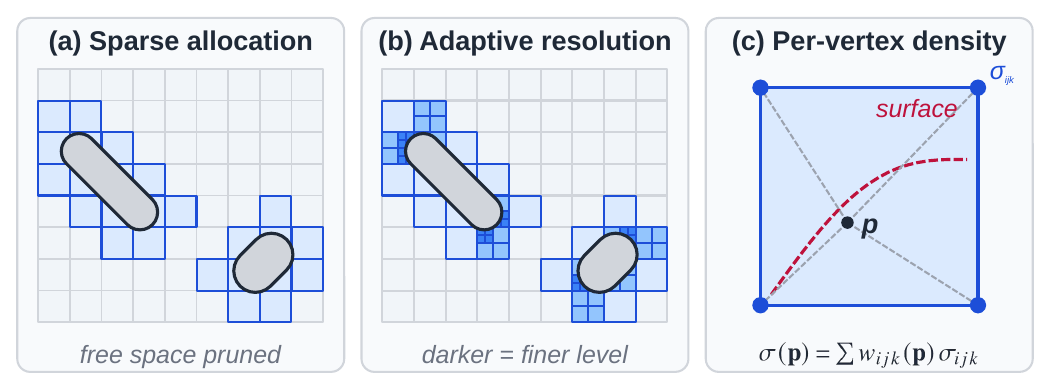}
  \vspace{-20pt}
  \caption{Sparse voxels on a two-object scene. (a)~Voxels are allocated only around the surfaces. (b)~Finer voxels (darker) concentrate where the detail is required. (c)~Trilinear interpolation of the eight vertex densities.}
  \label{fig:sparse_voxel_structure}
\end{figure}

\subsection{Sparse Voxels}
\label{sec:sparse_voxel_grid}

SVRaster allocates voxels following an octree space-partition layout.
A maximum level of detail $\maxlv$ defines the finest voxel resolution at $(2^\maxlv)^3$.
Let $\worldsiz \in \Real$ be the map size and $\worldcen \in \Real^3$ be its center in world space.
The voxel index $\voxidx = \{i,j,k\} \in [0, \ldots, 2^\maxlv{-}1]^3$ together with an octree level $l \in [1, \maxlv]$ define the voxel size $\voxsiz$ and voxel center $\voxcen$ as:
\begin{equation}
  \voxsiz = 2^{-l}\,\worldsiz, \quad
  \voxcen = \worldcen - 0.5\,\worldsiz + (\voxidx + 0.5)\,\voxsiz ~.
  \label{eq:voxel_size_center}
\end{equation}
Internally, this index is mapped to a Morton code for efficient spatial indexing. The voxels are kept in a compact flat 1D array indexed by $(\text{octpath}, \text{octlevel})$ pairs.
Fig.~\ref{fig:sparse_voxel_structure}~(a)--(b) illustrate the resulting sparsity and the detail-adaptive resolution obtained by the voxel adaptation of Sec.~\ref{ssec:voxel_init_adapt}.

The density field is piecewise trilinear: the value at a query point is determined by the voxel that contains it, with closed-form spatial derivatives inside each voxel (Fig.~\ref{fig:sparse_voxel_structure}~(c)).
Learnable densities reside at the vertices, which adjacent voxels share. For a query point $\pt$ inside a voxel, the density is the trilinear interpolant of the eight vertex values:
\begin{equation}
  \sigma(\pt) = \interp(\voxgeo, \pt) = \sum_{i,j,k \in \{0,1\}} w_{ijk}(\pt)\,\sigma_{ijk} ~,
  \label{eq:density_interp}
\end{equation}
where $\voxgeo \in \Real^{2 \times 2 \times 2}$ holds the densities $\sigma_{ijk}$ at the eight voxel vertices~\cite{svraster} and $w_{ijk}(\pt)$ are the trilinear weights of $\pt$.

For view-dependent appearance, SVRaster stores spherical harmonic (SH) coefficients per voxel.
A per-voxel coefficient is constant inside the voxel, giving no gradient with respect to the sample position.
Camera optimization backpropagates the photometric residual through that position and requires an appearance gradient with respect to it.
SpaRV stores the degree-0 (DC) component $\vox_{\mathrm{sh}_0}$ at the voxel vertices and interpolates it like the density, while the higher-order terms remain per voxel:
\begin{equation}
  \mathrm{SH}_0(\pt) = \interp(\vox_{\mathrm{sh}_0}, \pt) ~.
  \label{eq:sh_interp}
\end{equation}
The interpolated DC field supplies that gradient throughout the voxel, and directional shading stays compact.

\begin{figure}[t]
  \centering
  \includegraphics[width=\linewidth]{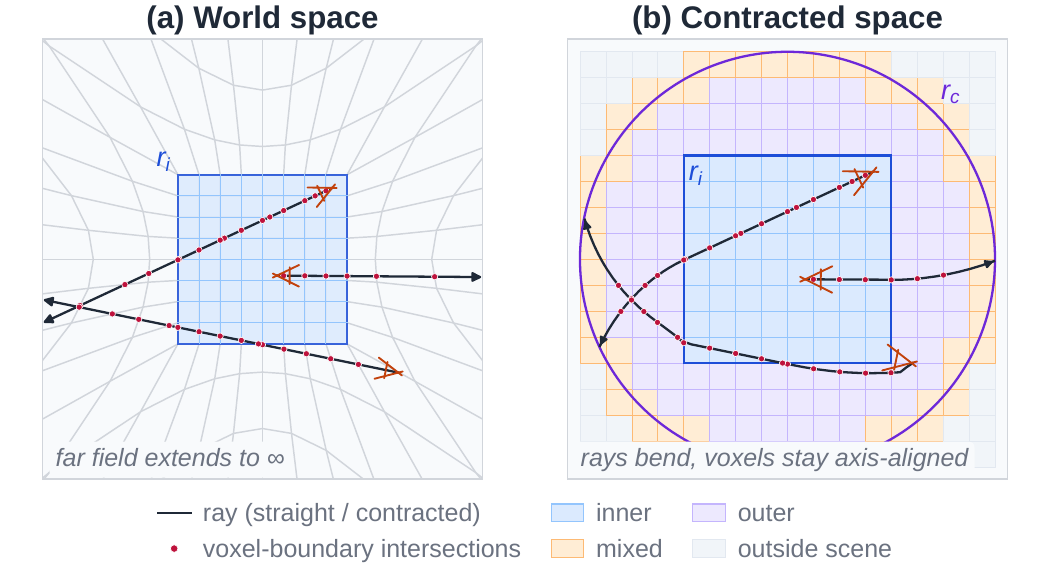}
  \vspace{-20pt}
  \caption{Contraction map $\contract$ in 2D.
  (a)~World space: rays are straight, the identity box of half-extent $r_i$ is undistorted, and the warped regions map to axis-aligned voxels in contracted space. (b)~Contracted space: the same rays curve, and the voxels fall into the inner, mixed, outer, and outside-scene classes. Dots mark corresponding voxel-boundary intersections.}
  \label{fig:contraction}
\end{figure}

\subsection{Contracted-Space Representation}
\label{sec:contraction_space}

The sparse-voxel map occupies a bounded domain of size $\worldsiz$.
Outdoor scenes, however, contain distant content that lies outside this domain.
SVRaster~\cite{svraster} covers such content by enlarging the domain in world space by a fixed number of octree levels.
At a fixed maximum level $\maxlv$, voxel size scales with $\worldsiz$ (Eq.~\eqref{eq:voxel_size_center}), so each added level doubles the size of the finest voxels in the near field, and content beyond the enlarged bound can only be fitted onto the outermost voxels, without correct geometry.
Following unbounded-scene radiance-field practice~\cite{zhang2020nerf++,barron2022mip}, SpaRV therefore maps the unbounded world into a finite domain by parameterizing the voxels in a contracted space as shown in Fig.~\ref{fig:contraction}.
Let $\mathbf{x} \in \Real^{3}$ be a world-space point, $r_i > 0$ the half-extent of the inner identity box centered at $\worldcen$, and $r_c = 0.5\,\worldsiz$ the outer contraction radius.
We define a contraction map $\contract: \mathbf{x} \mapsto \tilde{\mathbf{x}}$ such that
\begin{equation}
  \tilde{\mathbf{x}} = \contract(\mathbf{x}) =
  \begin{cases}
    \mathbf{x}, & \|\mathbf{x} - \worldcen\|_{\infty} \le r_i, \\
    \worldcen + s(\mathbf{x}) \, (\mathbf{x} - \worldcen), & \text{otherwise}.
  \end{cases}
  \label{eq:contraction_map}
\end{equation}
Let $\boldsymbol{\delta} = \mathbf{x} - \worldcen$, $m = \|\boldsymbol{\delta}\|_\infty$, and $\rho = \|\boldsymbol{\delta}\|_2$. For $\|\boldsymbol{\delta}\|_{\infty} > r_i$, the scale factor is
\begin{equation}
  s(\mathbf{x}) = s(m, \rho) =
  \frac{m^2\,r_c - r_i^2\,\rho}{m\,\left(m\,(r_c + \rho) - 2\,r_i\,\rho\right)} ~.
  \label{eq:contraction_scale_prelim}
\end{equation}
The near-field geometry is represented without distortion, while distant scene content is compressed into a finite domain and can still be indexed by the same sparse voxels.
Voxels fall into four classes relative to the identity box and the outer radius $r_c$: \emph{inner} voxels lie entirely inside the identity box, \emph{outer} voxels entirely outside it, \emph{mixed} voxels straddle its boundary, and \emph{outside-scene} voxels fall beyond $r_c$. Fig.~\ref{fig:contraction}~(b) shows these four classes. The classification determines how a ray leaves a voxel during marching (Sec.~\ref{sec:raymarch}).

\begin{algorithm}[t]
  \caption{Sparse Voxel Query}
  \label{alg:octree_query}
  \KwIn{Query point $\rayq$, sorted occupied octpaths $\{p_i\}_{i=0}^{M-1}$, and their levels $\{l_i\}_{i=0}^{M-1}$.}
  \KwOut{Voxel level $l$, octpath $p$, and occupancy flag $\delta \in \{0,1\}$.}
  Let $\left[I^-(p,l), I^+(p,l)\right)$ denote the finest-level Morton interval covered by voxel $(p,l)$\;

  $q \leftarrow \text{MortonEncode}(\rayq, \maxlv)$\label{algline:morton}\;
  Binary search the predecessor $(p^-, l^-)$ and successor $(p^+, l^+)$ of $q$\label{algline:bsearch}\;

  \If{$p^+ = q$}{
    \Return $(l^+, p^+, 1)$\label{algline:hit}\tcp*{exact leaf hit}
  }

  \If{$I^-(p^-, l^-) \le q < I^+(p^-, l^-)$}{
    \Return $(l^-, p^-, 1)$\label{algline:cover}\tcp*{the predecessor leaf covers the query}
  }

  \For{$\ell = 0$ \KwTo $\maxlv$}{
    $p \leftarrow \text{MortonEncode}(\rayq, \ell)$\;
    \If{$I^+(p^-, l^-) \le I^-(p,\ell)$ and $I^+(p,\ell) \le I^-(p^+, l^+)$}{
      \Return $(\ell, p, 0)$\label{algline:empty}\tcp*{largest empty voxel containing $q$}
    }
  }
\end{algorithm}

\vspace{-10pt}
\subsection{Octree Query}
\label{sec:octree_query}

LiDAR supervision and camera optimization query the field along independent rays in arbitrary directions.
Most of a ray crosses free space where no voxel is materialized, so empty space must be skipped in large steps. SpaRV therefore queries the field by an octree lookup, and rendering marches over the resulting voxel segments (Sec.~\ref{sec:rendering}). Since only occupied leaves are stored in a flat Morton-indexed array, the renderer does not perform a pointer-based hierarchical octree walk.
Instead, Alg.~\ref{alg:octree_query} maps each query point to its finest-level Morton code and locates it in the sorted leaf set by binary search (lines~\ref{algline:morton}--\ref{algline:bsearch}). If the queried code matches a stored leaf, or falls inside the finest-level Morton interval covered by the predecessor leaf, the ray enters that occupied voxel directly (lines~\ref{algline:hit} and~\ref{algline:cover}). Otherwise, the predecessor and successor occupied leaves bound the query in Morton order, and the largest empty voxel between them is skipped (line~\ref{algline:empty}). The marcher can therefore skip free space without traversing unoccupied fine voxels individually. The lookup thus returns either an occupied voxel to be evaluated or an inferred empty skip voxel that advances the ray to the next relevant event. The event-driven ray marcher of Sec.~\ref{sec:rendering} runs entirely on this occupied-or-empty interface.

\section{Sparse Radiance Voxel Rendering}
\label{sec:rendering}

\begin{figure}[h]
  \centering
  \includegraphics[width=\linewidth]{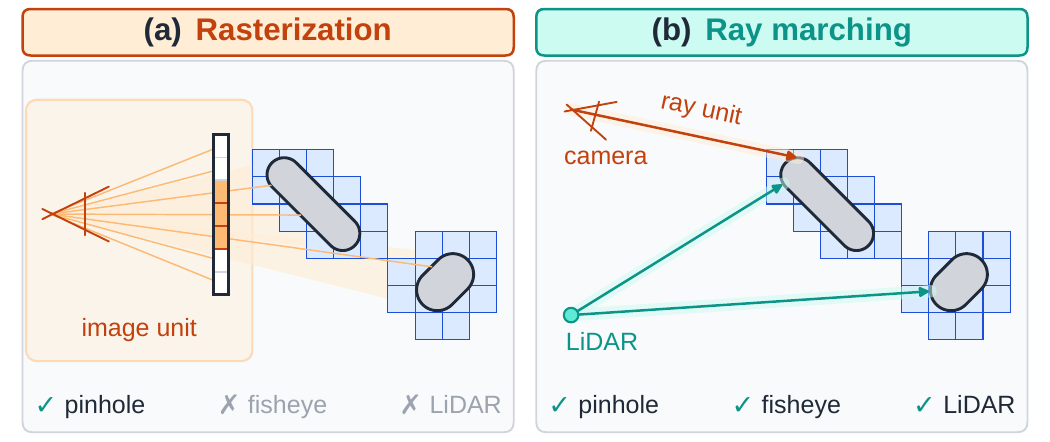}
  \vspace{-15pt}
  \caption{Rasterization versus ray marching. (a)~Rasterization batches a pinhole pixel lattice. (b)~Ray marching evaluates each ray independently.}
  \label{fig:raster_vs_raymarch}
\end{figure}

SpaRV renders the sparse voxels with a differentiable ray marcher that traverses them along each ray independently. Camera pixels, LiDAR rays, and rays of non-pinhole cameras then supervise the map through the same renderer. In contrast, the rasterizer of SVRaster~\cite{svraster} batches a pinhole pixel lattice and does not directly apply to LiDAR rays or non-pinhole cameras (Fig.~\ref{fig:raster_vs_raymarch}). Rendering converts the occupied-or-empty lookup of Sec.~\ref{sec:octree_query} into pixel values through event-driven ray marching and volumetric alpha blending.

\subsection{Ray Marching Rendering}
\label{sec:raymarch}

Ray marching proceeds by advancing from one returned voxel event to the next, as summarized in Alg.~\ref{alg:raymarch}. At each step, the current world-ray parameter $t$ determines the query point $\mathbf{x}(t)=\contract(\mathbf{o}+t\mathbf{d})$, and the lookup returns either an occupied voxel to be rendered or an empty skip voxel to be traversed (line~\ref{algline:query}). In either case, the next step requires the exit parameter $t_{\mathrm{exit}}$ at which the ray leaves the voxel. How it is found depends on the voxel's class relative to the identity box and the contracted domain (line~\ref{algline:class}): inner, outer, mixed, or outside-scene, as drawn in Fig.~\ref{fig:contraction}~(b).
For inner voxels, the exit is a ray--box intersection (line~\ref{algline:raybox}).
In outer and mixed voxels, the straight world ray maps to a curve in contracted space, so the exit has no closed-form ray--box solution.
For $m > r_i$, the Jacobian of Eq.~\eqref{eq:contraction_map} is
\begin{equation}
  \mathbf{J}(\mathbf{x}) = s(m, \rho)\,\mathbf{I} + (\mathbf{x} - \worldcen) \otimes \nabla s(\mathbf{x}) ~,
  \label{eq:jacobian}
\end{equation}
where $\nabla s$ is the gradient of the scale factor in Eq.~\eqref{eq:contraction_scale_prelim}.
The Jacobian maps the world direction $\mathbf{d}$ to a local contracted direction.
The curved-exit solve (line~\ref{algline:curved}) uses a slab test along $\mathbf{J}\mathbf{d}$ to propose a candidate and evaluates $\contract$ exactly to accept or reject it.
Once an occupied segment is identified, SpaRV divides the secant between its mapped entry and exit points into equal steps and evaluates density and appearance at each step's midpoint (line~\ref{algline:secant}). These samples are then alpha-blended front to back (Sec.~\ref{sec:alpha_blending}), which accumulates the color and depth of the ray and attenuates its transmittance (line~\ref{algline:blend}).

\begin{algorithm}[t]
  \caption{Ray Marching Rendering}
  \label{alg:raymarch}
  \KwIn{Ray origin $\mathbf{o}$, direction $\mathbf{d}$, sparse voxels.}
  \KwOut{Color $\pixrgb$, depth $D$, transmittance $T$.}
  Let $\mathbf{x}(t) = \contract(\mathbf{o}+t\mathbf{d})$ and let $B(l,p)$ be the contracted-space box of voxel $(l,p)$\;

  $t \leftarrow t_{\mathrm{near}}$, the entry of the world ray into the scene bounds\;
  $\pixrgb \leftarrow \mathbf{0}$, $D \leftarrow 0$, $T \leftarrow 1$\;

  \While{$T > T_{\min}$}{
    $(l, p, \delta) \leftarrow \text{OctreeQuery}(\mathbf{x}(t))$\label{algline:query}\tcp*{Alg.~\ref{alg:octree_query}}
    $\kappa \leftarrow \text{VoxelClass}(l,p)$\label{algline:class}\;
    \If{$\kappa = \text{outside}$}{
      \Return $\pixrgb, D, T$\;
    }
    \eIf{$\kappa = \text{inner}$}{
      $t_{\mathrm{exit}} \leftarrow \text{RayBoxExit}\big(t, B(l,p)\big)$\label{algline:raybox}\; %
    }{
      $t_{\mathrm{exit}} \leftarrow \text{CurvedExit}\big(t, B(l,p), \delta\big)$\label{algline:curved}\;
    }
    \If(\tcp*[f]{occupied voxel}){$\delta = 1$}{
      $\mathbf{x}_k \leftarrow \mathbf{x}(t) + \frac{k-1/2}{\nsamp}\bigl(\mathbf{x}(t_{\mathrm{exit}})-\mathbf{x}(t)\bigr),\ k=1,\ldots,\nsamp$\label{algline:secant}\;
      $(\pixrgb, D, T) \leftarrow \text{AlphaBlending}\big(\{\mathbf{x}_k\}, l, p\big)$\label{algline:blend}\; %
    }
    $t \leftarrow t_{\mathrm{exit}} + \epsilon(t_{\mathrm{exit}})$ \tcp*{cross the boundary}
  }
  \Return $\pixrgb, D, T$\;
\end{algorithm}

\vspace{-10pt}
\subsection{Sparse-Voxel Alpha Blending}
\label{sec:alpha_blending}

For a point $\pt$ inside a voxel, the density is obtained by trilinear interpolation $\interp(\voxgeo, \pt)$ (Eq.~\eqref{eq:density_interp}), followed by an exponential-linear activation~\cite{sun2022direct, karnewar2022relu} to ensure non-negativity.
Each ray--voxel intersection segment of Sec.~\ref{sec:raymarch} is discretized into $\nsamp$ midpoint samples: in the $i$-th intersected voxel, the $k$-th sample is located at $\mathbf{x}_{i,k}$ with spacing $\Delta t_i$, the segment length divided by $\nsamp$. The sample opacity is computed as
\begin{equation}
  \alpha_{i,k}
  = 1 - \exp\ \!\Bigl(
  -\Delta t_i \,
  \explin\bigl(\interp({\voxgeo}_{i}, \mathbf{x}_{i,k})\bigr)
  \Bigr) ~.
  \label{eq:voxel_alpha}
\end{equation}
The view-dependent color $\mathbf{c}_{i,k}$ is evaluated at the same sample using the query-ray direction relative to the voxel. The DC component $\mathrm{SH}_0$ is trilinearly interpolated from the vertex-stored coefficients at $\mathbf{x}_{i,k}$, while the higher-order SH coefficients remain shared by the voxel. Spherical-harmonic evaluation then combines the two into $\mathbf{c}_{i,k}$. The low-frequency color therefore varies continuously within each voxel, and view-dependent shading stays compact.

For a ray passing through $N$ intersected voxels, the rendered pixel color $\pixrgb$ is accumulated over the voxel-sample sequence encountered along the ray:
\begin{align} \label{eq:alpha_blend}
    \pixrgb
    &= \sum\nolimits_{i=1}^{N} \sum\nolimits_{k=1}^{\nsamp}
    T_{i,k}\,\alpha_{i,k}\,\mathbf{c}_{i,k} ~, \\
    T_{i,k}
    &= \prod\nolimits_{(j,l) \prec (i,k)}
    \left(1 - \alpha_{j,l}\right) ~,
\end{align}
where $\prec$ denotes the ray-marching sampling order.
Unlike autograd implementations such as NeRF~\cite{mildenhall2021nerf}, which store every sample for back-propagation, SpaRV differentiates Eq.~\eqref{eq:alpha_blend} in a custom CUDA kernel whose memory grows with the number of rays rather than samples.

\section{Geometry-Grounded Map Training}
\label{sec:methodology}

This section trains the sparse radiance voxel map from RGB images, LiDAR points, and the initial camera parameters, and adapts its voxels to the scene during training (Sec.~\ref{ssec:voxel_init_adapt}).
The goal is a map that renders the captured views photorealistically and keeps correct geometry in regions that the images alone under-constrain, so that it stays geometrically consistent beyond the captured trajectory.
To this end, photometric supervision fits the images, LiDAR depth supervision and occupied voxel guidance anchor the density field at the measured surfaces, and normal and smoothness regularizers keep the surface smooth between LiDAR points (Sec.~\ref{ssec:rf_optimization}).

\vspace{-10pt}
\subsection{Voxel Initialization and Adaptation}
\label{ssec:voxel_init_adapt}

SpaRV inherits the voxel initialization and adaptation scheme of SVRaster~\cite{svraster}. Initialization uses the same foreground-background layout, with voxel maximum sampling rate as the guiding quantity for image-space coverage.
The implementation starts from uniform dense voxels and iteratively subdivides the voxels with the highest sampling rate until the initial target voxel budget is reached. 
Voxels with zero camera visibility are then removed. This initialization provides a coarse scene-covering layout that is subsequently refined by adaptive subdivision and pruning during training.

During training, we adopt the same adaptive structural update criteria. 
At each adaptation, voxels whose maximum blending weight $T_{i,k}\alpha_{i,k}$ in Eq.~\eqref{eq:alpha_blend} across training views falls below a threshold are pruned.
Voxels are thus kept only where the rendering needs them, which makes the map sparse, as shown in Fig.~\ref{fig:sparse_voxel_structure}~(a).
For subdivision, each leaf voxel is ranked by the accumulated alpha-weighted loss gradient
\begin{equation}
  \voxpriority = \sum_{\ray \in R} \left\| \alpha(\ray)\,\frac{\partial \loss(\ray)}{\partial \alpha(\ray)} \right\| ~,
  \label{eq:voxel_priority}
\end{equation}
where $R$ denotes the training rays collected over one adaptation interval and $\loss(\ray)$ is the per-ray training loss defined in Sec.~\ref{ssec:rf_optimization}. 
The priority selects which voxels are subdivided, and the maximum sampling rate suppresses subdivision in poorly sampled regions.
Finer voxels therefore concentrate where detail is required, as shown in Fig.~\ref{fig:sparse_voxel_structure}~(b).

\vspace{-10pt}
\subsection{Radiance Field Training}
\label{ssec:rf_optimization}

\begin{figure}[t]
  \centering
  \includegraphics[width=\linewidth]{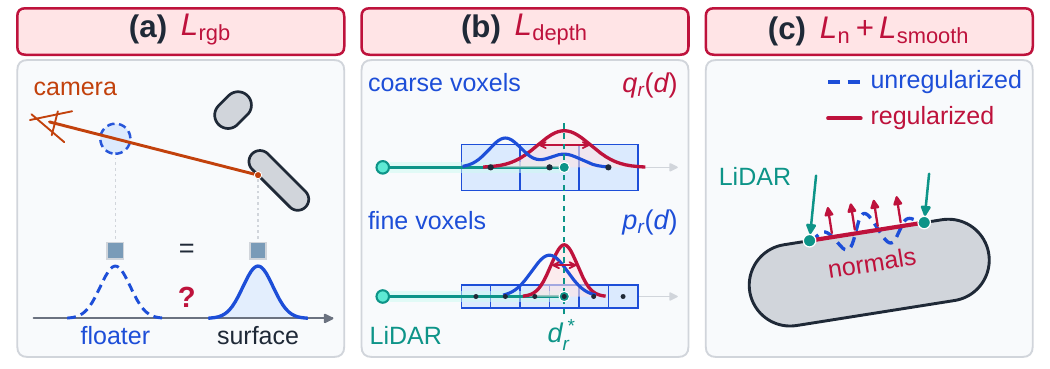}
  \vspace{-15pt}
  \caption{Loss terms of Eq.~\eqref{eq:total_loss} and the regions they constrain. (a)~$\mathcal{L}_{\text{rgb}}$: a floater (dashed) and a surface-aligned density profile (solid) render the same pixel color. (b)~$\mathcal{L}_{\text{depth}}$: the target $q_r(d)$ (rose) pulls the rendered depth distribution $p_r(d)$ (blue) onto the measured range $d_r^*$. (c)~$\mathcal{L}_{\text{n}}$ and $\mathcal{L}_{\text{smooth}}$: between the LiDAR anchors, a rippling iso-surface (blue, dashed) fits the images as well as a smooth one, and the regularizers smooth it (rose).}
  \vspace{-10pt}
  \label{fig:loss_coverage}
\end{figure}

We train the sparse voxel radiance field using the standard rendering losses and render-normal consistency regularization~\cite{svraster}, together with LiDAR depth supervision, occupied voxel guidance, and structural smoothness regularization.
The map-training objective is
\begin{equation}
  \mathcal{L} =
  \mathcal{L}_{\text{rgb}}
  + \lamnormal \mathcal{L}_{\text{n}}
  + \lamdepth \mathcal{L}_{\text{depth}}
  + \lampocc \mathcal{L}_{\text{occ}}
  + \lamsmooth \mathcal{L}_{\text{smooth}} ~,
  \label{eq:total_loss}
\end{equation}
where $\mathcal{L}_{\text{rgb}}$, $\mathcal{L}_{\text{n}}$, $\mathcal{L}_{\text{depth}}$, $\mathcal{L}_{\text{occ}}$, and $\mathcal{L}_{\text{smooth}}$ denote color reconstruction, normal consistency, LiDAR depth supervision, occupied voxel guidance, and density-field smoothness, respectively.
They constrain the field in complementary regions, as summarized in Fig.~\ref{fig:loss_coverage}.
Color supervision fits every camera ray but under-constrains geometry, since different density profiles can render the same pixel color (Fig.~\ref{fig:loss_coverage}~(a)).
LiDAR depth supervision anchors the field metrically, but only along measured rays (Fig.~\ref{fig:loss_coverage}~(b)).
Between the anchors, where no range measurement is available, the normal and smoothness regularizers keep the surface locally smooth (Fig.~\ref{fig:loss_coverage}~(c)).
Occupied voxel guidance suppresses floaters that depth supervision leaves off the measured rays (Fig.~\ref{fig:point_occ_guidance}).

\subsubsection{Photometric Supervision}
\label{ssec:training_objective}

For each training camera, SpaRV synthesizes an RGB image $\pixrgb$ from the current sparse-voxel field using the camera pose, and compares it with the observed color $\pixrgb^*$. The color loss combines a pixel-wise reconstruction term with SSIM:
\begin{equation}
  \mathcal{L}_{\text{rgb}} =
  (1 - \lamssim) \sum_{\text{pixels}}\| \pixrgb - \pixrgb^* \|_2^2
  + \lamssim \mathcal{L}_{\text{SSIM}} ~,
  \label{eq:color_loss}
\end{equation}
where $\lamssim$ controls the SSIM contribution.

\subsubsection{LiDAR Depth Supervision}
\label{sec:geometric_consistency}
\label{ssec:depth_aided}

Photometry constrains depth only through the parallax between views, and a short baseline gives little parallax.
Where the captured trajectory observes a region from few and uneven viewpoints, the color loss alone leaves its geometry under-constrained.
SpaRV therefore uses LiDAR range measurements as metric anchors along their rays.
Because these rays are sparse and irregular, SpaRV evaluates them with the event-driven ray marcher of Sec.~\ref{sec:raymarch} rather than a rasterization-only camera renderer, as illustrated in Fig.~\ref{fig:raster_vs_raymarch}.
The rendered depth along a ray is:
\begin{equation}
  D = \sum\nolimits_{i=1}^{N} T_i\,\alpha_i\,d_i ~,
  \label{eq:rendered_depth}
\end{equation}
where $d_i$ is the depth of the $i$-th sample along the ray.

LiDAR depth supervision treats the volume-rendering weights $w_i=T_i\alpha_i$ as a discrete depth distribution and minimizes a KL-divergence loss~\cite{deng2022depth} to a unimodal target distribution centered at the measured LiDAR range:
\begin{equation}
  \begin{aligned}
  \mathcal{L}_{\text{depth}}
  &=
  \sum_{r \in \mathcal{R}_{\text{LiDAR}}}
  D_{\mathrm{KL}}\!\left(q_r(d) \,\|\, p_r(d)\right) \\
  &\equiv
  -\sum_{r \in \mathcal{R}_{\text{LiDAR}}}
  \sum_i q_r(d_i)\log(w_i) ~,
  \end{aligned}
  \label{eq:depth_loss}
\end{equation}
where $p_r(d_i)=w_i$ is the rendered depth distribution along LiDAR ray $r$, and $q_r(d_i)\propto\Delta t_i\exp[-(d_i-d_r^*)^2/(2{\voxsiz}_i^2)]$ is the target distribution centered at the observed range $d_r^*$, with $\Delta t_i$ the sample spacing and ${\voxsiz}_i$ the size of the voxel containing sample $i$. The target therefore sharpens as the map refines.

\subsubsection{Structural Regularization}
\label{ssec:neighbor_reg}

Depth supervision constrains the field only along measured LiDAR rays. Between those anchors, the density can still ripple or drift while reproducing the training images, as shown in Fig.~\ref{fig:loss_coverage}~(c).
The normal-consistency loss $\mathcal{L}_{\text{n}}$~\cite{huang20242d} penalizes the inconsistency between the rendered density-gradient normal and the normals estimated from the rendered depth maps, including both the expected-depth normal $\pixnormalmean$ and the median-depth normal $\pixnormalmed$:
\begin{equation}
  \mathcal{L}_{\text{n}} =
  \left\| \pixnormalsigma - \pixnormalmean \right\|_2^2
  +
  \left\| \pixnormalsigma - \pixnormalmed \right\|_2^2 ~,
  \label{eq:normal_loss}
\end{equation}
where $\pixnormalsigma$ is the rendered density-gradient normal.

To further smooth the local geometry, we apply a geometry-aware smoothness regularization $\mathcal{L}_{\text{smooth}}$ across face-adjacent voxels. For a neighboring voxel pair $(\alpha,\beta)$, it penalizes the difference between their local density-gradient normals:
\begin{equation}
  \mathcal{L}_{\text{smooth}}
  =
  \sum_{(\alpha,\beta)\in\mathcal{N}}
  \omega_{\alpha\beta}
  \left\|
  {\nabla}\sigma_{\alpha}
  -
  {\nabla}\sigma_{\beta}
  \right\|_2^2 ~,
  \label{eq:neighbor_smooth_loss}
\end{equation}
where $\mathcal{N}$ denotes the set of face-neighbor voxel pairs, ${\nabla}\sigma_{\alpha}$ and ${\nabla}\sigma_{\beta}$ denote the normalized density gradients of the two neighboring voxels, and $\omega_{\alpha\beta}$ is a geometry-aware edge weight determined by the shared face area and the distance between voxel centers.

\subsubsection{Occupied Voxel Guidance}
\label{ssec:point_occ_guidance}

\begin{figure}[t]
  \centering
  \includegraphics[width=\linewidth]{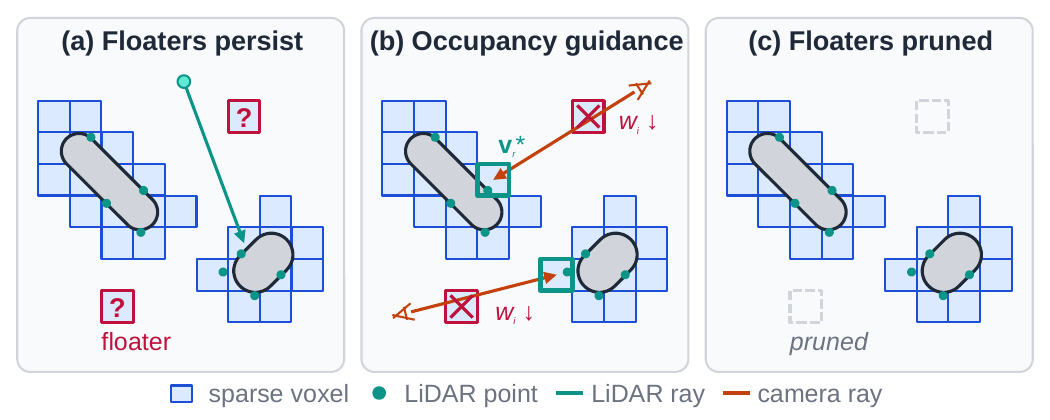}
  \vspace{-20pt}
  \caption{Occupied voxel guidance on the sparse voxel map. Teal dots represent input LiDAR points. (a)~LiDAR depth supervision leaves floaters off the measured rays unconstrained. (b)~Guiding camera rays to terminate in the first occupied voxel suppresses floaters in front of it. (c)~Voxel adaptation removes floater voxels whose blending weights fall below the pruning threshold.}
  \vspace{-5pt}
  \label{fig:point_occ_guidance}
\end{figure}

A LiDAR scan is sparse, and its field of view does not coincide with the camera's.
LiDAR depth supervision therefore acts only along measured rays, and free-space floaters off those rays cannot be supervised by Eq.~\eqref{eq:depth_loss} (Fig.~\ref{fig:point_occ_guidance}~(a)).
The point cloud still marks which voxels contain measured surface.
Occupied voxel guidance converts this occupancy into a termination target for camera rays that intersect an occupied voxel.

\begin{figure}[t]
  \centering
  \setlength{\subfigcapskip}{3pt}
  \setlength{\subfigcapmargin}{0pt}
  \subfigure[w/o $\mathcal{L}_{\text{occ}}$]{%
    \begin{tabular}{@{}c@{}}
      \includegraphics[width=0.485\columnwidth]{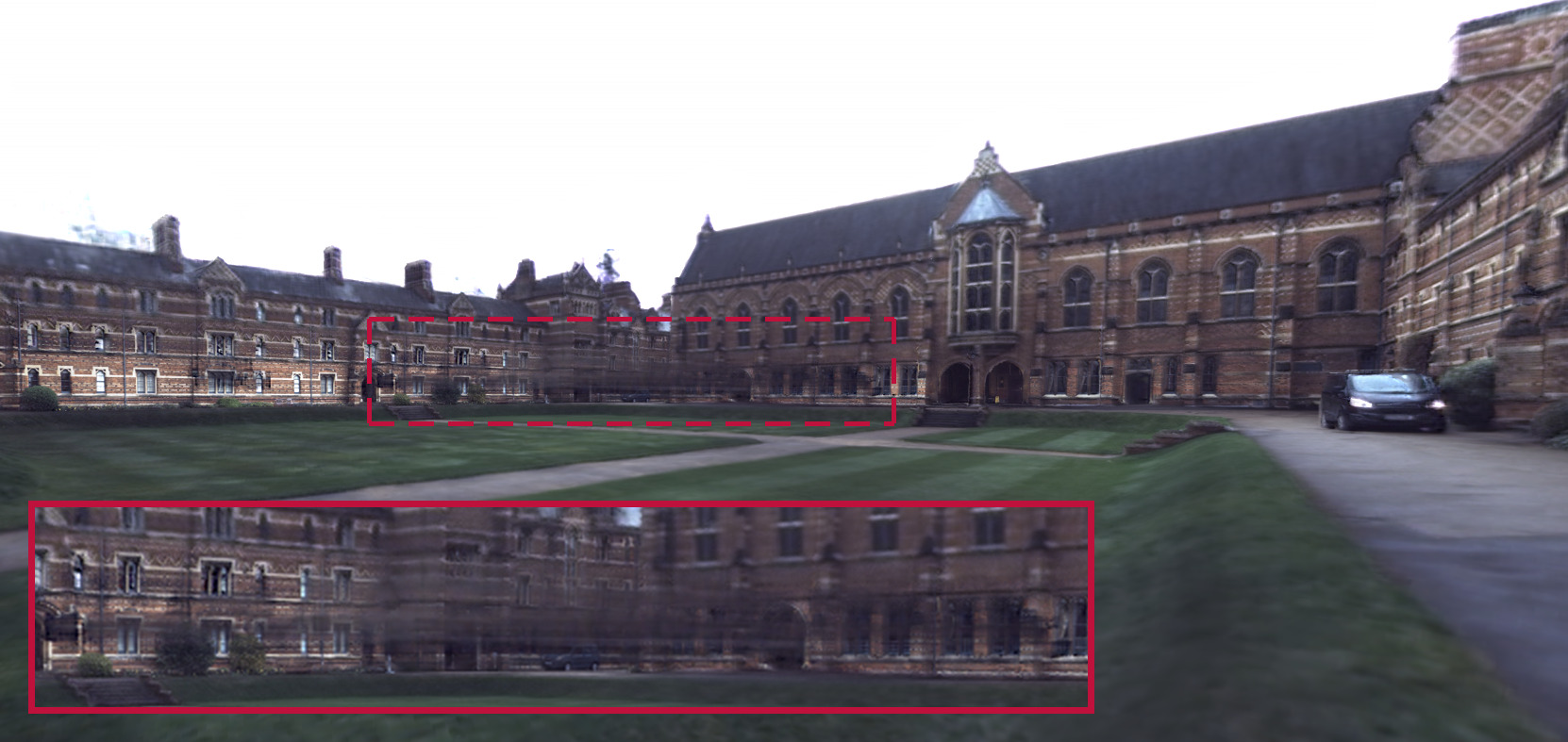}\\[-2pt]
      \includegraphics[width=0.485\columnwidth]{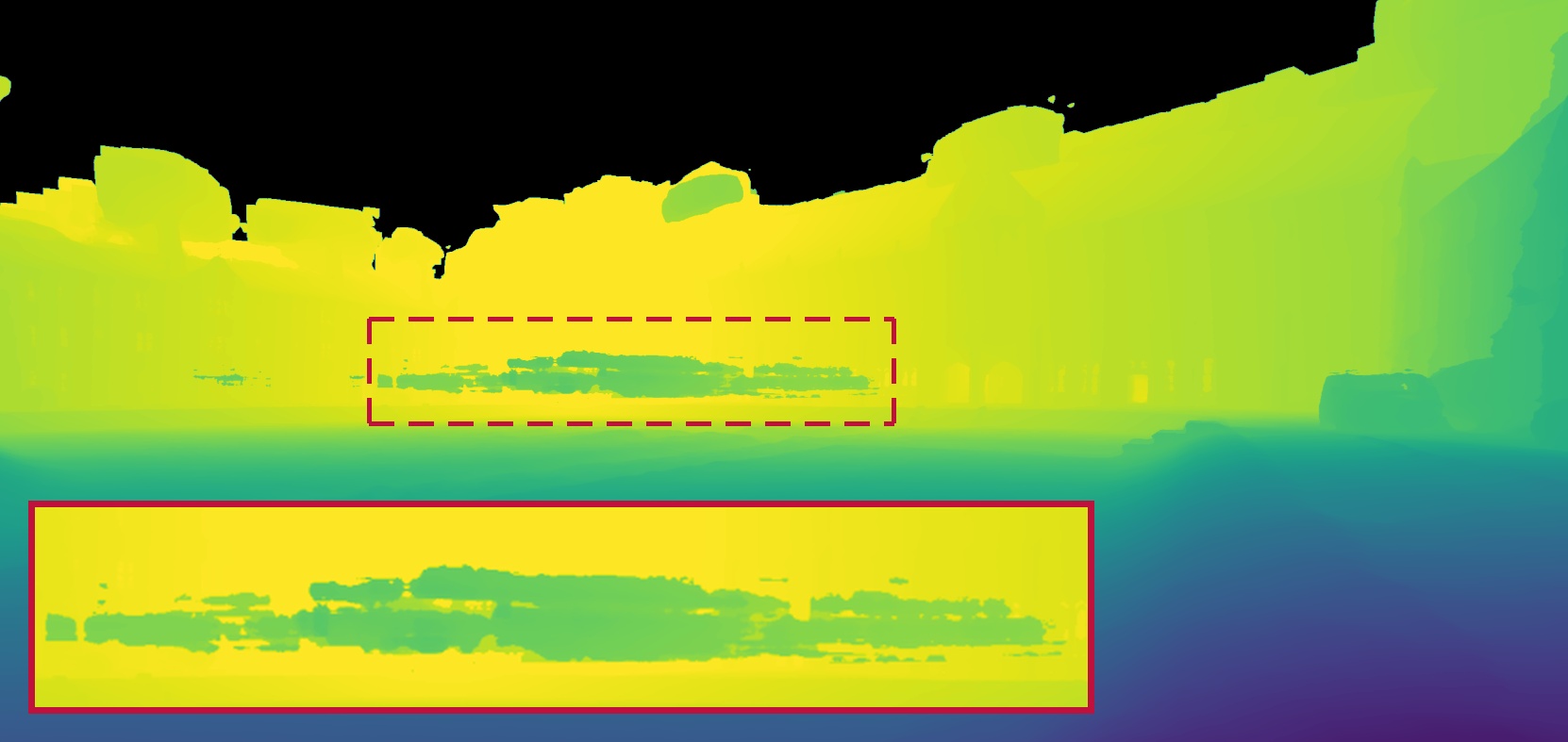}\\[-2pt]
      \includegraphics[width=0.485\columnwidth]{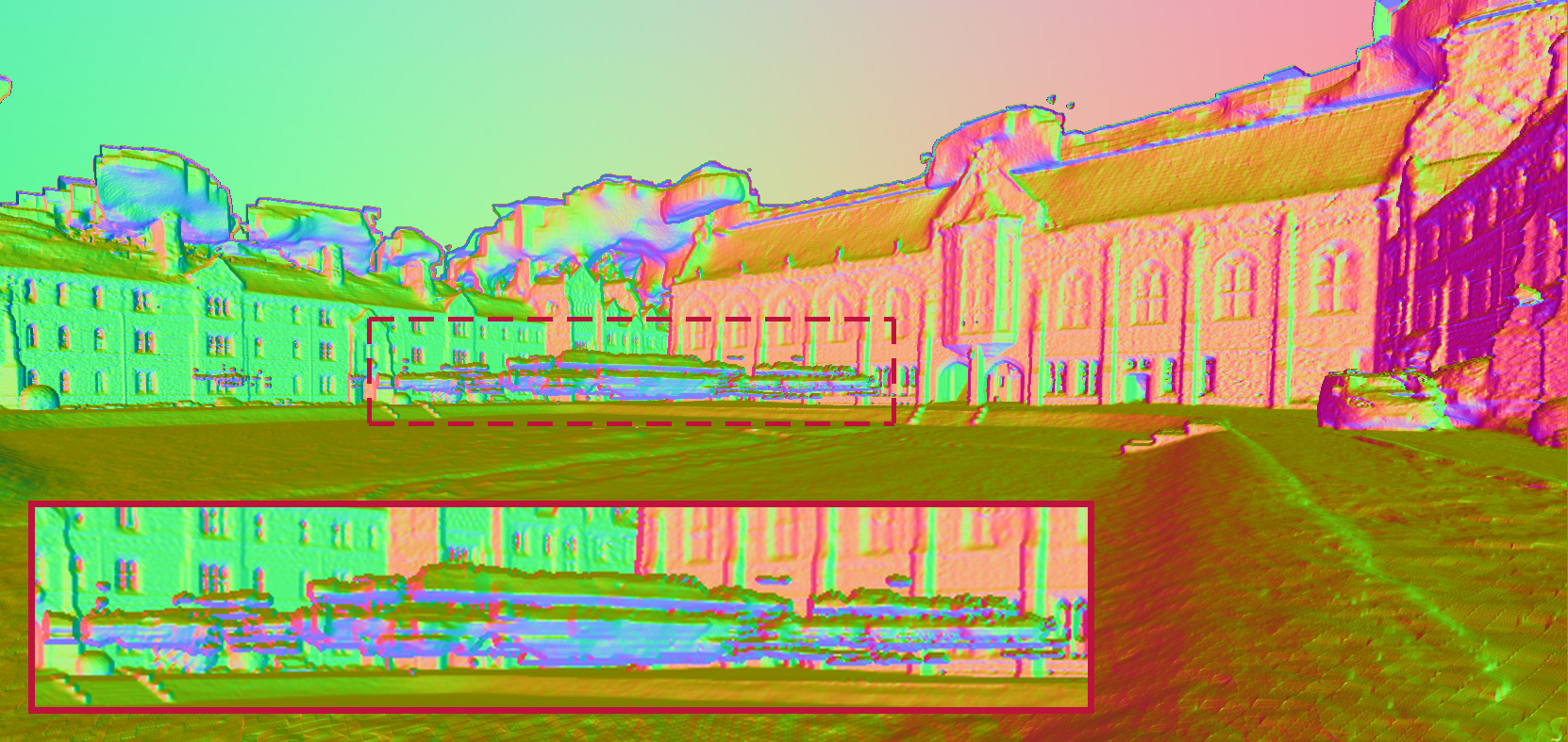}
    \end{tabular}%
  }%
  \hspace{3pt}%
  \subfigure[w/ $\mathcal{L}_{\text{occ}}$]{%
    \begin{tabular}{@{}c@{}}
      \includegraphics[width=0.485\columnwidth]{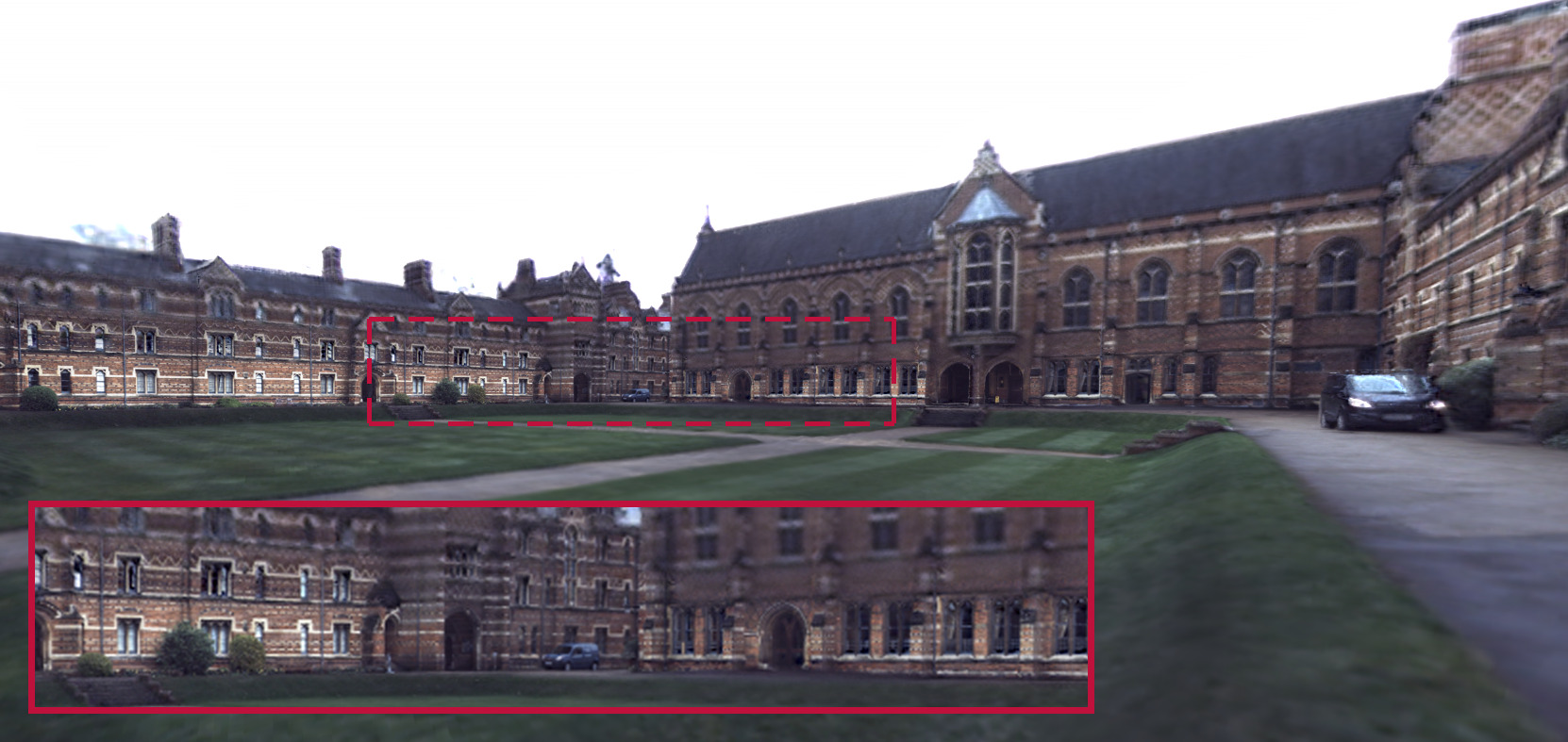}\\[-2pt]
      \includegraphics[width=0.485\columnwidth]{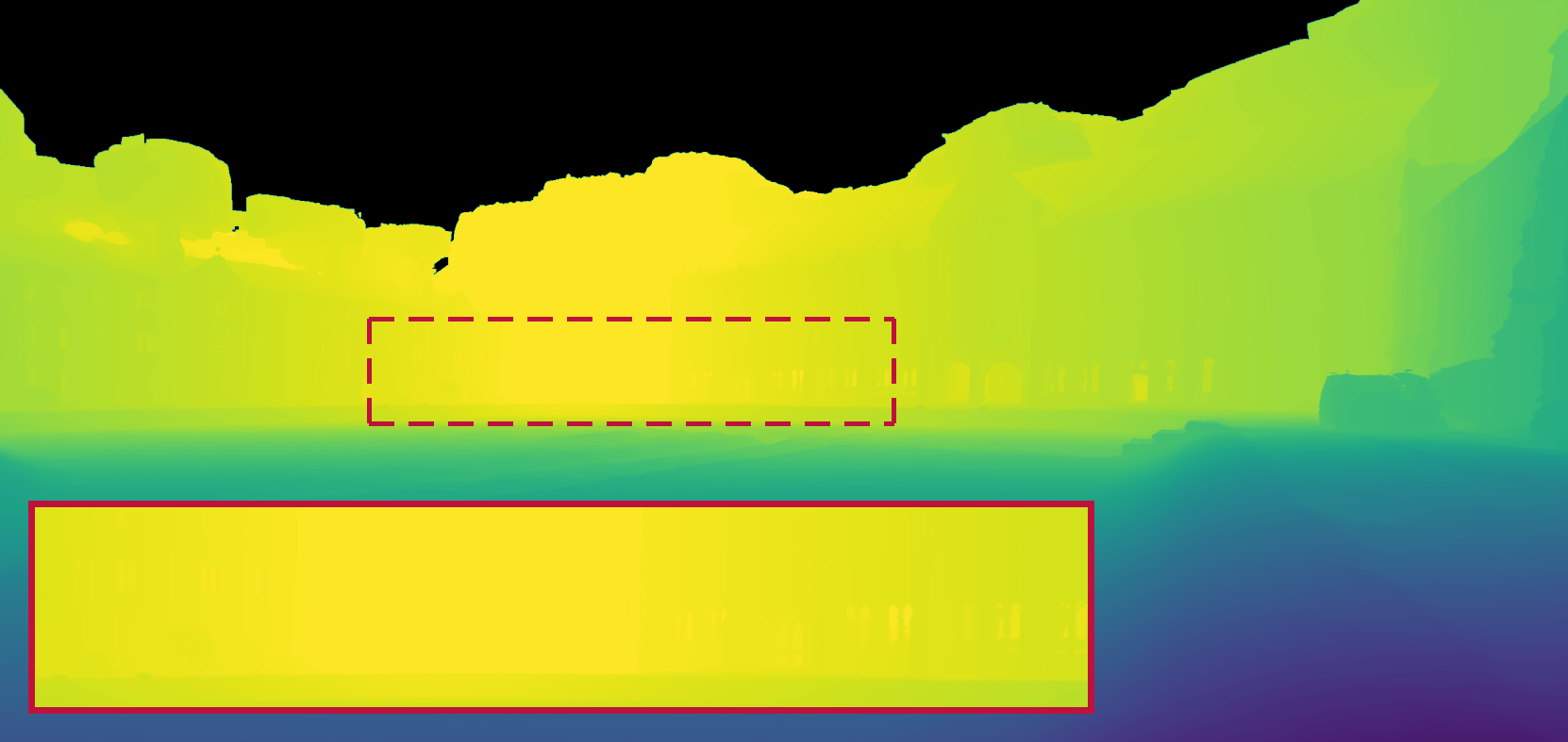}\\[-2pt]
      \includegraphics[width=0.485\columnwidth]{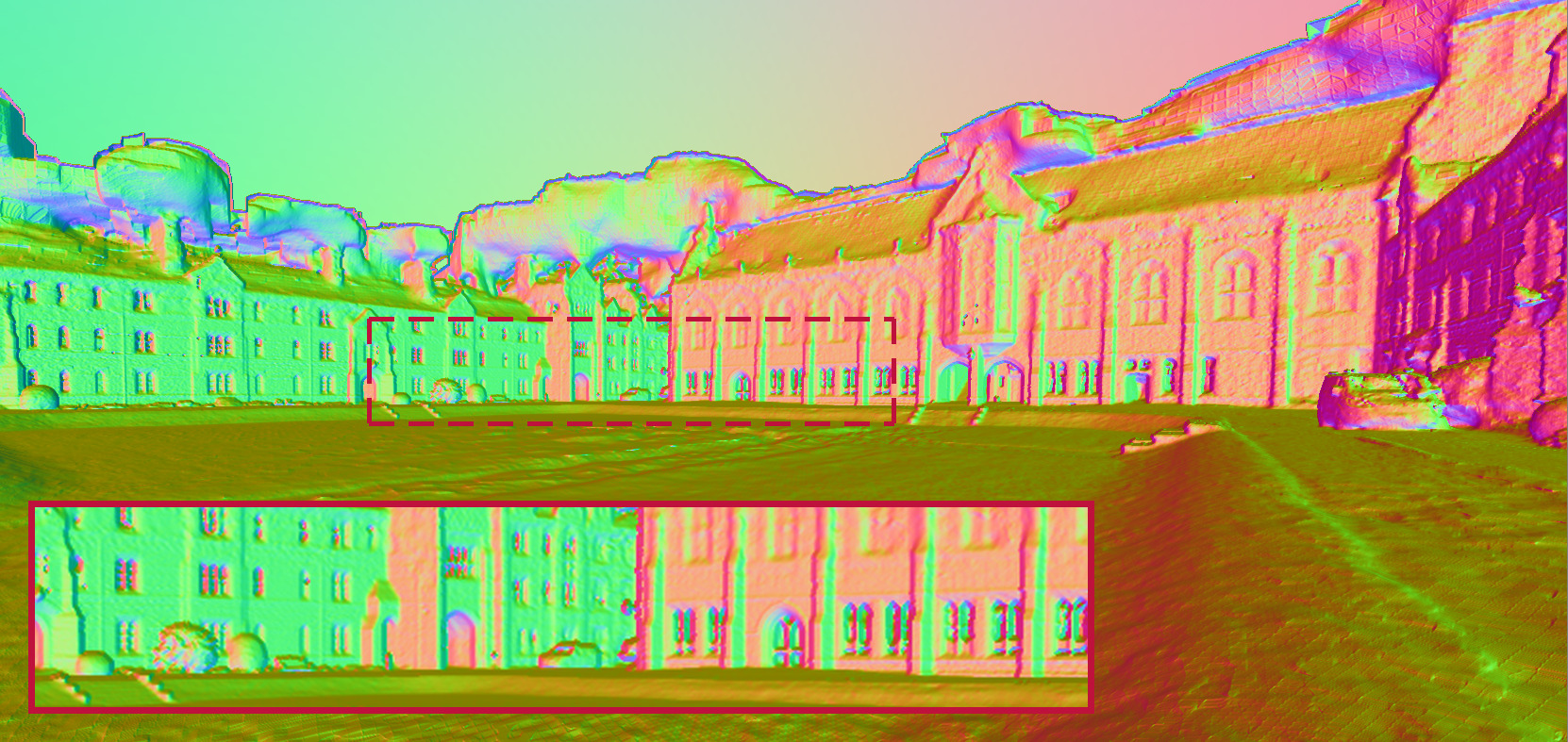}
    \end{tabular}%
  }%
  \caption{Effect of the occupied voxel guidance on an Oxford Spires scene: the same view from a map trained (a)~without and (b)~with $\mathcal{L}_{\text{occ}}$ of Eq.~\eqref{eq:point_occ_loss}, as color, median depth, and the normal map derived from that depth.}
  \vspace{-15pt}
  \label{fig:occ_guidance_ablation}
\end{figure}

\begin{figure*}[t]
  \centering
  \includegraphics[width=\textwidth]{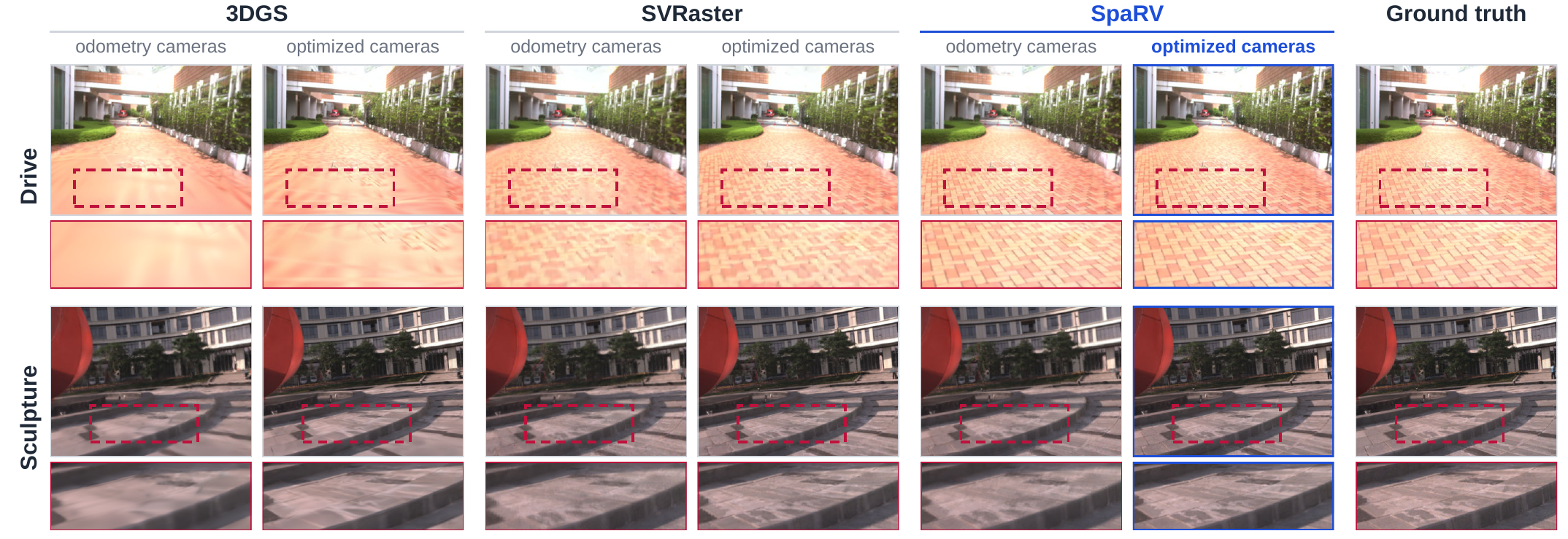}
  \vspace{-20pt}
  \caption{Rendering results with the odometry and the optimized cameras (Sec.~\ref{sec:pose_refinement}) on training views of FAST-LIVO2 Drive (top) and Sculpture (bottom).}
  \label{fig:fast_livo_cam_opt_qual}
  \vspace{-10pt}
\end{figure*}

Let $\mathcal{V}_{\mathrm{occ}}$ denote the set of current sparse voxels that contain at least one input LiDAR point. 
For each camera ray that intersects this set, let $\vox_r^\star$ denote the first occupied voxel along the ray.
The guidance encourages the ray to terminate inside $\vox_r^\star$ by maximizing the sum of its blending weights $w_i=T_i\alpha_i$:
\begin{equation}
  \mathcal{L}_{\text{occ}}
  =
  -\frac{1}{|\mathcal{R}_{\mathrm{occ}}|}
  \sum_{r\in\mathcal{R}_{\mathrm{occ}}}
  \log\left(\sum_{i\in\vox_r^\star} w_i+\epsilon\right) ~,
  \label{eq:point_occ_loss}
\end{equation}
where $\mathcal{R}_{\mathrm{occ}}$ is the set of camera rays that intersect an occupied voxel and $\epsilon$ is a small constant for numerical stability.

As illustrated in Fig.~\ref{fig:point_occ_guidance}~(b), the guidance increases the accumulated blending weight in the first occupied voxel.
It encourages ray termination near the LiDAR-observed surface and suppresses the weights of floaters in front of that surface.
As the rendered surface converges toward the observed surface, floater voxels whose maximum blending weight across training views falls below the pruning threshold are removed by voxel adaptation (Sec.~\ref{ssec:voxel_init_adapt}; Fig.~\ref{fig:point_occ_guidance}~(c)).
Selecting only the \emph{first} occupied voxel focuses the guidance on the foreground surface along each ray.
Occupied voxels behind it are not additional termination targets and can remain occluded when photometric supervision has already recovered the foreground surface.
The guidance applies only to occupied voxels that a camera ray actually reaches.
When photometric supervision recovers an opaque surface in front of an occupied voxel, the ray terminates before that voxel (Alg.~\ref{alg:raymarch}), and the guidance on that ray lapses.
LiDAR occupancy and the images thus compete for where each ray terminates, and a surface recovered from the images survives as long as it explains them.
The guidance is also applied only in the early stage of training, while the geometry is still forming.
Fig.~\ref{fig:occ_guidance_ablation} shows the resulting difference on an Oxford Spires scene: trained without the term, the map reproduces the view in color yet leaves floaters in regions observed only by cameras; with it, the same view renders a clean structure.

\section{Camera Optimization}
\label{sec:pose_refinement}

This section optimizes the camera poses and focal lengths against the map.
The initial camera parameters are metrically plausible but carry residual errors, which blur the textures of the map.
The map under training predicts the image each camera should see, and the photometric residual against the captured image exposes these errors.
The trilinear density and the vertex-interpolated DC appearance of Sec.~\ref{sec:sparse_voxel_grid} give this residual dense gradients with respect to the camera parameters.
SpaRV alternates this camera optimization with map training to sharpen the textures of the map.

\vspace{-10pt}
\subsection{Camera Parameterization}

Camera poses are parameterized as rigid transformations $\mathbf{T}_c^w \in SE(3)$ from camera to world coordinates:
\begin{equation}
  \mathbf{T}_c^w =
  \begin{bmatrix} \mathbf{R} & \mathbf{t} \\ \mathbf{0} & 1
  \end{bmatrix} ~,
  \label{eq:c2w}
\end{equation}
where $\mathbf{R} \in SO(3)$ is the rotation matrix and $\mathbf{t} \in \Real^3$ is the translation vector. Since the initial poses are already metrically plausible, SpaRV optimizes an error-state correction $\boldsymbol{\xi} = [\boldsymbol{\omega}, \delta\mathbf{t}] \in \Real^6$ in the tangent space of $SE(3)$, where $\boldsymbol{\omega} \in so(3)$ is the rotation perturbation and $\delta\mathbf{t} \in \Real^3$ is the translation perturbation~\cite{eade2013lie,GSSLAM2024}. The optimized pose is written as
\begin{equation}
  \mathbf{T}_c^{w} = \exp(\boldsymbol{\xi}^{\wedge})\, \mathbf{T}_{c}^{w,(0)} ~,
  \label{eq:pose_update}
\end{equation}
where $\mathbf{T}_{c}^{w,(0)}$ denotes the odometry initialization, which stays fixed across rounds.
Intrinsic optimization is illustrated with a shared pinhole model for each intrinsic group; other camera models replace only the pixel-to-ray map:
\begin{equation}
  \mathbf{K} =
  \begin{bmatrix}
    f_x & 0 & c_x \\
    0 & f_y & c_y \\
    0 & 0 & 1
  \end{bmatrix} ~.
  \label{eq:pinhole_ray}
\end{equation}
Within each group, the principal point $(c_x,c_y)$ remains fixed at its initialized value, while the focal lengths are represented by the vector $\mathbf{f}=[f_x,f_y]^\top$ and optimized jointly with the camera poses. For numerical stability, SpaRV optimizes a log-space focal error state $\boldsymbol{\phi}_f \in \Real^2$ relative to the initialization $\mathbf{f}^{(0)}$ such that
\begin{equation}
  \log \mathbf{f} = \log \mathbf{f}^{(0)} + \boldsymbol{\phi}_f ~.
  \label{eq:focal_update}
\end{equation}
For a pinhole pixel $(u,v)$, the camera-space ray direction is $\ray_{\text{cam}} \propto \mathbf{K}^{-1}[u, v, 1]^\top$. The loss is backpropagated through the rendered rays to both the pose and focal error states.

\vspace{-10pt}
\subsection{Photometric Camera Optimization}

Each camera-optimization step updates the camera poses and focal lengths by minimizing a direct photometric objective between the rendered image and the observed RGB image through differentiable rendering. Given the rendered and observed images $\pixrgb$ and $\pixrgb^*$, the camera-optimization objective is defined as
\begin{equation}
  \mathcal{L}_{\mathrm{cam}} =
  \mathcal{L}_{\text{rgb}}\left(\pixrgb, \pixrgb^*\right)
  + \lambda_{\xi}\, \rho_{\xi}(\boldsymbol{\xi})
  + \lambda_{f}\,\rho_{f}\!\left(\boldsymbol{\phi}_f\right) ~,
  \label{eq:camera_opt_loss}
\end{equation}
where $\mathcal{L}_{\text{rgb}}(\cdot,\cdot)$ denotes the color loss defined in Eq.~\eqref{eq:color_loss}. Robust kernels regularize both the pose correction and the shared focal update around their initialized values~\cite{huber1992robust}.
The vertex-interpolated DC SH component of Sec.~\ref{sec:sparse_voxel_grid} varies continuously within each voxel and across faces shared by voxels, so the photometric gradient of this objective does not vanish inside the voxels.

\begin{figure*}[!t]
  \centering
  \includegraphics[width=\linewidth]{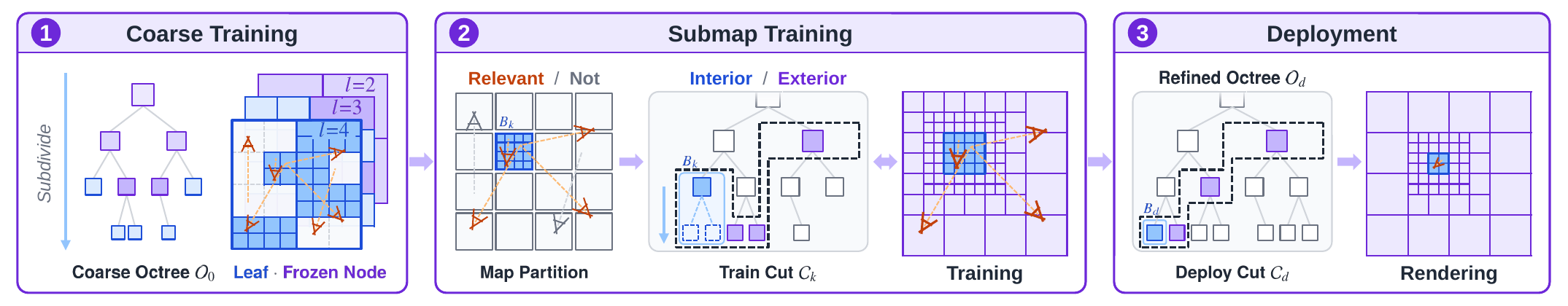}
  \vspace{-20pt}
  \caption{Large-scale training and deployment of SpaRV.
  (1)~Coarse training (Sec.~\ref{ssec:coarse_training}) subdivides the octree over the whole scene up to the memory budget and keeps each coarser level as frozen nodes.
  (2)~Submap training (Sec.~\ref{ssec:submap_train}) trains one submap $B_k$ at a time, with its relevant views, on the train cut $\mathcal{C}(\tau_t,B_k,\mathcal{O}_0)$: the interior (blue) is subdivided (dashed leaves), the exterior (violet) stays frozen, and grey nodes are not in the cut.
  (3)~Deployment (Sec.~\ref{ssec:deploy_render}) renders from the deploy cut $\mathcal{C}(\tau_d,B_d,\mathcal{O}_d)$ of the refined octree around the deploy box $B_d$ that contains the camera.}
  \vspace{-10pt}
  \label{fig:lod_pipeline}
\end{figure*}

\subsection{Alternating Optimization Schedule}

SpaRV alternates between updating the map parameters and updating the camera parameters. During map updates, the camera parameters are held fixed and the voxel field is trained with Eq.~\eqref{eq:total_loss}.
Camera optimization runs on the same fixed period as voxel adaptation.
At each camera-optimization round, SpaRV freezes the map and collects the cameras that will be sampled within the next voxel-adaptation interval.
SpaRV then optimizes this look-ahead set for a small number of steps.
The updated camera parameters are rolled back if they do not reduce the color loss $\mathcal{L}_{\text{rgb}}(\cdot,\cdot)$.
The optimized camera parameters are then reused in the subsequent map-training iterations.
As shown in Fig.~\ref{fig:fast_livo_cam_opt_qual}, alternating optimization improves rendering quality as the camera parameters are updated.

\section{Large-Scale Reconstruction}
\label{sec:large_scale}

The GPU memory required by an explicit voxel map grows with both the extent of the scene and the requested level of detail.
Consequently, the map of a large scene fits within a fixed GPU memory budget only at a coarser level of detail.
A voxel's projected size decreases with its distance from the camera, so a view needs fine voxels only near the camera.
SpaRV therefore keeps the full octree in host memory and places only a selection of its voxels, called a granularity-driven \emph{cut}, on the GPU.
The cut selects voxels by their projected size: fine voxels inside a box of interest, and progressively coarser voxels farther away.
Throughout this section, $N_{\max}$ denotes the maximum number of voxels resident on the GPU at one time.
During training, voxel subdivision stops when the number of resident voxels reaches this budget.
During deployment, the cut is loaded subject to the same budget.
GPU memory then stays bounded as the scene grows.

Fig.~\ref{fig:lod_pipeline} summarizes the pipeline.
Coarse training first builds the coarse map of the whole scene and initializes the octree.
Submap training then visits the disjoint submaps one at a time and refines each within its own cut.
The box of each cut bounds the submap in focus, so the cut holds that submap at the finest available level and the rest of the scene at coarser levels.
Deployment applies the same cut rule to the refined octree, with the box placed around the moving camera.
We first describe the octree and the cut, then detail their roles in training and rendering.

  \vspace{-7pt}
\subsection{Level-of-Detail Data Structure}
\label{ssec:lod_hierarchy}

\subsubsection{Structure and Growth}
\label{ssec:lod_growth}

\begin{figure}[h]
  \centering
  \setlength{\subfigcapskip}{-3pt}
  \subfigure[Level 10]{
    \includegraphics[width=0.23\columnwidth]{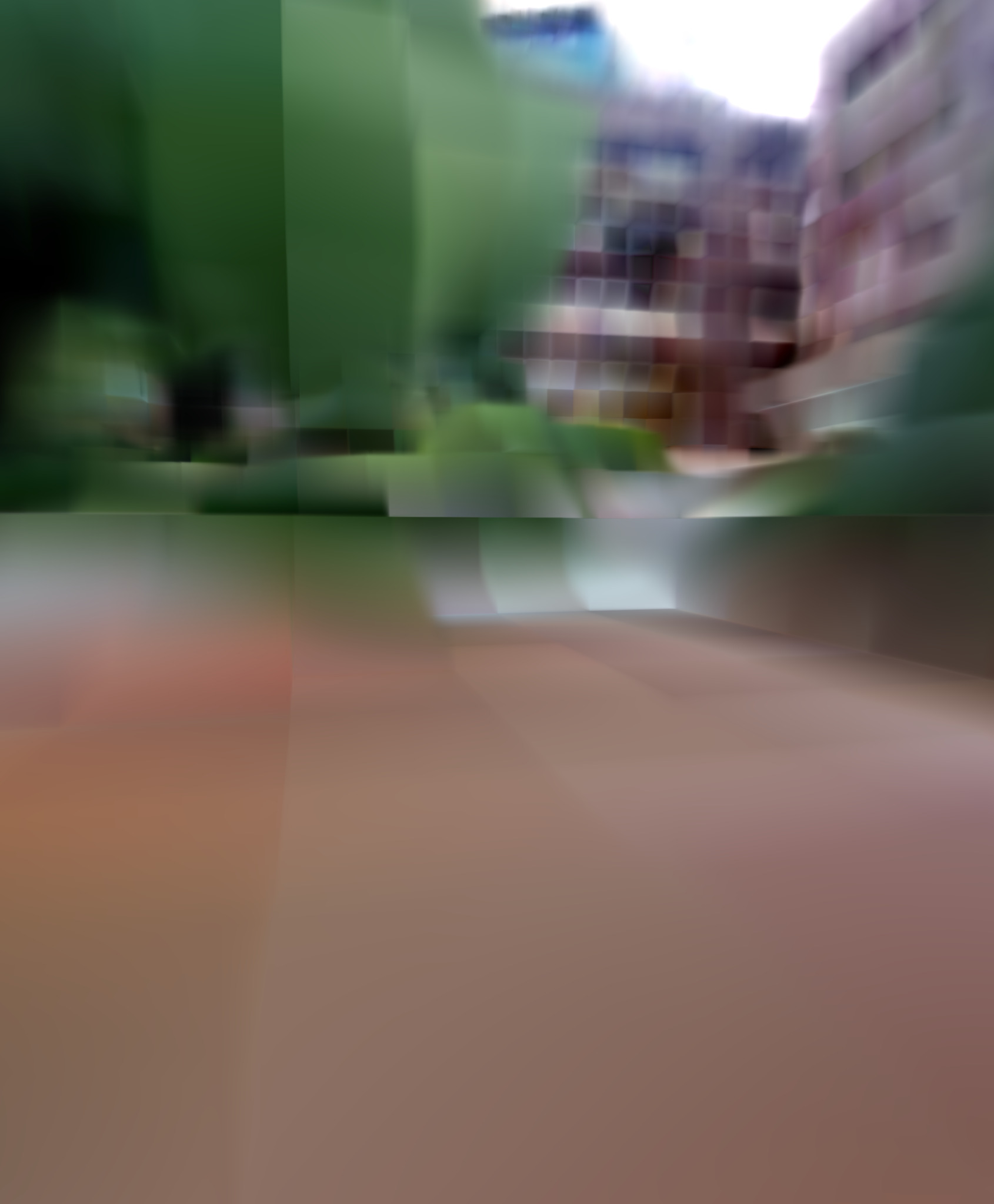}
  }%
  \hspace{-3pt}%
  \subfigure[Level 11]{
    \includegraphics[width=0.23\columnwidth]{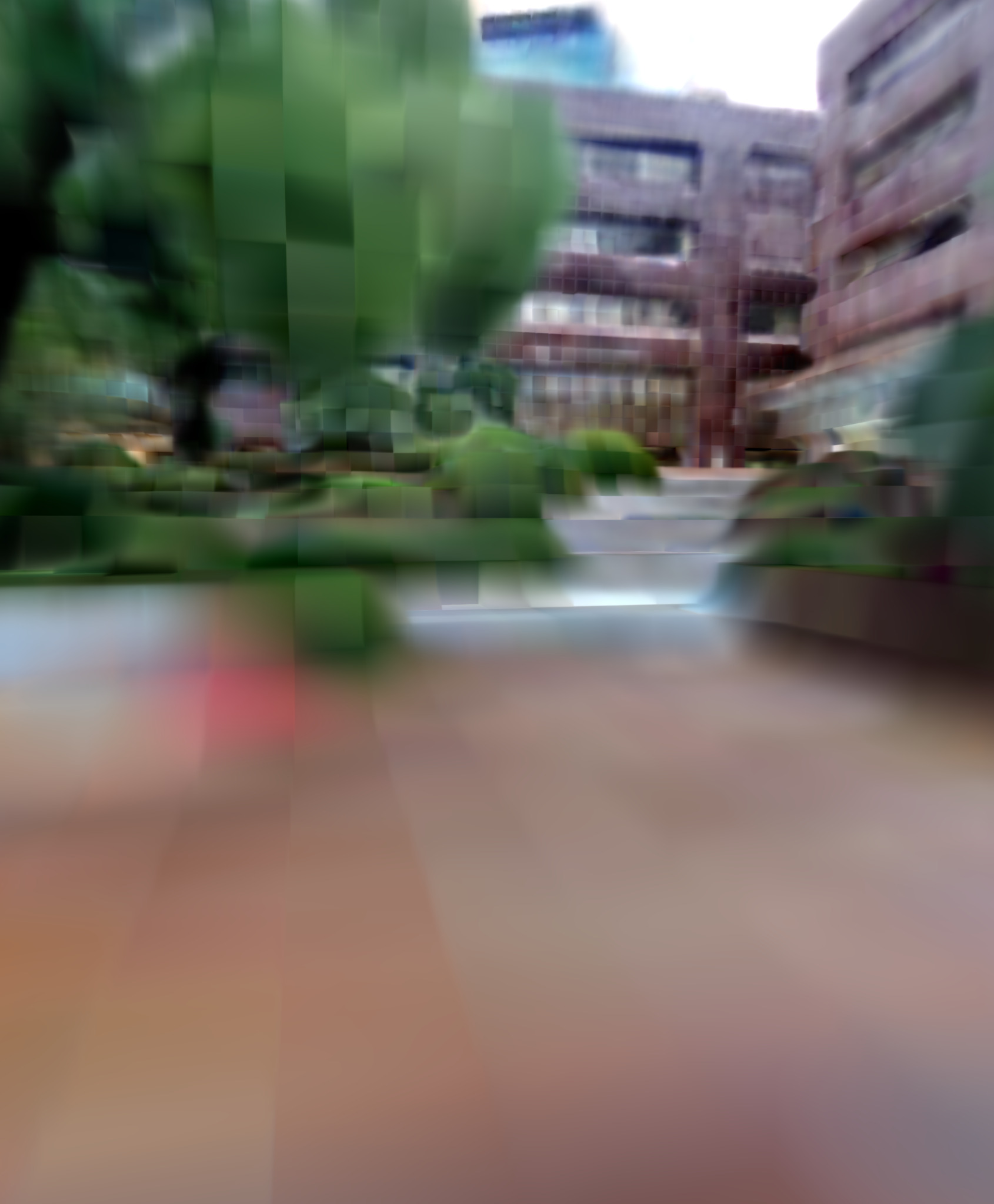}
  }%
  \hspace{-3pt}%
  \subfigure[Level 12]{
    \includegraphics[width=0.23\columnwidth]{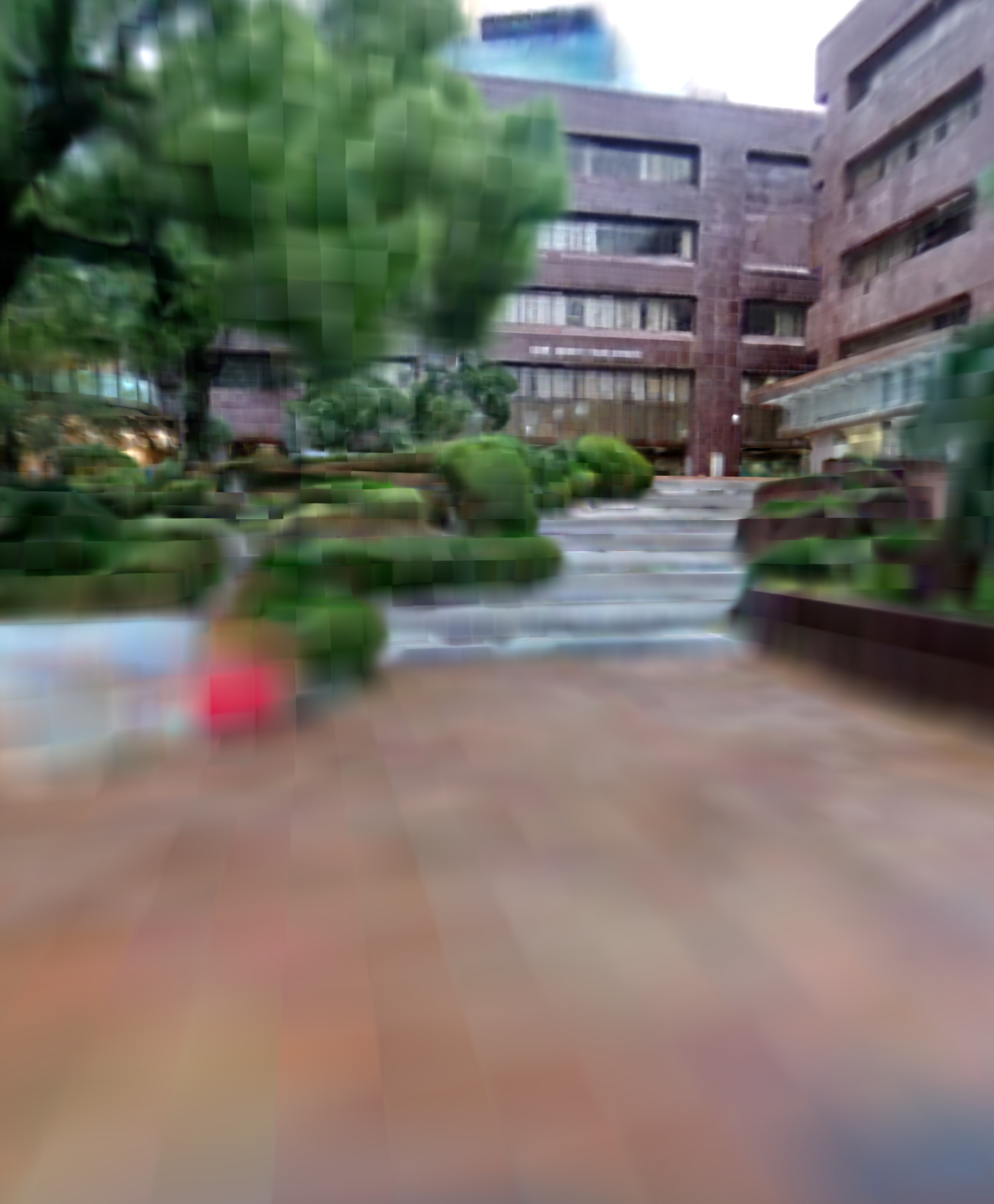}
  }%
  \hspace{-3pt}%
  \subfigure[Level 16]{
    \includegraphics[width=0.23\columnwidth]{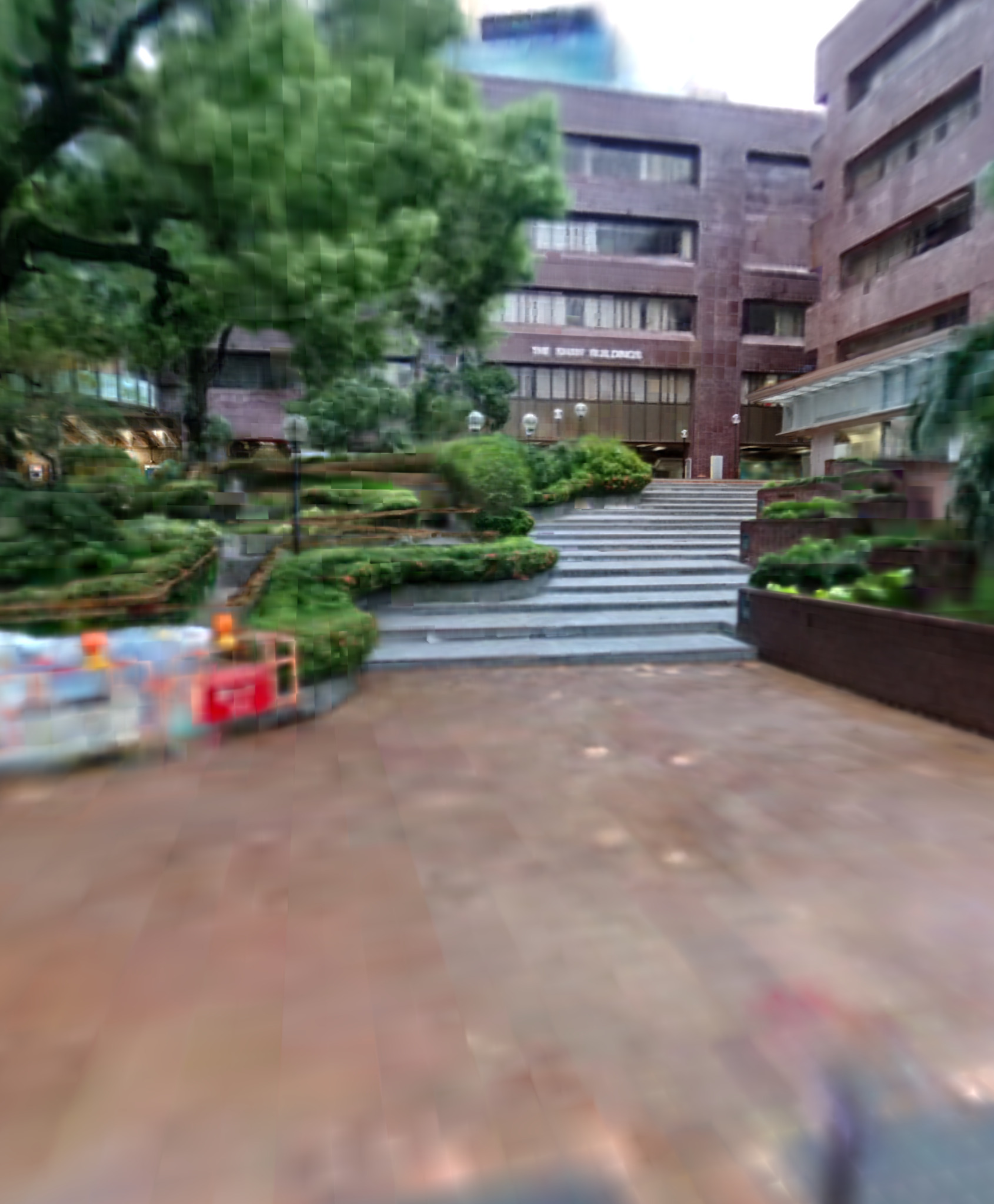}
  }%

  \vspace{-3pt}

  \subfigure[$\tau=1$]{
    \includegraphics[width=0.23\columnwidth]{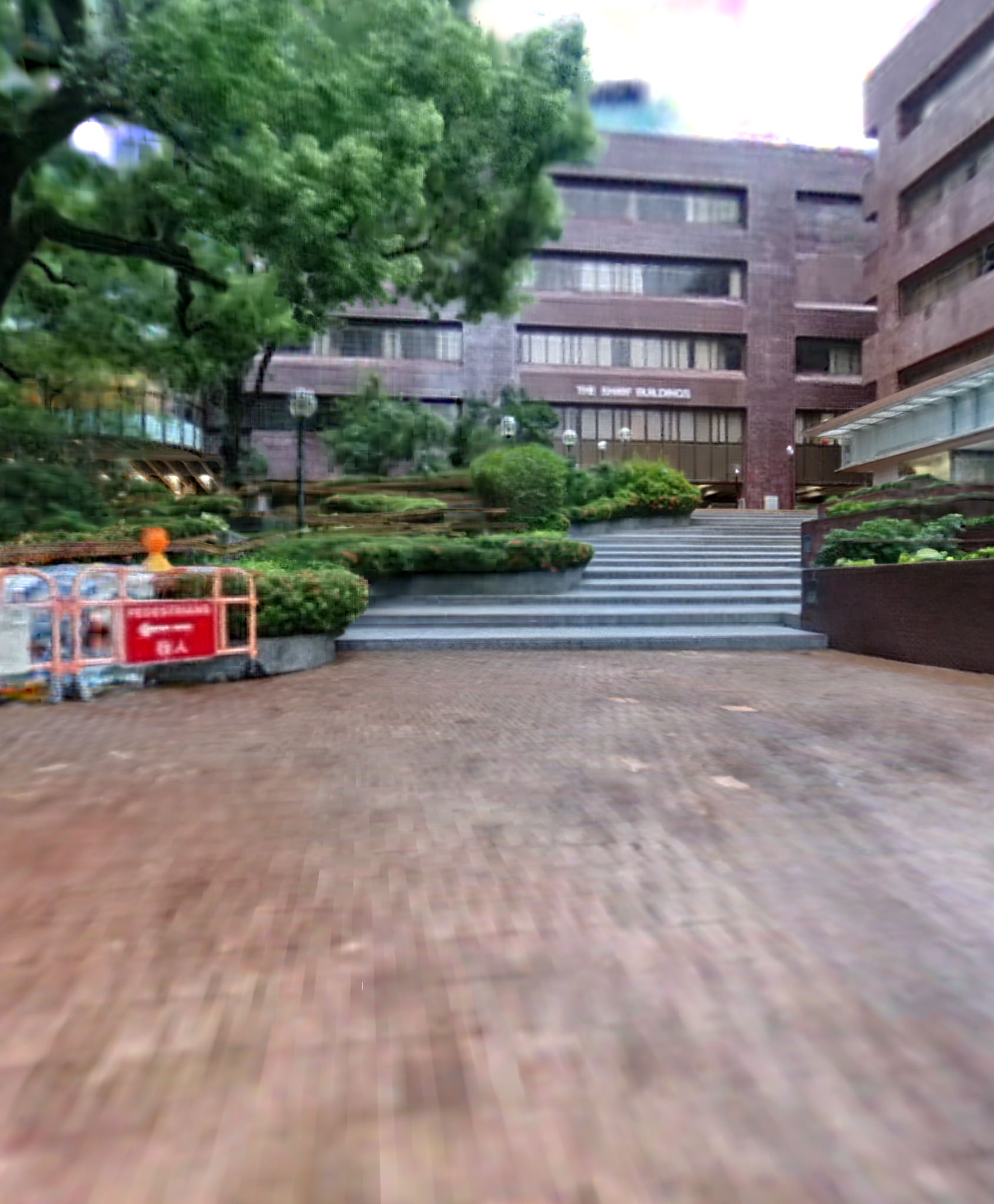}
  }%
  \hspace{-3pt}%
  \subfigure[$\tau=5$]{
    \includegraphics[width=0.23\columnwidth]{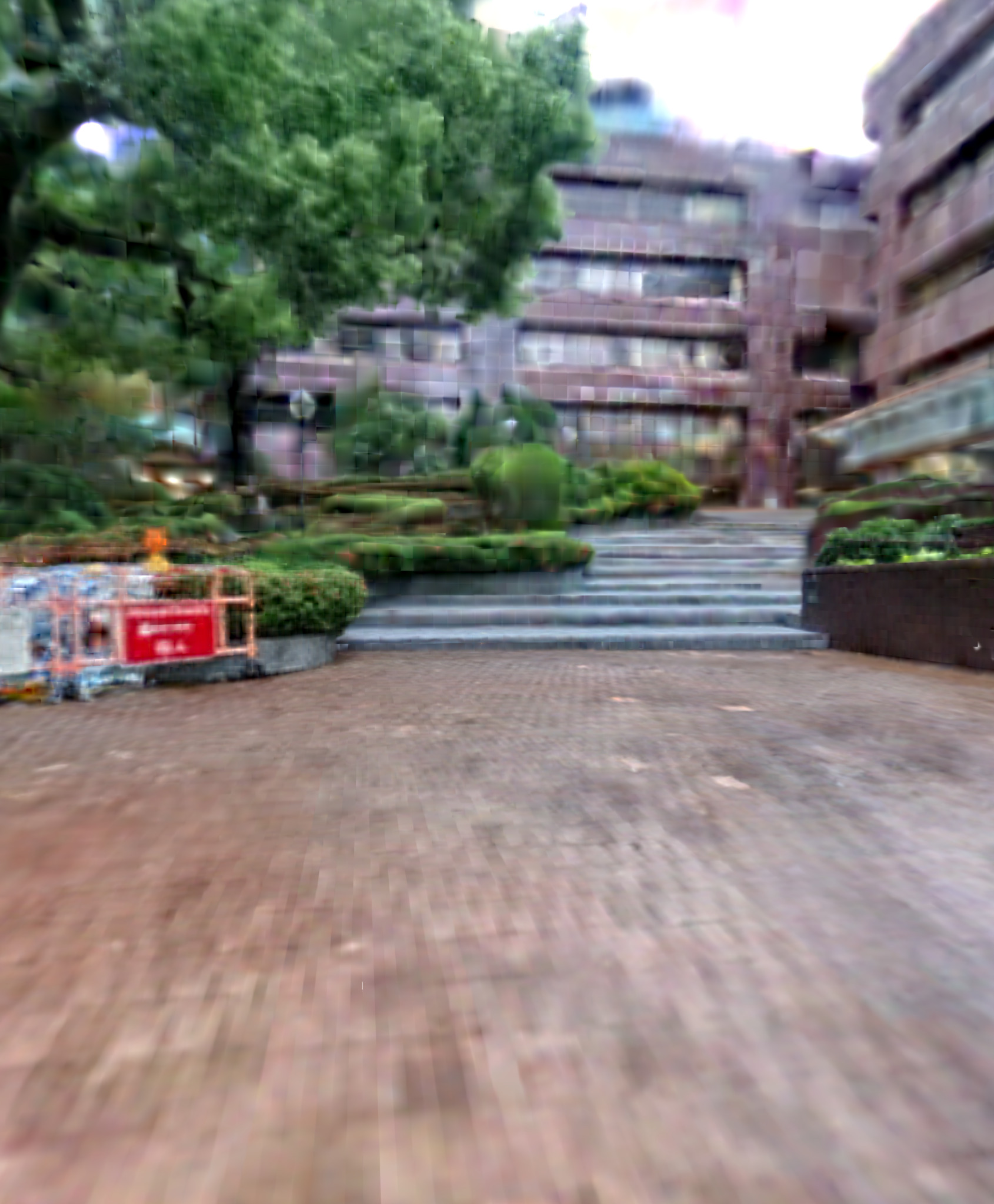}
  }%
  \hspace{-3pt}%
  \subfigure[$\tau=9$]{
    \includegraphics[width=0.23\columnwidth]{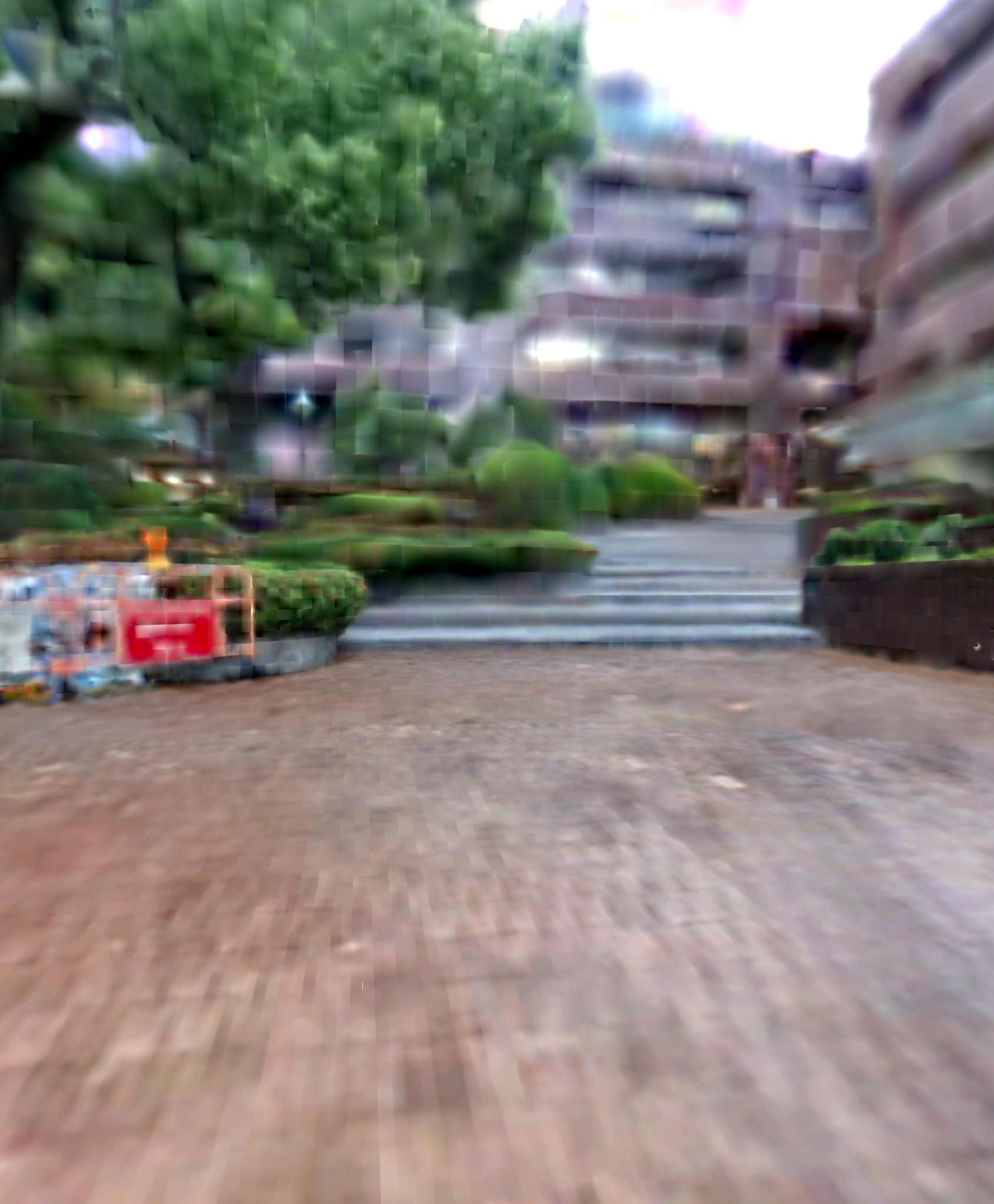}
  }%
  \hspace{-3pt}%
  \subfigure[$\tau=17$]{
    \includegraphics[width=0.23\columnwidth]{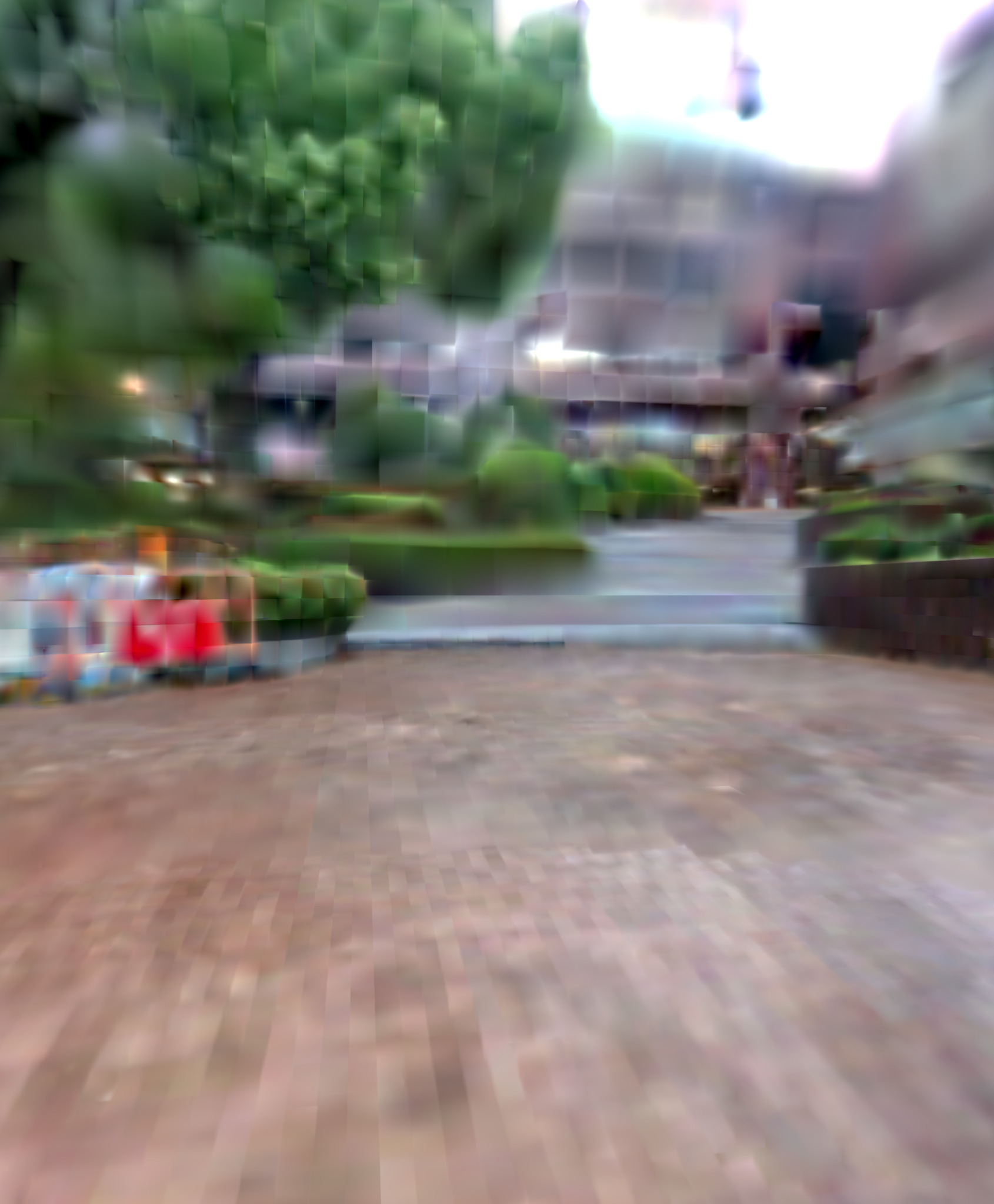}
  }%
  \caption{Levels of detail of one octree, seen from a single viewpoint. \emph{(a)--(d)}~Increasingly fine levels. \emph{(e)--(h)}~Cuts at increasing $\tau$.}
  \label{fig:lod_levels}
\end{figure}

The octree $\mathcal{O}$ stores both the current leaves and the coarser internal nodes created during subdivision (Fig.~\ref{fig:lod_pipeline}~(1)), so every level that training passes through stays available to the cut.
It uses the flat $(\text{octpath}, \text{octlevel})$ indexing of Sec.~\ref{sec:sparse_voxel_grid}.
When a voxel subdivides according to Sec.~\ref{ssec:voxel_init_adapt}, its trained parameters are frozen and retained as the parent node, while its eight children become trainable leaves.
The parent is not trained further because a voxel is split only after it has been sufficiently trained at its current level.
Its parameters therefore remain a valid approximation at that coarser level.
Training continues on the children, which add finer detail while the parent stays a snapshot of the preceding level.
Fig.~\ref{fig:lod_levels}~(a)--(d) illustrates the detail represented at different levels of the same octree.

\begin{figure}[!t]
  \centering
  \includegraphics[width=\linewidth]{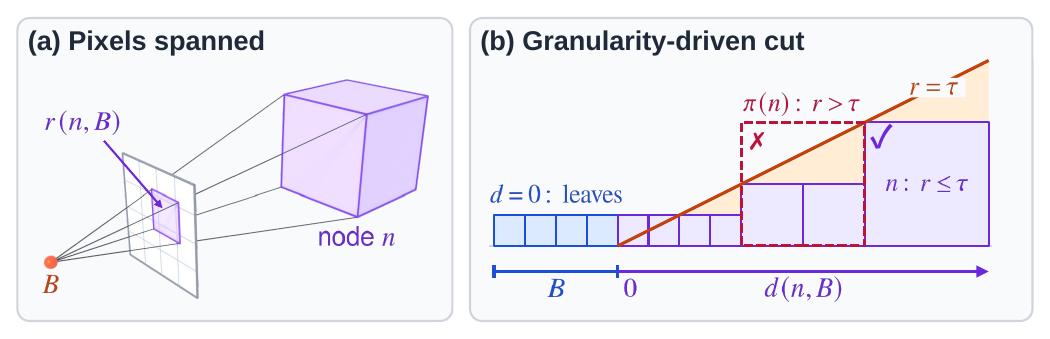}
  \vspace{-20pt}
  \caption{Granularity-driven cut selection. (a)~With the box $B$ collapsed to the camera's focal point, $r(n,B)$ of Eq.~\eqref{eq:size_ratio} is the number of pixels node $n$ spans on the image plane. (b)~The cut of Eq.~\eqref{eq:cut_predicate} against the distance $d(n,B)$: on each root-to-leaf path, it takes the coarsest node that projects to at most $\tau$ pixels. Here the parent $\pi(n)$ projects larger than $\tau$ and is rejected (dashed, crossed), and its child $n$ is taken. Inside $B$, where $d(n,B)=0$, the cut keeps the leaves.}
  \vspace{-10pt}
  \label{fig:lod_cut}
\end{figure}

\subsubsection{Granularity-Driven Cut}
\label{ssec:cut_selection}

A \emph{cut} selects voxels from different levels of the octree to cover the whole scene: full detail inside a box $B$, and coarser voxels outside it.
It is defined by the box $B$ and by a granularity target $\tau$ in pixels~\cite{kerbl2024hierarchical}.
The cut should select as few voxels as possible, while no selected voxel projects to more than $\tau$ pixels in any camera inside $B$ unless the octree has no finer voxel there.
To measure how large a voxel node $n$ appears from $B$, we use the projected-size proxy
\begin{equation}
  r(n, B) = f\,\frac{\voxsiz(n)}{d(n, B)} ~,
  \label{eq:size_ratio}
\end{equation}
where $f=\max(f_x,f_y)$ is the focal length of a pinhole camera, $\voxsiz(n)$ is the voxel size, and $d(n,B)$ is the minimum Euclidean distance between voxel $n$ and box $B$.
When $B$ collapses to the camera's focal point, $r(n,B)$ is the number of pixels the voxel spans on the image plane (Fig.~\ref{fig:lod_cut}~(a)).
For a larger box, $d(n,B)$ never exceeds the distance from $n$ to any camera inside $B$, so $r(n,B)$ bounds the projected size of $n$ in every camera inside $B$.
Coarser nodes lie closer to the root, so along each root-to-leaf path the first node with $r(n,B) \le \tau$ is the coarsest node that is fine enough.
The cut takes that node, or the leaf when no node on the path is fine enough:
\begin{equation}
  \begin{aligned}
    \mathcal{C}(\tau,B,\mathcal{O})
    = \big\{n \in \mathcal{O}:\;&\big[r(\pi(n),B) > \tau\big] \\
    &{}\land \big[r(n,B) \le \tau \;\lor\; \text{$n$ is a leaf}\big]\big\} ,
  \end{aligned}
  \label{eq:cut_predicate}
\end{equation}
where $\pi(n)$ denotes the parent of $n$, with $r(\pi(n),B)=\infty$ for the root.
The cut thus takes exactly one node along each root-to-leaf path.
Since a parent covers the same space as its children together, the selected voxels cover the scene without overlaps or holes and form a sparse radiance voxel map that the renderer uses directly.
Inside $B$, $d(n,B)=0$ and $r(n,B)=\infty$, so the cut descends to the leaves and keeps the finest available voxels, while outside $B$ the selected voxels grow coarser with distance (Fig.~\ref{fig:lod_cut}~(b)).
Increasing $\tau$ reduces the number of selected voxels at the cost of coarser detail (Fig.~\ref{fig:lod_levels}~(e)--(h)), whereas enlarging $B$ keeps a larger region at the finest available level and increases the number of selected voxels.

  \vspace{-10pt}
\subsection{Coarse-to-Fine Training}
\label{ssec:hier_opt}

\subsubsection{Coarse Training}
\label{ssec:coarse_training}

Coarse training applies the training of Sec.~\ref{sec:methodology} to the whole scene under the budget $N_{\max}$: voxels subdivide to capture detail until the leaf count reaches $N_{\max}$.
This yields the coarse map covering the whole scene, with optimized cameras and the initial octree $\mathcal{O}_0$ (Fig.~\ref{fig:lod_pipeline}~(1)).

\subsubsection{Submap Training}
\label{ssec:submap_train}

As shown in Fig.~\ref{fig:lod_pipeline}~(2), submap training first partitions the scene into non-overlapping axis-aligned boxes.
The target submap size $s_t$ determines the octree level $l_t=\left\lceil\log_2(\worldsiz/s_t)\right\rceil$.
This level defines the box $B_k$ of each submap, with edge length $2^{-l_t}\,\worldsiz$.
The $k$-th submap owns the subtrees of $\mathcal{O}_0$ inside $B_k$, and submaps refine disjoint portions of the octree.
Each submap's training views are selected with the coarse map: a view is assigned to a submap if its rays reach voxels inside $B_k$.

Each submap is initialized from its cut $\mathcal{C}_k=\mathcal{C}(\tau_t,B_k,\mathcal{O}_0)$ with granularity target $\tau_t$.
The cut is partitioned into an \emph{interior} $\mathcal{I}_k=\{n\in\mathcal{C}_k : n\subseteq B_k\}$, the leaves inside $B_k$, and an \emph{exterior} $\mathcal{E}_k=\mathcal{C}_k\setminus\mathcal{I}_k$, the remaining voxels of the cut.
Only the interior is trained, following Sec.~\ref{sec:methodology}: subdivision adds finer leaves inside $B_k$ and appends their trained parents to the octree (Sec.~\ref{ssec:lod_growth}).
The exterior stays frozen but participates in rendering, so the training views are rendered with the surrounding scene in place and the interior does not have to explain content outside $B_k$.
Both parts count against the budget $N_{\max}$.
After all submaps are trained in turn, the octree already holds their refined subtrees and serves directly as the octree $\mathcal{O}_d$ for deployment.

  \vspace{-10pt}
\subsection{Deployment}
\label{ssec:deploy_render}

Deployment applies the same cut rule around the camera, so rendering preserves local detail without making all fine-resolution voxels of the entire scene resident on the GPU at once.
It selects the cut from the refined octree $\mathcal{O}_d$, using a granularity target $\tau_d$ and a \emph{deploy box} $B_d$ of target size $s_d$:
$\mathcal{C}_d=\mathcal{C}(\tau_d,B_d,\mathcal{O}_d)$.
The deploy box contains the camera and aligns with the octree partition at the level determined by $s_d$, following the same sizing rule as in training (Fig.~\ref{fig:lod_pipeline}~(3)).
The deploy box is smaller than a training box ($s_d<s_t$), so the finest voxels cover a smaller region and more of the memory budget remains for the exterior.
The granularity target $\tau_d$ sets how coarse the exterior is and is increased when the cut exceeds $N_{\max}$.
The cut is recomputed only when the camera moves beyond the current deploy box by a small margin, preventing repeated switching near the box boundary.
The resident voxels are then updated by incremental host-device transfer.
Training and rendering thus share one octree and one cut rule, and both keep the resident voxels within $N_{\max}$ as the scene grows.

\section{Experiments}
\label{sec:experiments}

This section tests the four contributions of the paper on five datasets.
Radiance-field methods fit their training views closely, so differences in geometry appear mainly at viewpoints away from the captured trajectory.
To evaluate geometric consistency, we compare the methods at interpolated viewpoints along the trajectory and at extrapolated viewpoints away from it, on the synthetic Replica dataset (Sec.~\ref{ssec:replica}) and the real FAST-LIVO2 dataset (Sec.~\ref{ssec:fast_livo2}).
Replica provides ground-truth poses, so its extrapolated views are evaluated against ground truth, while those of FAST-LIVO2 are compared qualitatively.
Since the cameras of FAST-LIVO2 come from odometry and carry residual errors, we also use this dataset to evaluate camera optimization (Sec.~\ref{sssec:camera_refinement}).
We further compare the methods with accurate cameras on the Oxford Spires benchmark (Sec.~\ref{ssec:oxford}), and ablate occupied voxel guidance on it.
To evaluate large-scale reconstruction under bounded GPU memory, we use two kilometer-scale scenes that we captured, Garden and Campus (Secs.~\ref{ssec:large_scale_recon}--\ref{ssec:runtime}).
To evaluate the ray-marching renderer beyond pinhole cameras, we train SpaRV on the EgoNeRF dataset (Sec.~\ref{ssec:egonerf}), where an omnidirectional camera supervises the map through the same renderer.
Finally, Sec.~\ref{ssec:applications} uses the trained maps for surface extraction and, through the same renderer, for the simulation of other camera and LiDAR models.

\vspace{-10pt}
\subsection{Experimental Setup}

\subsubsection{Baselines and Metrics}

We compare SpaRV against the following methods:
\textbf{InstantNGP}~\cite{muller2022instant}: hash-grid NeRF;
\textbf{3DGS}~\cite{kerbl20233d}: image-only 3D Gaussian splatting;
\textbf{SVRaster}~\cite{svraster}: sparse-voxel radiance field whose representation SpaRV shares;
\textbf{LetsGo}~\cite{cui2024letsgo}: 3DGS initialized from a LiDAR Poisson mesh and regularized by LiDAR depth;
\textbf{GS-SDF}~\cite{liu2025gs}: Gaussians coupled to a LiDAR-built neural SDF;
\textbf{EgoNeRF}~\cite{choi2023balanced}: NeRF on a balanced spherical grid for egocentric omnidirectional views (Sec.~\ref{ssec:egonerf});
\textbf{OmniGS}~\cite{li2025omnigs}: omnidirectional 3DGS (Sec.~\ref{ssec:egonerf});
\textbf{H3DGS}~\cite{kerbl2024hierarchical}: hierarchical 3DGS with divide-and-conquer training and LoD rendering (Sec.~\ref{ssec:large_scale_recon}).
We report PSNR, SSIM, and LPIPS~\cite{zhang2018unreasonable}. PSNR measures per-pixel fidelity and tolerates blur, whereas SSIM and LPIPS respond to lost structure and perceptual detail.

\begin{table*}[!t]
  \centering
  \caption{Quantitative rendering results on the Replica dataset. Per-scene values are at extrapolated viewpoints (E); the last two columns average the extrapolated and the interpolated (I) viewpoints over the 8 scenes.}
  \label{tab:replica_rendering}
  \resizebox{1.0\textwidth}{!}{%
    \begin{tabular}{cccccccccccc}
      \toprule
      \multirow{2}*{\textbf{Metrics}} & \multirow{2}*{\textbf{Methods}} & \multicolumn{9}{c}{\textbf{Extrapolated (E)}} & \textbf{Interp. (I)} \\
      \cmidrule(lr){3-11}\cmidrule(lr){12-12}
       & & \textbf{Office-0} & \textbf{Office-1} & \textbf{Office-2} & \textbf{Office-3} & \textbf{Office-4} & \textbf{Room-0} & \textbf{Room-1} & \textbf{Room-2} & \textbf{Avg.} & \textbf{Avg.} \\
      \midrule
      \multirow{6}*{PSNR$\uparrow$} & InstantNGP & 39.863 & \textbf{40.473} & \textbf{35.799} & \textbf{35.928} & 34.217 & 32.019 & 31.761 & 35.080 & 35.643 & 33.713 \\
       & 3DGS & 36.047 & 36.315 & 25.438 & 23.639 & 28.532 & 25.944 & 27.098 & 26.703 & 28.714 & 40.439 \\
       & SVRaster & 37.914 & 36.460 & 31.267 & 30.544 & 32.781 & 31.881 & 33.291 & 32.636 & 33.347 & 39.600 \\
       & LetsGo & 30.263 & 30.808 & 23.130 & 21.762 & 26.225 & 21.949 & 24.512 & 21.756 & 25.051 & 38.892 \\
       & GS-SDF & \underline{40.523} & 39.488 & 33.214 & 33.719 & \underline{37.017} & \underline{33.833} & \underline{35.015} & \textbf{35.615} & \underline{36.053} & \underline{40.548} \\
       & \textbf{SpaRV} & \textbf{40.568} & \underline{39.496} & \underline{35.326} & \underline{35.089} & \textbf{37.572} & \textbf{34.382} & \textbf{35.716} & \underline{35.374} & \textbf{36.690} & \textbf{41.001} \\
      \midrule
      \multirow{6}*{SSIM$\uparrow$} & InstantNGP & 0.967 & \underline{0.965} & 0.951 & 0.949 & 0.951 & 0.903 & 0.919 & 0.945 & 0.944 & 0.933 \\
       & 3DGS & 0.963 & 0.947 & 0.914 & 0.892 & 0.936 & 0.886 & 0.916 & 0.919 & 0.922 & 0.974 \\
       & SVRaster & 0.968 & 0.936 & 0.946 & 0.943 & \underline{0.953} & \underline{0.936} & 0.942 & 0.947 & 0.946 & \underline{0.975} \\
       & LetsGo & 0.924 & 0.895 & 0.892 & 0.890 & 0.913 & 0.851 & 0.892 & 0.831 & 0.886 & 0.965 \\
       & GS-SDF & \underline{0.976} & 0.960 & \underline{0.961} & \underline{0.958} & \textbf{0.969} & \textbf{0.948} & \textbf{0.957} & \textbf{0.963} & \underline{0.962} & 0.973 \\
       & \textbf{SpaRV} & \textbf{0.977} & \textbf{0.970} & \textbf{0.963} & \textbf{0.961} & \textbf{0.969} & \textbf{0.948} & \underline{0.956} & \underline{0.959} & \textbf{0.963} & \textbf{0.982} \\
      \midrule
      \multirow{6}*{LPIPS$\downarrow$} & InstantNGP & 0.095 & \underline{0.128} & 0.149 & 0.147 & 0.159 & 0.199 & 0.178 & 0.149 & 0.150 & 0.154 \\
       & 3DGS & 0.097 & 0.154 & 0.181 & 0.187 & 0.142 & 0.196 & 0.168 & 0.160 & 0.161 & 0.095 \\
       & SVRaster & 0.091 & 0.160 & 0.157 & 0.145 & 0.121 & 0.141 & 0.133 & 0.134 & 0.135 & 0.090 \\
       & LetsGo & 0.134 & 0.186 & 0.224 & 0.219 & 0.174 & 0.247 & 0.192 & 0.210 & 0.198 & 0.122 \\
       & GS-SDF & \underline{0.060} & 0.151 & \underline{0.115} & \underline{0.105} & \underline{0.085} & \underline{0.123} & \underline{0.091} & \underline{0.093} & \underline{0.103} & \underline{0.078} \\
       & \textbf{SpaRV} & \textbf{0.045} & \textbf{0.076} & \textbf{0.079} & \textbf{0.082} & \textbf{0.065} & \textbf{0.101} & \textbf{0.079} & \textbf{0.086} & \textbf{0.077} & \textbf{0.047} \\
      \bottomrule
    \end{tabular}
  }
  \vspace{-10pt}
\end{table*}

\subsubsection{Implementation Details}

SpaRV is implemented in PyTorch with custom CUDA kernels.
Every SpaRV result in this paper is produced on a single NVIDIA RTX~4090 with 24\,GB of device memory, an Intel Core i7-13700K CPU, and 64\,GB of host memory.
On the benchmark scenes of Secs.~\ref{ssec:replica}--\ref{ssec:egonerf}, training runs 20{,}000 iterations of sparse Adam~\cite{svraster}: voxels are subdivided every 1{,}000 iterations and pruned on each coverage epoch, and camera optimization alternates with map training on the same interval.
The large-scale scenes of Sec.~\ref{ssec:large_scale_recon} use the coarse-to-fine training of Sec.~\ref{ssec:hier_opt}.
Coarse training stretches the benchmark schedule by a factor that adapts to the dataset size.
Here the factor is 5.0, so coarse training runs 100,000 iterations and every milestone above scales with it.
Each submap is trained as an ordinary map-training run on the 20{,}000-iteration schedule.
The target submap size is $s_t = 100$\,m, and the granularity target is $\tau = 1$.
Deployment uses a target deploy-box size of $s_d = 10$\,m at the same granularity.

\subsection{Geometric Consistency on Replica}
\label{ssec:replica}

Replica~\cite{straub2019replica} is a synthetic indoor dataset with ground-truth camera poses, which remove camera error from the comparison.
We evaluate on 8 scenes at both interpolated and extrapolated viewpoints, the latter generated by uniformly sampling positions and orientations in each scene~\cite{liu2024neural}. In the extrapolation evaluation, pixels that are not seen in the training views are excluded.

\begin{figure}[h]
  \centering
  \renewcommand{\zoomtagfont}{\tiny}
  \setzoomcropa{0.191}{0.341}{0.485}{0.301}
  \setzoomcropb{1.379}{0.206}{0.379}{0.235}

  \imagetag{InstantNGP}{0.329\columnwidth}{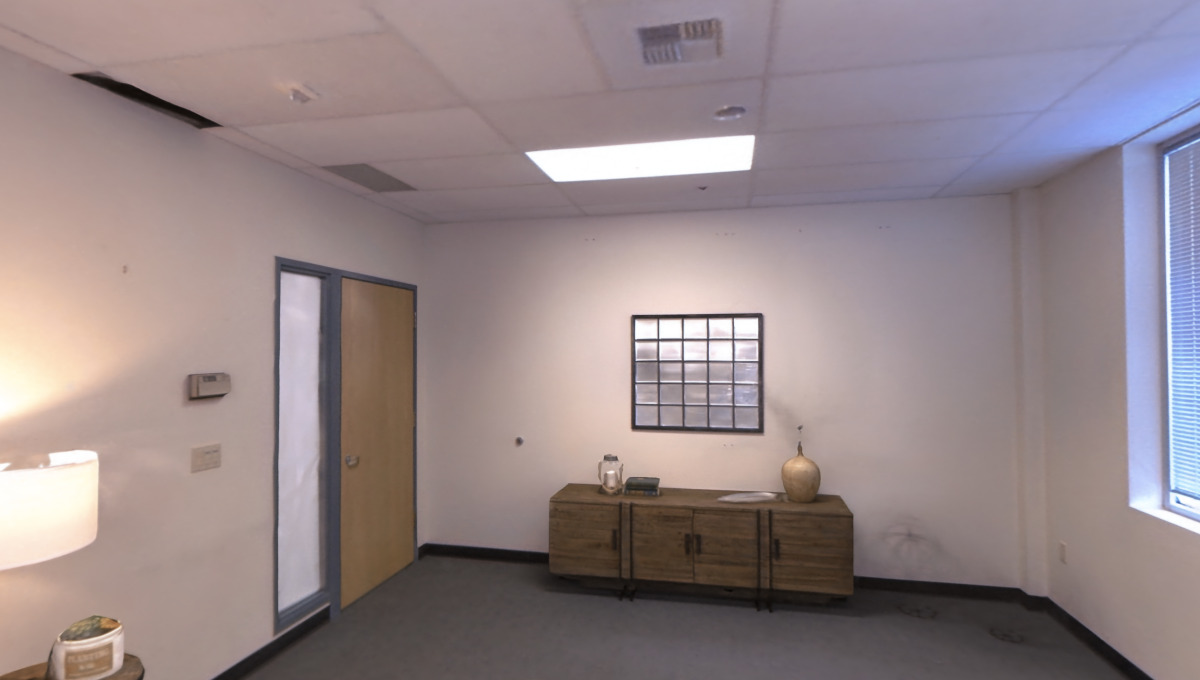}\hfill
  \imagetag{3DGS}{0.329\columnwidth}{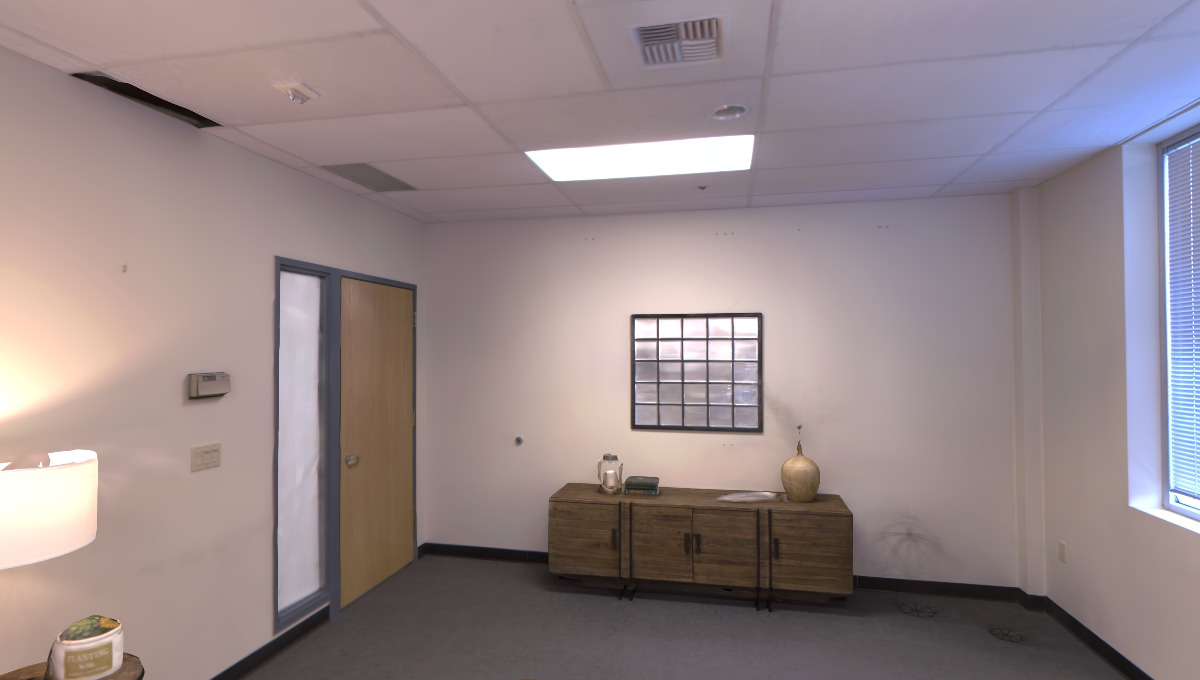}\hfill
  \imagetag{SVRaster}{0.329\columnwidth}{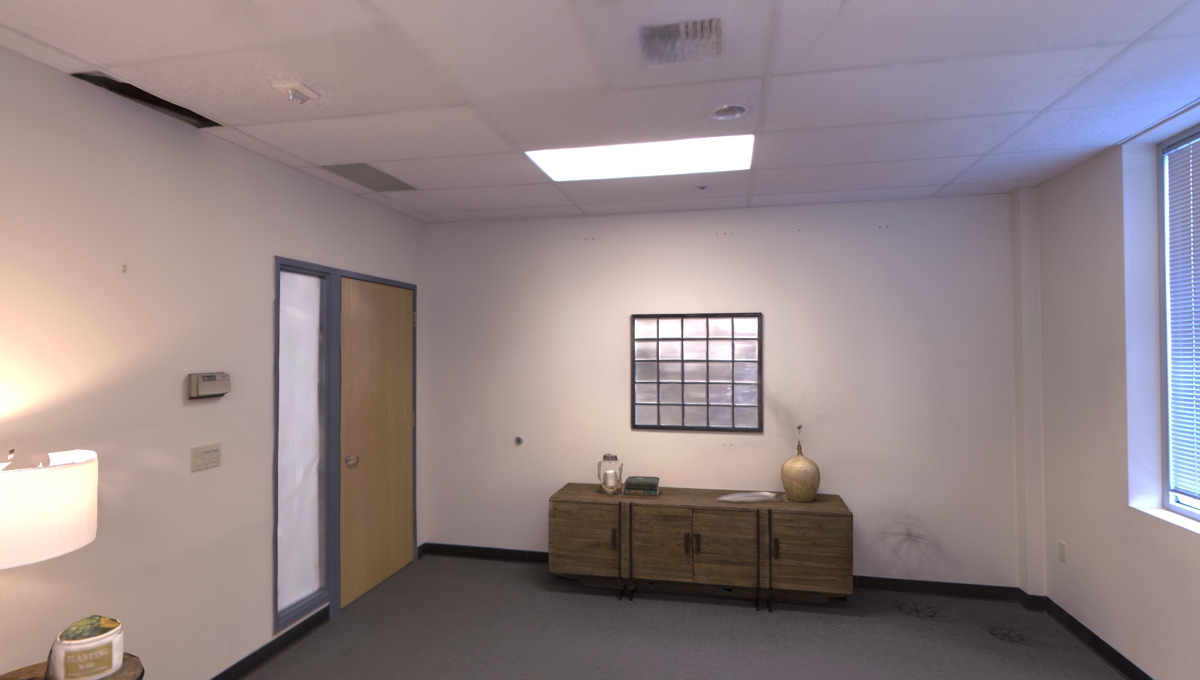}\par\vspace{2pt}
  \imagetag{LetsGo}{0.329\columnwidth}{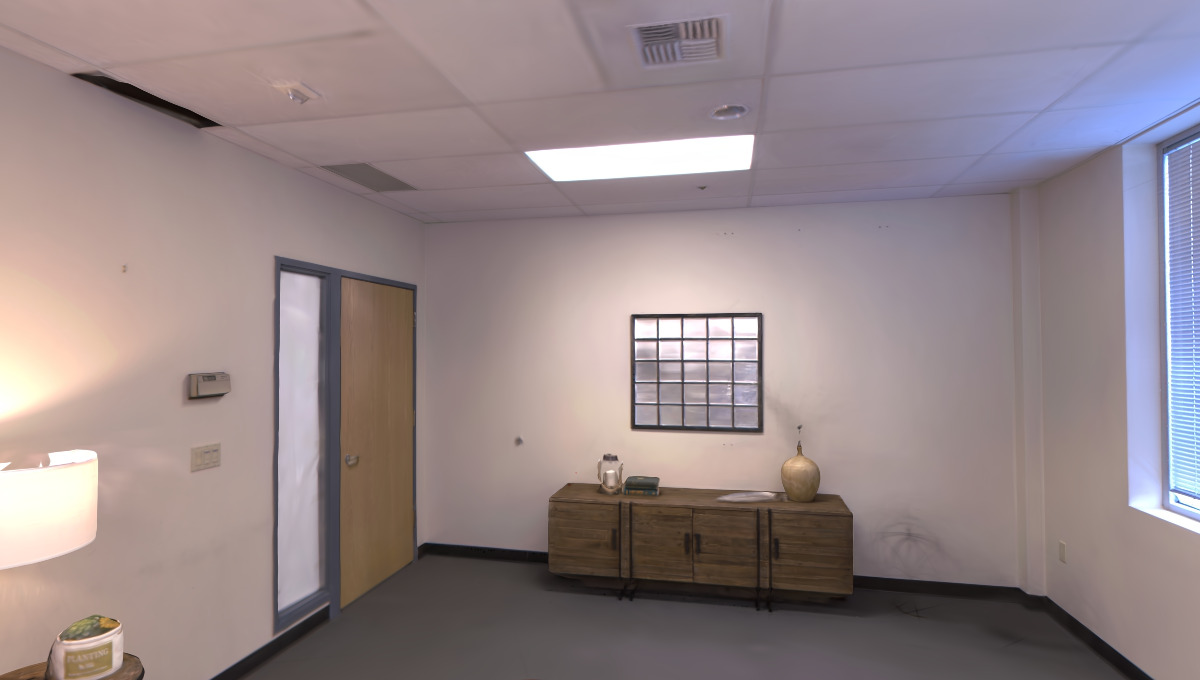}\hfill
  \imagetag{GS-SDF}{0.329\columnwidth}{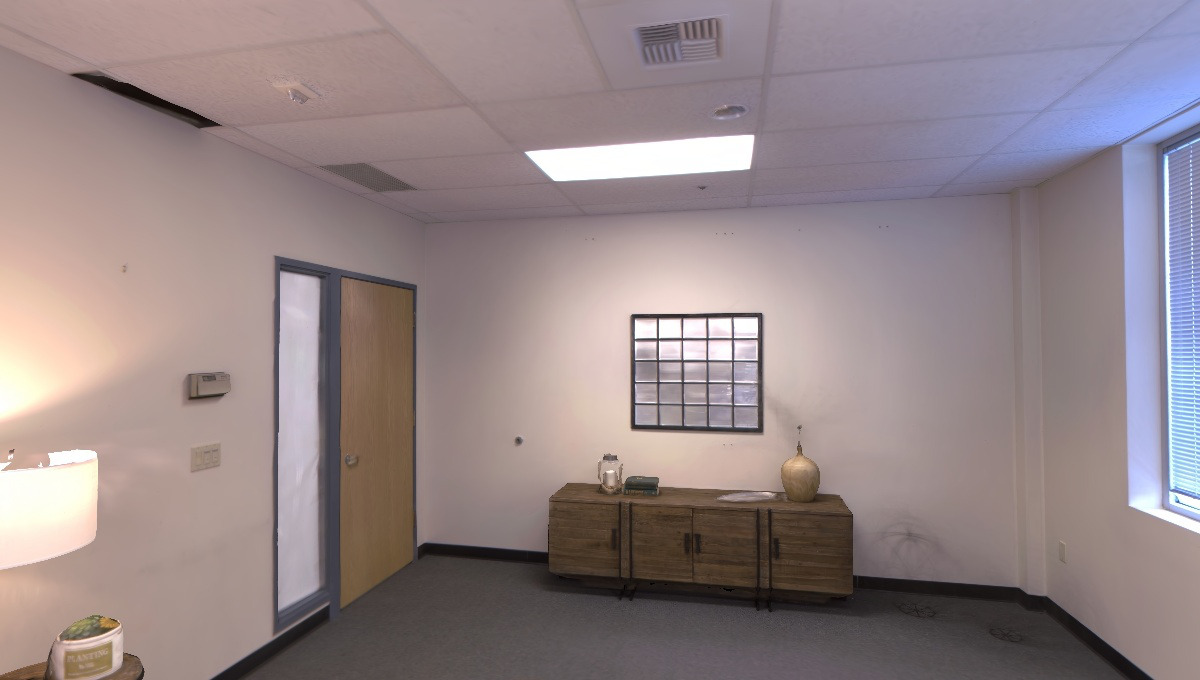}\hfill
  \imagetag{\textbf{SpaRV}}{0.329\columnwidth}{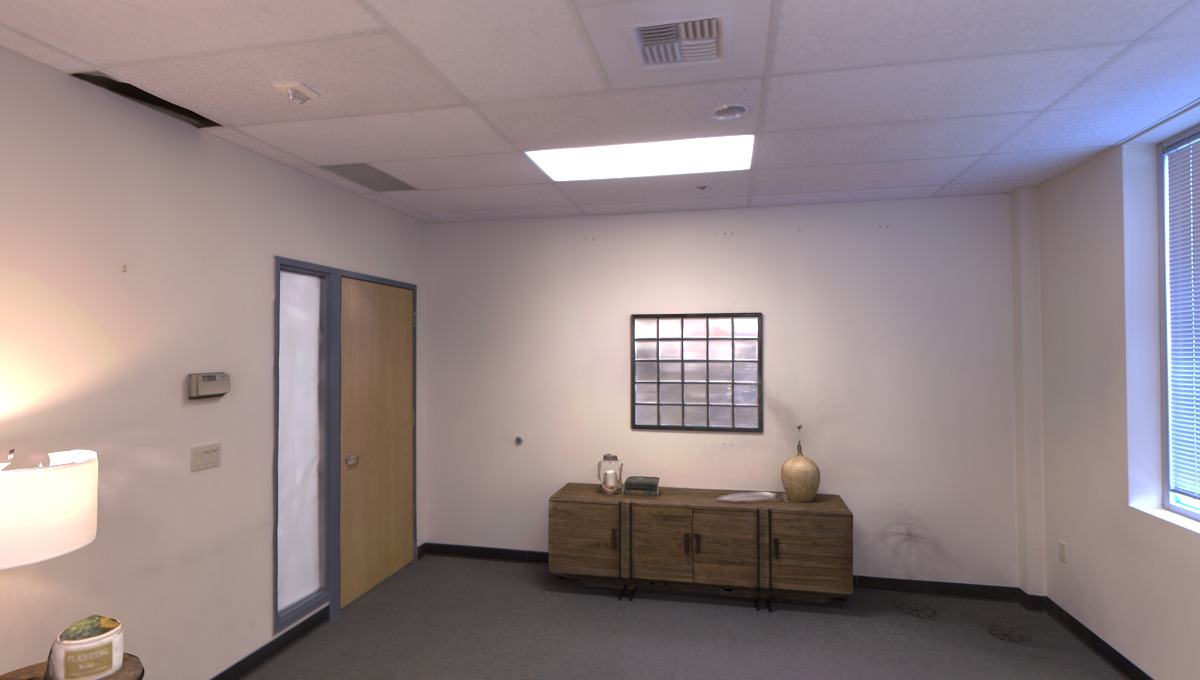}\par
  \vspace{3pt}{\color{zoomhair}\hrule height 0.6pt}\vspace{3pt}
  \cropboxcelltag{InstantNGP}{0.329\columnwidth}{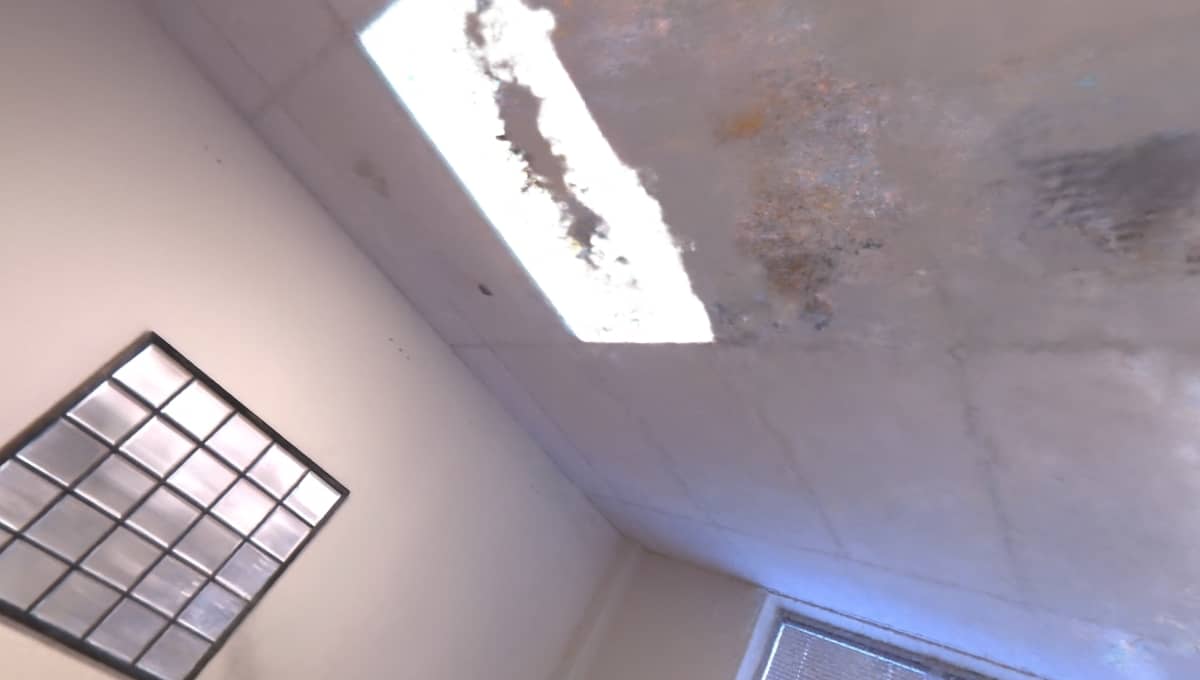}\hfill
  \cropboxcelltag{3DGS}{0.329\columnwidth}{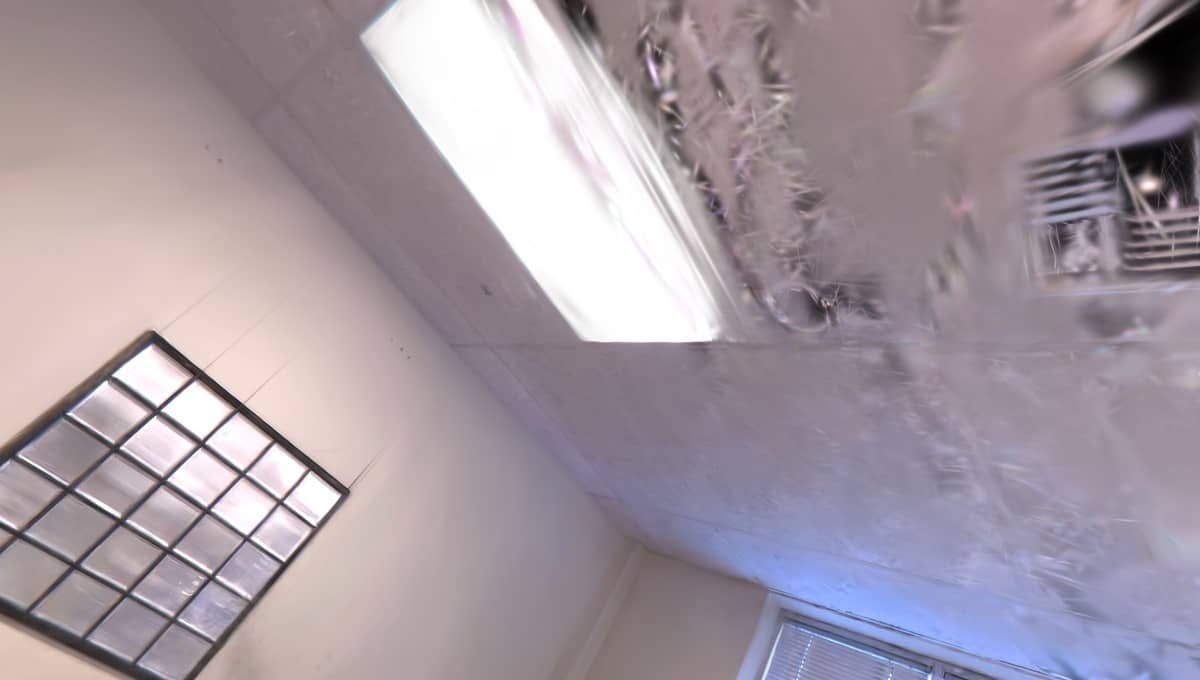}\hfill
  \cropboxcelltag{SVRaster}{0.329\columnwidth}{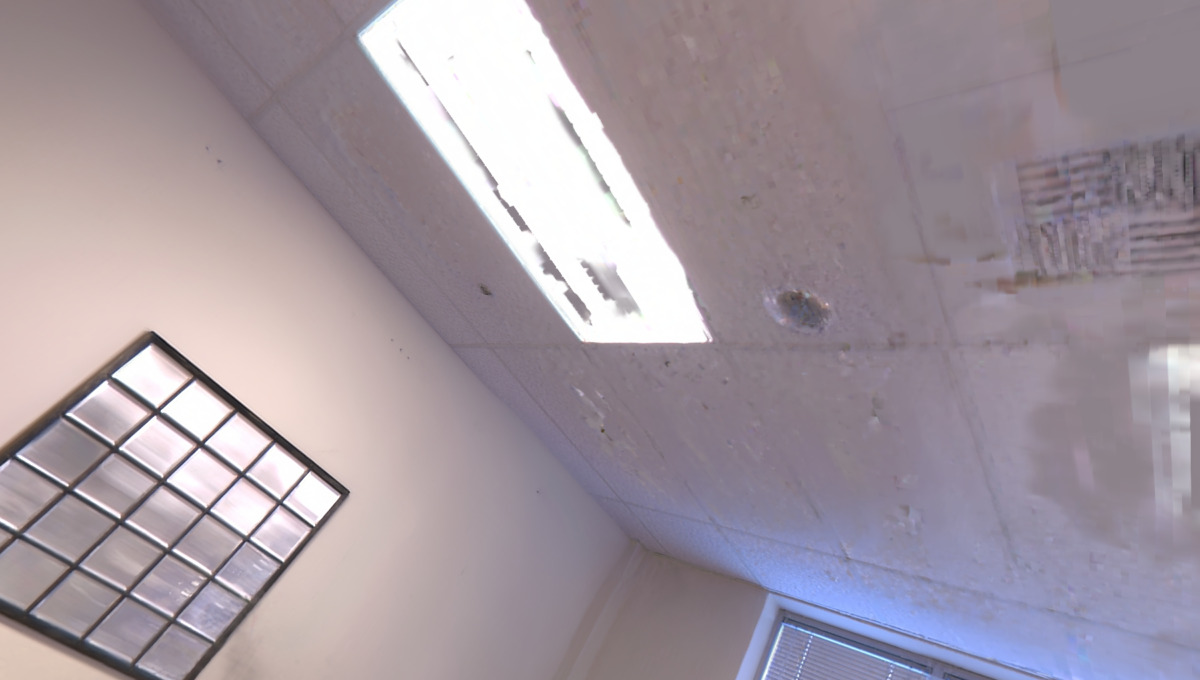}\par\vspace{2pt}
  \cropboxcelltag{LetsGo}{0.329\columnwidth}{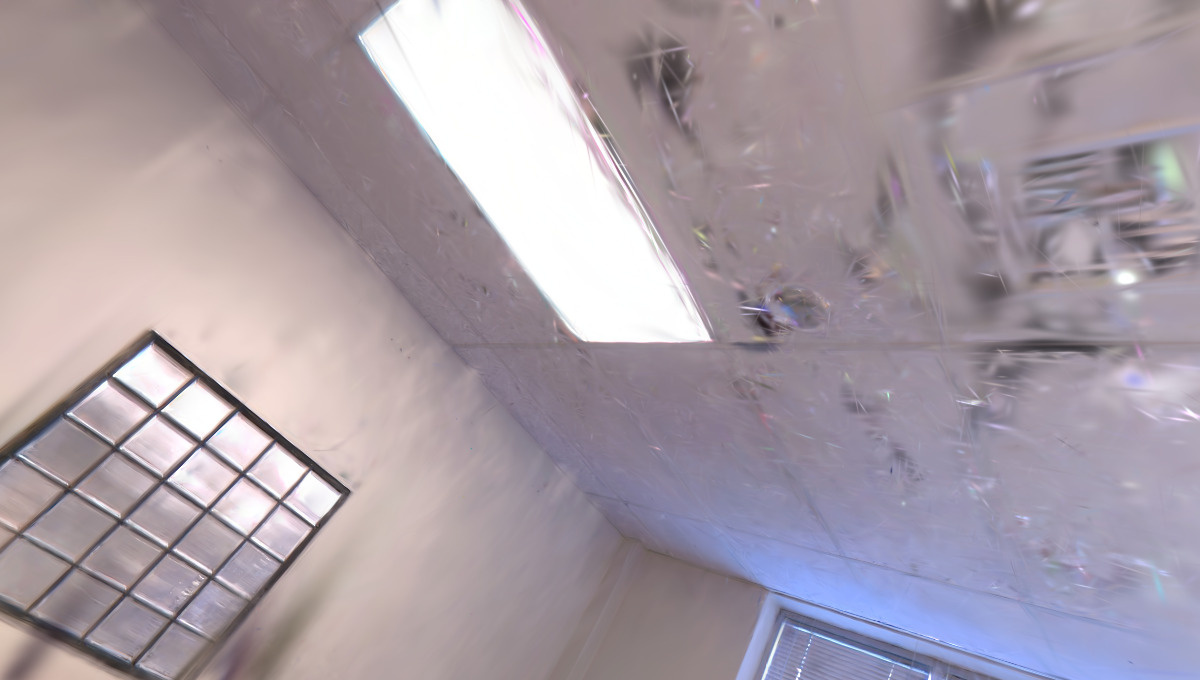}\hfill
  \cropboxcelltag{GS-SDF}{0.329\columnwidth}{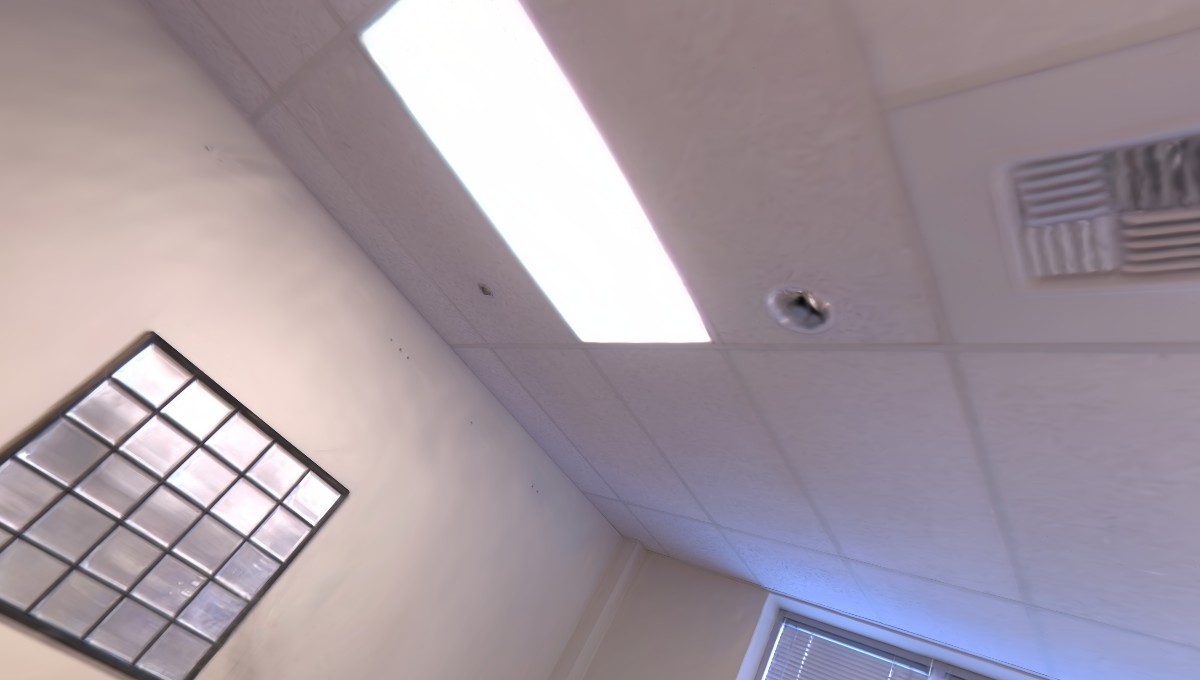}\hfill
  \cropboxcelltag{\textbf{SpaRV}}{0.329\columnwidth}{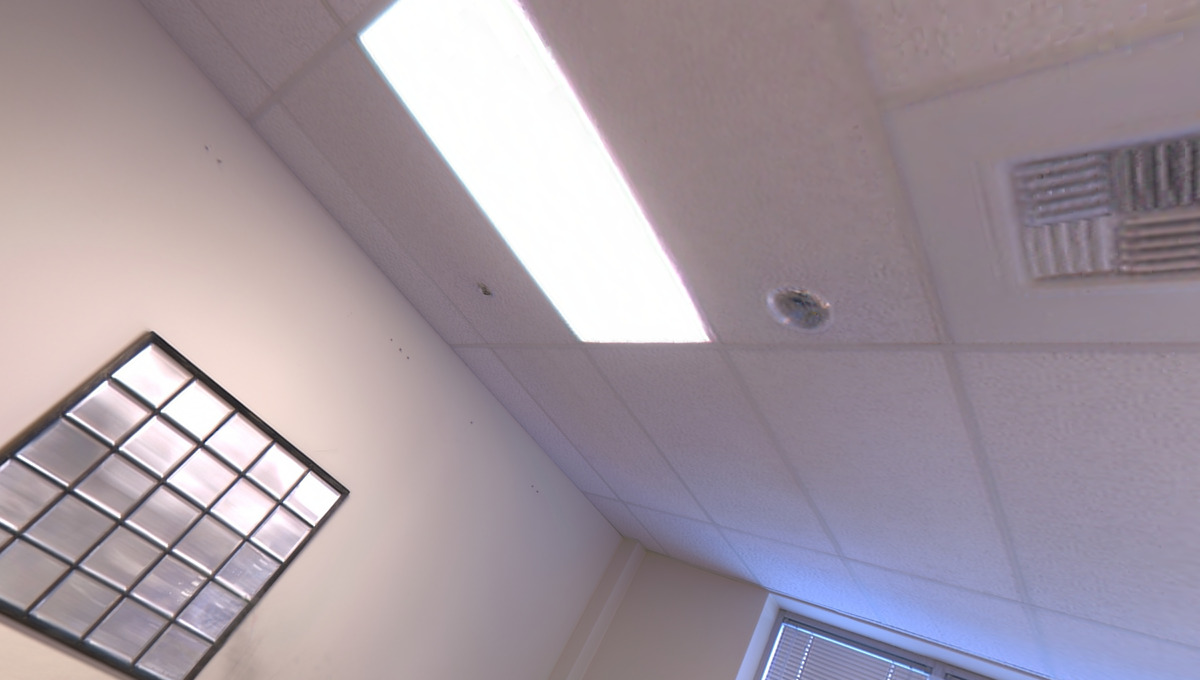}\par
  \vspace{-6pt}
  \caption{Rendering results on Replica Room-0. Top two rows: an interpolated
    training view, where every method is close to the reference. Bottom two
    rows: an extrapolated view of the same maps.
    }
  \label{fig:replica_qual_render}
  \vspace{-10pt}
\end{figure}

\begin{figure*}[!t]
  \centering
  \includegraphics[width=\linewidth]{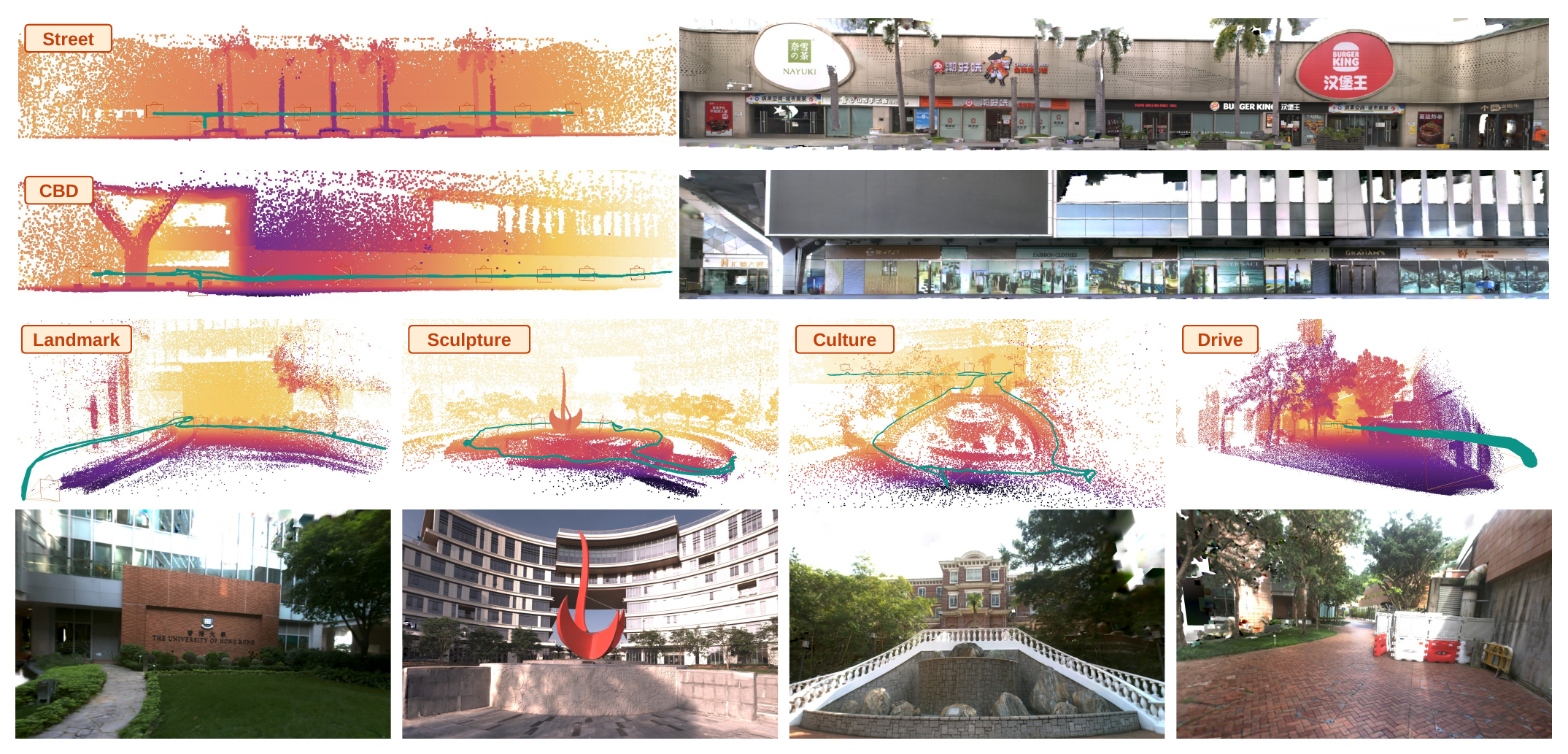}
  \vspace{-22pt}
  \caption{The six FAST-LIVO2 scenes: LiDAR map, trajectory, and a SpaRV render per scene.}
  \label{fig:fast_livo_overview}
  \vspace{-10pt}
\end{figure*}

Interpolation evaluates how well each method fits the observed views; extrapolation evaluates whether the map stays geometrically consistent at viewpoints away from the captured trajectory.
Table~\ref{tab:replica_rendering} reports the extrapolated results per scene and the averages of both, and Fig.~\ref{fig:replica_qual_render} shows one interpolated and one extrapolated view of each method on Replica Room-0.
At interpolated views the methods are not distinguishable: all of them fit the training images closely, and the six renderings in the top two rows are nearly identical.
The methods differ at the extrapolated view of the same maps. Geometric inconsistency appears first.
The captured trajectory covers the ceiling from few views and over a short baseline, so the image-only methods fit it to those views instead of recovering it as a surface.
The ceiling renders correctly from the training views, and degenerates into thin spikes and blurred blobs once the camera leaves the trajectory.
LetsGo, although its Gaussians are regularized by a LiDAR depth term, still smears the tile seams and leaves floaters.
GS-SDF and SpaRV keep the plane flat and the seams straight.
The extrapolated columns of Table~\ref{tab:replica_rendering} show the same result: SpaRV gains 3.3\,dB in extrapolated PSNR over SVRaster, whose sparse-voxel representation it shares, and leads every baseline on all three averaged metrics.

The two boxed regions of Fig.~\ref{fig:replica_qual_render} compare texture detail and shading, which can differ even where the geometry is correct.
In the ceiling exhaust vent, GS-SDF and SpaRV recover the individual slats that InstantNGP blurs into a single patch and that 3DGS, SVRaster and LetsGo break into thin spikes.
On the wall beside the ceiling light, GS-SDF places the surface correctly, yet the illumination falloff breaks into streaks.
Its overlapping Gaussians have no intrinsic compositing order, so the per-view sort leaves discontinuities at primitive boundaries~\cite{radl2024stopthepop}.
SpaRV renders the falloff smoothly, because its non-overlapping voxels keep the compositing order exact.
The LPIPS rows measure these appearance differences, and SpaRV leads at both interpolated and extrapolated viewpoints, ahead of GS-SDF.

\begin{figure*}[t]
  \centering
  \includegraphics[width=\linewidth]{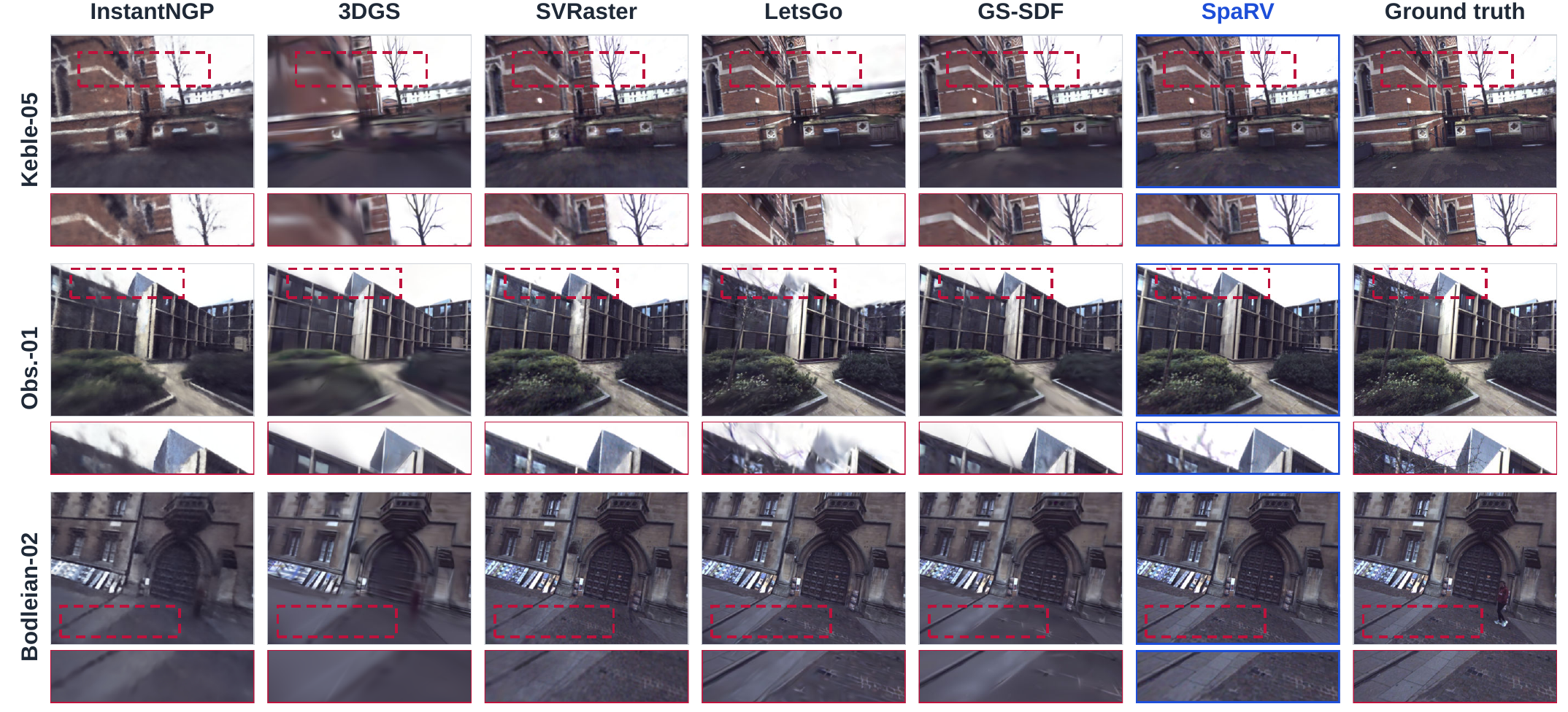}
  \vspace{-20pt}
  \caption{Rendering results on three Oxford Spires sequences: Keble-05, Obs.-01, and Bodleian-02.}
  \label{fig:oxford_rendering_qual_zoom}
  \vspace{-10pt}
\end{figure*}

\subsection{Real-World Mapping on FAST-LIVO2 Dataset}
\label{ssec:fast_livo2}

The FAST-LIVO2 dataset~\cite{zheng2024fast} is a real-world LiDAR-visual dataset captured with a solid-state Livox AVIA LiDAR. Its non-repetitive scanning pattern makes the range measurements sparse and irregular within any single frame. We evaluate on six outdoor scenes with three trajectory types (Fig.~\ref{fig:fast_livo_overview}): forward-facing (Landmark, CBD, Street), object-centric (Sculpture), and free-view (Culture, Drive).
Their initial poses and intrinsic parameters come from the FAST-LIVO2 odometry system~\cite{zheng2024fast}.

\begin{table}[t]
  \centering
  \caption{Training-view PSNR on FAST-LIVO2.}
  \label{tab:fast_livo2_nvs_psnr}
  \resizebox{\columnwidth}{!}{%
    \begin{tabular}{lccccccc}
    \toprule
    Method & Landmark & CBD & Culture & Drive & Sculpture & Street & Avg.\\
    \midrule
    \multicolumn{8}{c}{\textit{odometry cameras}}\\
    \cmidrule(lr){1-8}
    InstantNGP & {31.705} & 25.312 & 23.131 & 18.379 & \underline{25.276} & 23.429 & 24.539\\
    3DGS & \underline{32.260} & 22.033 & 21.571 & 25.570 & 24.294 & \underline{26.836} & 25.427\\
    SVRaster & 31.616 & \underline{26.419} & \underline{24.749} & {26.035} & {24.636} & 26.025 & \underline{26.580}\\
    \textbf{SpaRV} & 31.223 & {25.923} & {24.681} & \underline{26.871} & 24.483 & {26.130} & {26.552}\\
    \textbf{SpaRV} (cam.\ opt.) & \textbf{36.190} & \textbf{28.114} & \textbf{26.512} & \textbf{29.817} & \textbf{28.548} & \textbf{30.007} & \textbf{29.865}\\
    \addlinespace[2pt]
    \multicolumn{8}{c}{\textit{optimized cameras}}\\
    \cmidrule(lr){1-8}
    InstantNGP & 31.693 & 25.182 & 24.064 & 19.622 & 25.290 & 23.596 & 24.908\\
    3DGS & \underline{36.256} & 23.159 & 21.231 & 27.006 & 27.937 & \textbf{30.657} & 27.708\\
    SVRaster & 36.011 & \textbf{28.680} & 24.755 & 27.587 & \underline{28.411} & 29.443 & 29.148\\
    LetsGo & 32.066 & 27.107 & 24.218 & 17.712 & 23.177 & 28.408 & 25.448\\
    GS-SDF & 35.744 & \underline{28.584} & \textbf{27.143} & \underline{28.298} & 28.066 & \underline{30.493} & \underline{29.721}\\
    \textbf{SpaRV} & \textbf{36.403} & 28.157 & \underline{26.601} & \textbf{29.914} & \textbf{28.605} & 30.013 & \textbf{29.949}\\
    \bottomrule
    \end{tabular}%
  }
  \vspace{-5pt}
\end{table}

\subsubsection{Camera Optimization}
\label{sssec:camera_refinement}

Residual camera errors blur the map even at the training views, so we train on all images of each sequence and evaluate each method on those same images (Table~\ref{tab:fast_livo2_nvs_psnr}).
The table has two blocks, one per camera source.
In the \emph{odometry cameras} block, each method is trained on the FAST-LIVO2 odometry poses, and SpaRV is comparable to SVRaster and ahead of 3DGS and InstantNGP.
The last row of that block, SpaRV (cam.\ opt.), alternates map training with the camera optimization of Sec.~\ref{sec:pose_refinement} and outperforms every baseline.
The \emph{optimized cameras} block repeats the comparison with every method trained on the cameras that SpaRV (cam.\ opt.) optimized.

Camera optimization raises the PSNR of SpaRV on every scene, by 3.3\,dB on average, and retraining SpaRV on the optimized cameras reproduces the gain. The optimized cameras are not specific to SpaRV: every baseline trained under both settings in Table~\ref{tab:fast_livo2_nvs_psnr} improves when retrained on them.
Fig.~\ref{fig:fast_livo_cam_opt_qual} shows the textures on one training view of Drive and one of Sculpture, each rendered with 3DGS, SVRaster and SpaRV under both camera settings.
Both zooms fall on the paving.
The two scenes show two different effects of the odometry cameras.
On Drive the texture is incomplete: the brick pattern collapses to a flat blur under 3DGS and SVRaster. 
On Sculpture the pattern survives but is not sharp: the joint lines of the plaza slabs and the edge of the curved step are blurred for every method. 
The optimized cameras complete the pattern on Drive and sharpen it on Sculpture. 
Under the odometry cameras, SpaRV already yields the sharpest of the three renders; under the optimized cameras it is the closest to the captured frame.

\begin{figure}[t]
  \centering
  \setlength{\subfigtopskip}{0pt}
  \setlength{\subfigbottomskip}{6pt}
  \setlength{\subfigcapskip}{-1pt}

  \begin{minipage}[c]{0.955\columnwidth}
    \centering
    \subfigure[InstantNGP]{\includegraphics[width=0.327\linewidth]{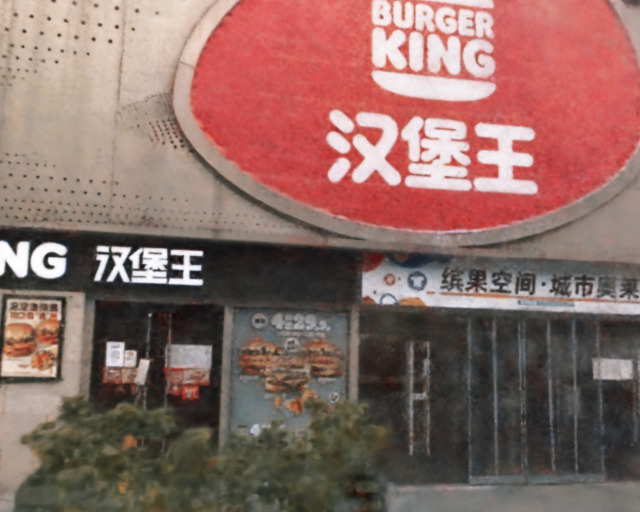}}\hspace{1pt}%
    \subfigure[3DGS]{\includegraphics[width=0.327\linewidth]{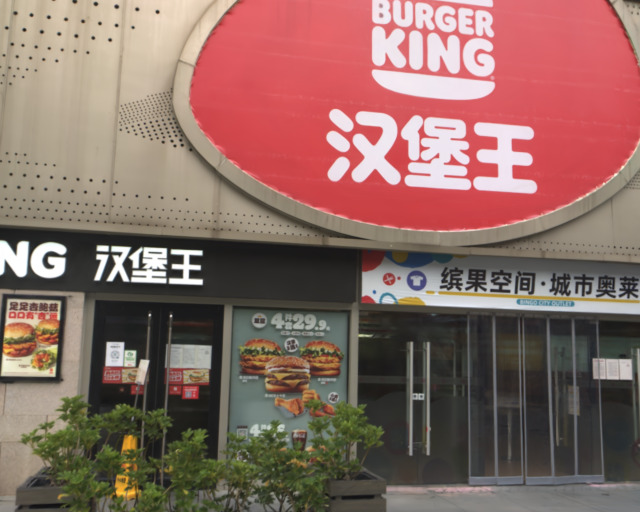}}\hspace{1pt}%
    \subfigure[SVRaster]{\includegraphics[width=0.327\linewidth]{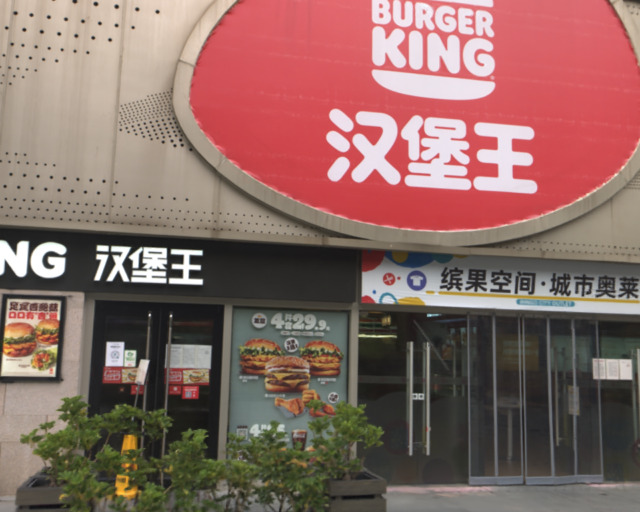}}\par
    \subfigure[LetsGo]{\includegraphics[width=0.327\linewidth]{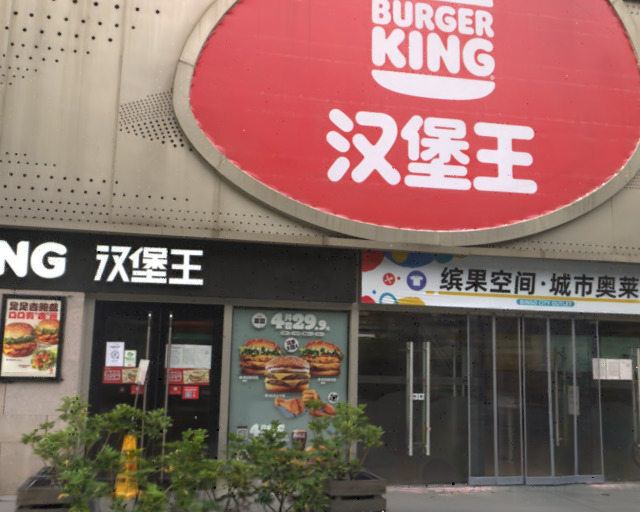}}\hspace{1pt}%
    \subfigure[GS-SDF]{\includegraphics[width=0.327\linewidth]{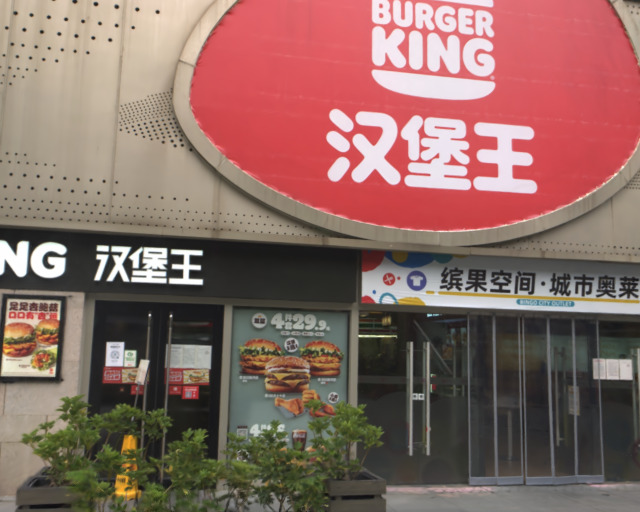}}\hspace{1pt}%
    \subfigure[SpaRV]{\includegraphics[width=0.327\linewidth]{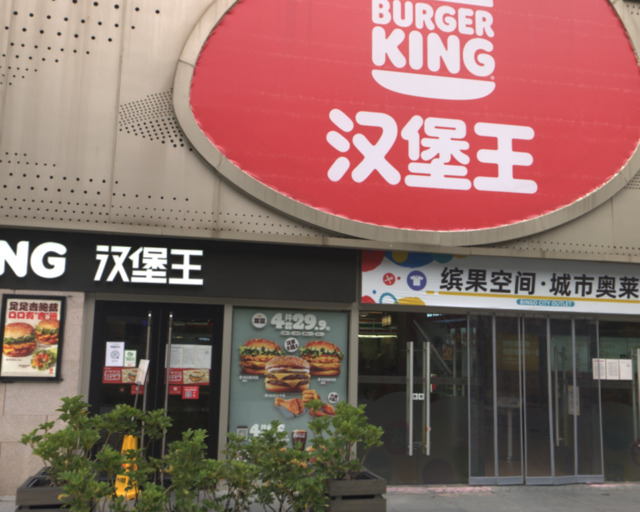}}
  \end{minipage}
  \vspace{-8pt}
  \caption{Rendering results on the FAST-LIVO2 Street scene at an interpolated view on the captured trajectory.}
  \vspace{-10pt}
  \label{fig:fast_livo_qual_interp}
\end{figure}

\begin{table}[h]
  \centering
  \caption{Quantitative results on the FAST-LIVO2 test split, under the optimized cameras.}
  \label{tab:fast_livo2_nvs_heldout}
  \resizebox{\columnwidth}{!}{%
    \begin{tabular}{ccccccccc}
      \toprule
      \textbf{Metrics} & \textbf{Methods} & \textbf{Landmark} &
\textbf{CBD} & \textbf{Culture} & \textbf{Drive} &
\textbf{Sculpture} & \textbf{Street} & \textbf{Avg.} \\
      \midrule
      \multirow{6}*{PSNR$\uparrow$} & InstantNGP & 30.983 & 25.080
& 23.339 & 24.111 & 25.080 & 23.506 & 25.350 \\
        & 3DGS & \textbf{34.567} & 22.659 & 21.112 & 27.234 & 27.496
& \textbf{30.123} & 27.198 \\
        & SVRaster & \underline{34.172} & 27.042 & 24.720 & 26.763 &
27.311 & 28.242 & 28.042 \\
        & LetsGo & 33.580 & \textbf{27.849} & \textbf{25.171} &
\textbf{29.386} & 24.588 & 29.798 & 28.395 \\
        & GS-SDF & 34.057 & \underline{27.157} & \underline{25.158}
& 28.555 & \underline{27.794} & \underline{30.093} &
\textbf{28.802} \\
        & \textbf{SpaRV} & 33.971 & 26.605 & 24.720 &
\underline{29.227} & \textbf{27.887} & 29.643 & \underline{28.676}
\\
      \midrule
      \multirow{6}*{SSIM$\uparrow$} & InstantNGP & 0.853 & 0.802 &
0.684 & 0.675 & 0.777 & 0.772 & 0.761 \\
        & 3DGS & \textbf{0.924} & 0.814 & 0.704 & 0.818 & 0.874 &
\underline{0.938} & 0.845 \\
        & SVRaster & 0.921 & 0.905 & 0.812 & 0.818 & 0.878 & 0.904 &
0.873 \\
        & LetsGo & 0.911 & 0.900 & 0.809 & \underline{0.918} & 0.823
& 0.929 & 0.881 \\
        & GS-SDF & 0.915 & \underline{0.905} & \underline{0.820} &
0.897 & \underline{0.879} & 0.936 & \underline{0.892} \\
        & \textbf{SpaRV} & \underline{0.923} & \textbf{0.914}
& \textbf{0.844} & \textbf{0.939} & \textbf{0.902} &
\textbf{0.947} & \textbf{0.911} \\
      \midrule
      \multirow{6}*{LPIPS$\downarrow$} & InstantNGP & 0.202 & 0.341
& 0.419 & 0.426 & 0.294 & 0.265 & 0.325 \\
        & 3DGS & 0.119 & 0.298 & 0.371 & 0.275 & 0.195 & 0.084 &
0.224 \\
        & SVRaster & \underline{0.107} & 0.143 & 0.210 & 0.255 &
\underline{0.169} & 0.127 & 0.168 \\
        & LetsGo & 0.137 & 0.166 & 0.229 & \underline{0.105} & 0.250
& 0.097 & 0.164 \\
        & GS-SDF & 0.138 & \underline{0.140} & \underline{0.203} &
0.146 & 0.190 & \underline{0.081} & \underline{0.149} \\
        & \textbf{SpaRV} & \textbf{0.100} & \textbf{0.099} &
\textbf{0.144} & \textbf{0.071} & \textbf{0.135} & \textbf{0.067} &
\textbf{0.103} \\
      \bottomrule
    \end{tabular}%
  }
  \vspace{-6pt}
\end{table}

\subsubsection{Geometrically Consistent Rendering}
\label{sssec:fast_livo2_nvs}

Table~\ref{tab:fast_livo2_nvs_heldout} evaluates rendering on a test split under the optimized cameras of Sec.~\ref{sssec:camera_refinement}.
The split excludes one image out of every eight from training and uses it only for evaluation.
SpaRV has the lowest LPIPS on all six scenes and the highest SSIM on five of them, ahead of GS-SDF, the closest baseline.
Figs.~\ref{fig:fast_livo_qual_interp} and~\ref{fig:fast_livo_qual_extrap} contrast interpolated with laterally extrapolated views on Street.
The methods are again indistinguishable at the interpolated view and separate once the camera moves off the captured trajectory.
At the extrapolated view, the image-only methods (InstantNGP, 3DGS, SVRaster) degrade, while the LiDAR-supervised methods (LetsGo, GS-SDF, SpaRV) keep the geometry and texture more stable.
The Gaussian-based LetsGo and GS-SDF show popping on the billboard, where SpaRV stays consistent.

\begin{table*}[t]
  \centering
  \caption{Oxford Spires dataset reconstruction benchmark results.}
  \label{tab:oxford_raw_camera_full_reconstruction_all_methods}
  \resizebox{\textwidth}{!}{%
    \begin{tabular}{ccccccccccccccc}
      \toprule
      \textbf{Metric} & \textbf{Method} & \textbf{Keble-02} & \textbf{Keble-03} & \textbf{Keble-04} & \textbf{Keble-05} & \textbf{Obs.-01} & \textbf{Obs.-02} & \textbf{Blenheim-01} & \textbf{Blenheim-02} & \textbf{Blenheim-05} & \textbf{Christ-02} & \textbf{Christ-03} & \textbf{Bodleian-02} & \textbf{Avg.} \\
      \midrule
      \multirow{7}{*}{PSNR$\uparrow$} & InstantNGP & 22.182 & 22.265 & 22.627 & 23.342 & 24.082 & 24.280 & 24.845 & 25.788 & 25.189 & 23.703 & 27.079 & 23.847 & 24.102 \\
       & 3DGS & \textbf{28.910} & \underline{29.146} & 24.652 & 25.082 & \underline{25.727} & \underline{26.027} & 24.631 & \textbf{25.992} & \underline{25.503} & 19.971 & 28.752 & 24.716 & 25.759 \\
       & SVRaster & 26.927 & 27.544 & 24.562 & 24.816 & 25.240 & 25.431 & 23.367 & 24.011 & 23.938 & 22.491 & \underline{28.819} & 26.286 & 25.286 \\
       & LetsGo & 27.801 & 27.819 & 24.612 & \underline{26.202} & 23.468 & 23.556 & \textbf{25.135} & 24.837 & 24.090 & \textbf{24.966} & \textbf{29.070} & \underline{27.009} & 25.714 \\
       & GS-SDF & 28.063 & 28.805 & \underline{25.270} & 25.728 & 25.667 & 25.765 & 24.624 & 25.541 & 25.054 & 23.834 & 26.676 & 25.536 & \underline{25.880} \\
       & \textbf{SpaRV} & \underline{28.452} & \textbf{29.159} & \textbf{26.171} & \textbf{26.705} & \textbf{26.937} & \textbf{27.217} & \underline{25.033} & \underline{25.850} & \textbf{25.936} & \underline{24.357} & 28.485 & \textbf{27.740} & \textbf{26.837} \\
      \cmidrule(lr){2-15}
       & \textbf{SpaRV} (occ off) & 28.320 & 29.031 & 26.041 & 26.577 & 26.762 & 27.014 & 24.580 & 25.204 & 25.444 & 24.231 & 28.392 & 27.682 & 26.607 \\
      \midrule
      \multirow{7}{*}{SSIM$\uparrow$} & InstantNGP & 0.750 & 0.753 & 0.722 & 0.737 & 0.756 & 0.759 & 0.782 & 0.783 & 0.779 & 0.795 & 0.894 & 0.752 & 0.772 \\
       & 3DGS & 0.865 & 0.865 & 0.767 & 0.776 & 0.795 & 0.801 & 0.754 & 0.773 & 0.768 & 0.720 & 0.919 & 0.746 & 0.796 \\
       & SVRaster & 0.863 & 0.866 & 0.787 & 0.800 & \underline{0.818} & \underline{0.823} & 0.765 & 0.781 & 0.772 & 0.779 & 0.916 & 0.800 & 0.814 \\
       & LetsGo & \underline{0.893} & \underline{0.892} & \underline{0.807} & \underline{0.838} & 0.805 & 0.811 & \underline{0.798} & \underline{0.793} & 0.784 & \textbf{0.843} & \textbf{0.928} & \underline{0.823} & \underline{0.835} \\
       & GS-SDF & 0.866 & 0.870 & 0.796 & 0.810 & 0.810 & 0.811 & 0.779 & 0.791 & \underline{0.785} & 0.805 & 0.905 & 0.780 & 0.817 \\
       & \textbf{SpaRV} & \textbf{0.899} & \textbf{0.901} & \textbf{0.832} & \textbf{0.846} & \textbf{0.864} & \textbf{0.869} & \textbf{0.810} & \textbf{0.825} & \textbf{0.817} & \underline{0.839} & \underline{0.927} & \textbf{0.849} & \textbf{0.857} \\
      \cmidrule(lr){2-15}
       & \textbf{SpaRV} (occ off) & 0.899 & 0.901 & 0.834 & 0.847 & 0.865 & 0.870 & 0.808 & 0.825 & 0.813 & 0.838 & 0.926 & 0.853 & 0.857 \\
      \midrule
      \multirow{7}{*}{LPIPS$\downarrow$} & InstantNGP & 0.412 & 0.412 & 0.433 & 0.419 & 0.422 & 0.421 & 0.460 & 0.457 & 0.459 & 0.444 & 0.325 & 0.475 & 0.428 \\
       & 3DGS & 0.267 & 0.273 & 0.368 & 0.365 & 0.337 & 0.331 & 0.448 & 0.422 & 0.429 & 0.490 & 0.227 & 0.432 & 0.366 \\
       & SVRaster & 0.228 & 0.226 & \underline{0.300} & 0.294 & \underline{0.268} & \underline{0.264} & \underline{0.385} & \underline{0.362} & \underline{0.370} & 0.377 & 0.216 & \underline{0.328} & \underline{0.302} \\
       & LetsGo & \underline{0.203} & \underline{0.205} & 0.308 & \underline{0.278} & 0.313 & 0.305 & 0.389 & 0.386 & 0.411 & \underline{0.324} & \underline{0.196} & \underline{0.328} & 0.304 \\
       & GS-SDF & 0.244 & 0.246 & 0.318 & 0.309 & 0.308 & 0.309 & 0.408 & 0.387 & 0.403 & 0.388 & 0.248 & 0.384 & 0.329 \\
       & \textbf{SpaRV} & \textbf{0.160} & \textbf{0.160} & \textbf{0.232} & \textbf{0.225} & \textbf{0.196} & \textbf{0.192} & \textbf{0.321} & \textbf{0.294} & \textbf{0.304} & \textbf{0.274} & \textbf{0.164} & \textbf{0.256} & \textbf{0.232} \\
      \cmidrule(lr){2-15}
       & \textbf{SpaRV} (occ off) & 0.159 & 0.159 & 0.228 & 0.221 & 0.193 & 0.189 & 0.317 & 0.288 & 0.305 & 0.274 & 0.167 & 0.245 & 0.229 \\
      \bottomrule
    \end{tabular}
  }
  \vspace{-10pt}
\end{table*}

\subsection{Benchmarking with Accurate Trajectories on Oxford Spires}
\label{ssec:oxford}

Oxford Spires~\cite{tao2024oxford} is an outdoor LiDAR-visual benchmark of 12 sequences around Oxford landmarks, captured with a 64-channel spinning Hesai QT64 LiDAR and three synchronized global-shutter cameras. Its ground-truth trajectories are registered against a millimeter-accurate Leica RTC360 laser scan, so they are much more accurate than robotic odometry.
Table~\ref{tab:oxford_raw_camera_full_reconstruction_all_methods} summarizes the benchmark results.
SpaRV leads on all three averaged metrics. Its LPIPS is the lowest on all 12 scenes and its SSIM on 10 of them, while PSNR separates the methods less.
Fig.~\ref{fig:oxford_rendering_qual_zoom} shows Keble-05, Obs.-01, and Bodleian-02.
On Keble-05, the baselines blur the bricks and the tree trunks are missing. On Obs.-01, they blur the roof and the building corner is missing. On Bodleian-02, the paving joints become a grey patch. SpaRV reconstructs these details.

The \textit{SpaRV (occ off)} row of Table~\ref{tab:oxford_raw_camera_full_reconstruction_all_methods} turns off the occupied voxel guidance $\mathcal{L}_{\text{occ}}$ (Sec.~\ref{ssec:point_occ_guidance}) and keeps every other loss and camera setting fixed. PSNR then drops on all 12 scenes, SSIM remains nearly unchanged, and LPIPS is slightly better without the term.
Fig.~\ref{fig:occ_guidance_ablation} compares the same view with and without the term. The color images are nearly indistinguishable, which is consistent with SSIM and LPIPS, but the depth and normal maps without the term reveal a band of floaters over the lawn.

\subsection{Large-Scale Reconstruction on Garden and Campus}
\label{ssec:large_scale_recon}

We captured Garden and Campus (Fig.~\ref{fig:manifold_overview}) with a handheld Manifold Pocket 2 scanner\footnote{\url{https://www.3dmanifold.com/products/pocket2}} carrying a solid-state Livox MID360 LiDAR and three fisheye cameras. Their poses are produced by the vendor's commercial software with loop closure and bundle adjustment.
Garden is an outdoor district densely covered with complex objects: 44,550 images and 146.7\,M LiDAR points along a 2,239\,m trajectory covering 45,199\,m$^2$.
Campus spans an entire university: 63,582 images and 151.6\,M LiDAR points along a 3,534\,m trajectory covering 234,035\,m$^2$.

\begin{figure}[t]
  \centering
  \includegraphics[width=\columnwidth]{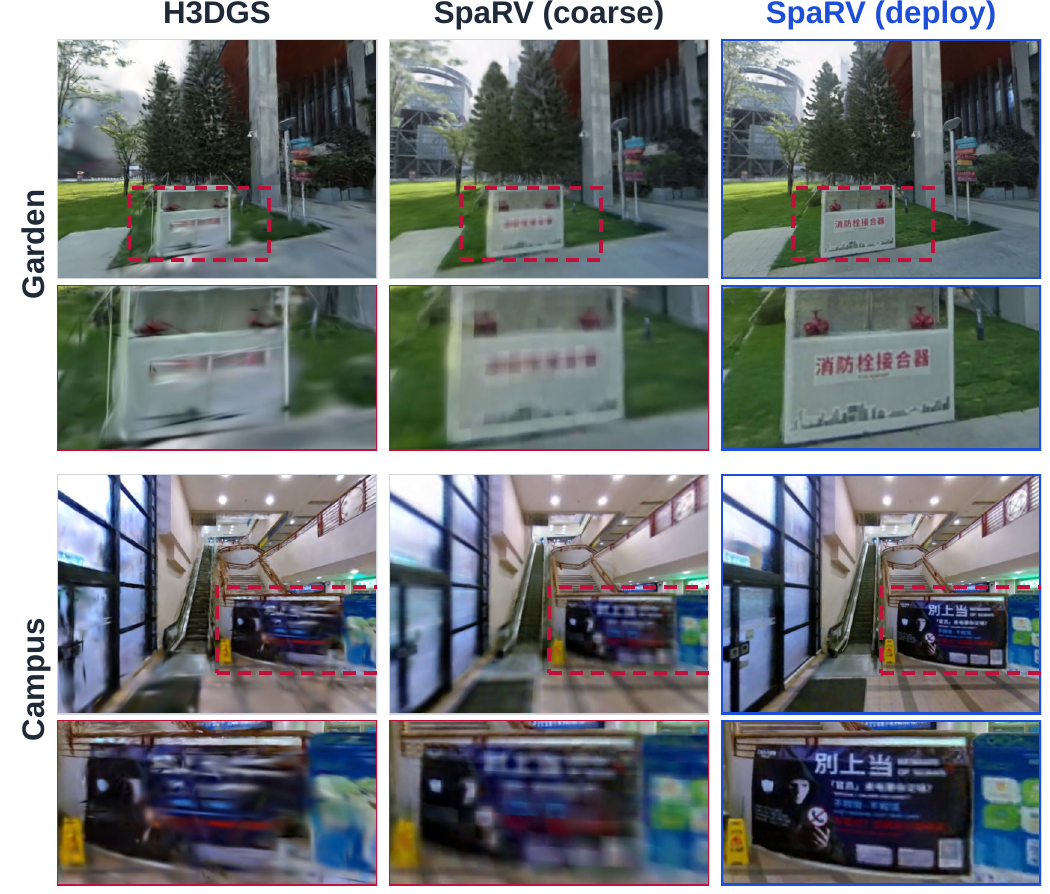}
  \vspace{-20pt}
  \caption{Rendering results on Garden (top) and Campus (bottom): H3DGS, SpaRV (coarse), and SpaRV (deploy).}
  \label{fig:university_submap}
  \vspace{-10pt}
\end{figure}

On each scene, we compare SpaRV at two stages with H3DGS.
SpaRV (coarse) is the output of coarse training, which spreads the budget $N_{\max}$ over the whole scene.
SpaRV (deploy) adds submap training (Sec.~\ref{ssec:submap_train}), which refines one submap at a time as far as the budget allows, and renders through a cut around the camera (Sec.~\ref{ssec:deploy_render}).
H3DGS~\cite{kerbl2024hierarchical} is the large-scale baseline.
Table~\ref{tab:large_scale_recon} reports the rendering quality of H3DGS, SpaRV (coarse), and SpaRV (deploy) on the training views of both scenes.
Submap training improves SpaRV (coarse) on every metric, most on Campus ($+2.5$\,dB PSNR), whose larger extent spreads the coarse budget more thinly.
SpaRV (deploy) has the highest PSNR and SSIM on both scenes.
Fig.~\ref{fig:university_submap} shows one matched viewpoint per scene.
SpaRV (coarse) already recovers the layout of both scenes, but fine content is blocky: the lettering on the Garden hydrant cabinet and on the Campus poster is unreadable.
SpaRV (deploy) makes both legible and restores smaller details, such as the two red valves on the cabinet and the wet-floor sign in front of the poster.
H3DGS smears the same regions and leaves streaks on the Garden lawn and the Campus glass wall.

\begin{table}[t]
  \centering
  \caption{Rendering quality on the two self-collected scenes, evaluated on the training views.}
  \label{tab:large_scale_recon}
  \setlength{\tabcolsep}{4pt}
  \renewcommand{\arraystretch}{1.05}
  \footnotesize
  \begin{tabular}{llccc}
  \toprule
  Scene & Method & PSNR$\uparrow$ & SSIM$\uparrow$ & LPIPS$\downarrow$ \\
  \midrule
  \multirow{3}{*}{Garden}
     & H3DGS          & 18.02 & 0.620 & \textbf{0.282} \\
     & \textbf{SpaRV} (coarse) & \underline{20.89} & \underline{0.712} & 0.346 \\
     & \textbf{SpaRV} (deploy) & \textbf{21.12} & \textbf{0.713} & \underline{0.295} \\
  \midrule
  \multirow{3}{*}{Campus}
     & H3DGS          & 20.01 & \underline{0.660} & \underline{0.295} \\
     & \textbf{SpaRV} (coarse) & \underline{20.62} & 0.652 & 0.451 \\
     & \textbf{SpaRV} (deploy) & \textbf{23.16} & \textbf{0.760} & \textbf{0.283} \\
  \bottomrule
  \end{tabular}
  \vspace{-10pt}
\end{table}

\subsection{Runtime Analysis}
\label{ssec:runtime}

Table~\ref{tab:runtime_efficiency} reports the training and rendering cost of the two systems on Campus. Both systems are evaluated on the same $3960$-frame fly-through at $1280{\times}960$, each cutting at $\tau{=}1$, one pixel of projected size.
SpaRV needs $8.7$\,h in total to produce the deployment map.
The coarse map alone trains in $0.8$\,h and renders at $248.4$\,FPS in $4.3$\,GiB, so it already serves as a fast preview of the whole scene, although with less detail than the deployment map (Table~\ref{tab:large_scale_recon}).
To render the deployment map, SpaRV applies the cut around the camera under a $16$\,GiB GPU memory budget.
It renders at $68.8$\,FPS in $15.6$\,GiB, against H3DGS's $29.6$\,FPS in $22.5$\,GiB, and its rate includes the cost of the cut.
The cut is recomputed only when the camera leaves the deploy box, which happens on $12\%$ of frames and adds a bounded $63.6$\,ms to each of them; the remaining frames reuse the voxels already resident. 
Training peaks at $12.5$\,GiB and rendering at $15.6$\,GiB, so SpaRV reconstructs and renders the kilometer-scale Campus scene on a single $24$\,GB GPU.

\begin{table}[h]
  \centering
  \caption{Training and rendering cost on Campus, over one $3960$-frame fly-through at $1280{\times}960$. Prim.\ is the largest number of primitives rendered in a frame.}
  \label{tab:runtime_efficiency}
  \setlength{\tabcolsep}{3pt}
  \renewcommand{\arraystretch}{1.05}
  \footnotesize
  \begin{tabular}{lccccc}
  \toprule
  \multirow{2}{*}{Method} & \multirow{2}{*}{Train\,[h]$\downarrow$} & \multirow{2}{*}{FPS$\uparrow$} & \multirow{2}{*}{Prim.\,[M]} & \multicolumn{2}{c}{Peak VRAM\,[GiB]$\downarrow$} \\
  \cmidrule(lr){5-6}
  & & & & Train & Render \\
  \midrule
  H3DGS & \underline{7.3} & 29.6 & 15.5 & \textbf{8.3} & 22.5 \\
  \textbf{SpaRV} (coarse) & \textbf{0.8} & \textbf{248.4} & 9.9 & 12.5 & \textbf{4.3} \\
  \textbf{SpaRV} (deploy) & 8.7 & \underline{68.8} & 14.3 & \underline{11.7} & \underline{15.6} \\
  \bottomrule
  \end{tabular}
  \vspace{-10pt}
\end{table}

\subsection{Omnidirectional Cameras on the EgoNeRF Dataset}
\label{ssec:egonerf}

The EgoNeRF dataset~\cite{choi2023balanced} is an egocentric view-synthesis benchmark of 22 equirectangular sequences, 11 synthetic scenes in OmniBlender and 11 real captures in Ricoh360, each recorded along a short orbit around a fixed center.
The benchmark carries no range data, so SpaRV is trained from images alone in this setting.
The only change from the benchmark experiments of Secs.~\ref{ssec:replica}--\ref{ssec:oxford} is the ray generator: each pixel of an equirectangular image is mapped to a ray direction by its longitude and latitude, instead of through the pinhole intrinsics.
The map, the renderer, and the training schedule are unchanged.
Following the OmniGS~\cite{li2025omnigs} protocol, we reconstruct from the OpenMVG poses shipped with the benchmark, train EgoNeRF~\cite{choi2023balanced} on the same poses and split, and evaluate all methods on the same test frames.

\begin{table}[h]
\centering
\footnotesize
\setlength{\tabcolsep}{4pt}
\caption{Novel-view synthesis results on the test frames of the EgoNeRF dataset.}
\label{tab:egonerf_train_reconstruction}
\begin{tabular}{@{}lcccccc@{}}
\toprule
\multirow{2}{*}[-2pt]{\textbf{Method}} & \multicolumn{3}{c}{\textbf{OmniBlender (11)}} & \multicolumn{3}{c@{}}{\textbf{Ricoh360 (11)}} \\
\cmidrule(lr){2-4}\cmidrule(l){5-7}
 & PSNR$\uparrow$ & SSIM$\uparrow$ & LPIPS$\downarrow$ & PSNR$\uparrow$ & SSIM$\uparrow$ & LPIPS$\downarrow$ \\
\midrule
EgoNeRF~\cite{choi2023balanced} & 30.93 & 0.892 & 0.207 & 24.89 & 0.752 & 0.353 \\
OmniGS~\cite{li2025omnigs} & 33.09 & 0.918 & 0.174 & 25.99 & 0.825 & 0.262 \\
\textbf{SpaRV} & \textbf{33.43} & \textbf{0.926} & \textbf{0.129} & \textbf{26.14} & \textbf{0.838} & \textbf{0.233} \\
\bottomrule
\end{tabular}
\end{table}

\begin{figure}[h]
  \centering
  \makebox[1.1em]{}%
  \makebox[\dimexpr(\columnwidth-1.1em)/2\relax][c]{\small Farm}%
  \makebox[\dimexpr(\columnwidth-1.1em)/2\relax][c]{\small Bridge}\\[1.5pt]
  \parbox[b]{1.1em}{\offinterlineskip\centering
    \rotatebox{90}{\makebox[\dimexpr(\columnwidth-1.1em)*3791/10000\relax][c]{\small EgoNeRF}}\par
    \vspace{\dimexpr(\columnwidth-1.1em)*109/10000\relax}
    \rotatebox{90}{\makebox[\dimexpr(\columnwidth-1.1em)*3791/10000\relax][c]{\small OmniGS}}\par
    \vspace{\dimexpr(\columnwidth-1.1em)*109/10000\relax}
    \rotatebox{90}{\makebox[\dimexpr(\columnwidth-1.1em)*3791/10000\relax][c]{\small SpaRV}}}%
  \includegraphics[width=\dimexpr\columnwidth-1.1em\relax]{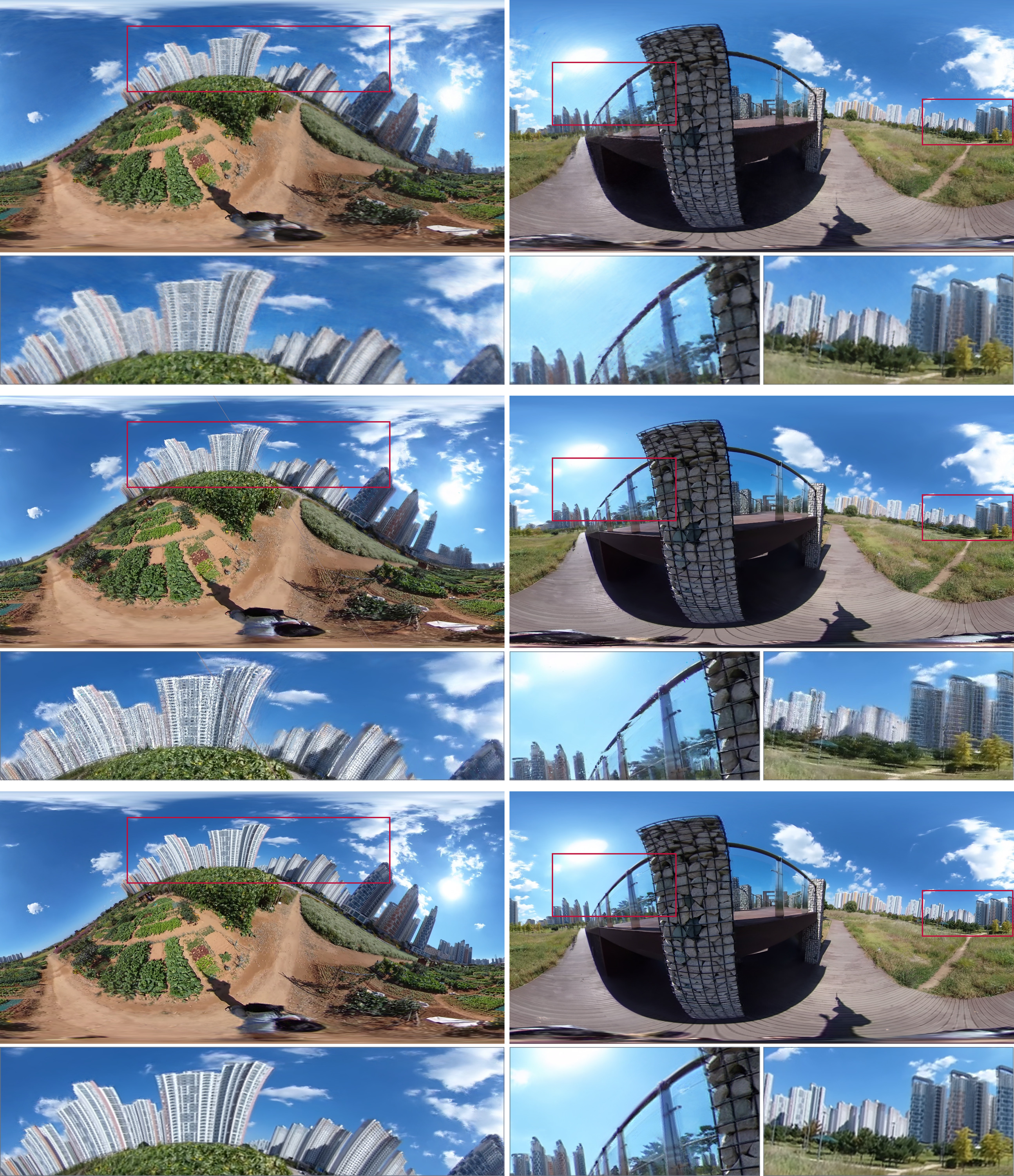}
  \vspace{-15pt}
  \caption{Rendering results on two Ricoh360 scenes of the EgoNeRF dataset.}
  \label{fig:egonerf_qual}
  \vspace{-5pt}
\end{figure}

SpaRV outperforms OmniGS on all three metrics on both splits (Table~\ref{tab:egonerf_train_reconstruction}). The PSNR margin is small, $0.34$\,dB on the synthetic OmniBlender scenes and $0.15$\,dB on the real Ricoh360 captures, while LPIPS drops by $26\%$ and $11\%$. Fig.~\ref{fig:egonerf_qual} compares the three methods on two Ricoh360 scenes.
OmniGS specializes its rasterizer for the equirectangular camera, whereas SpaRV uses the renderer and training schedule of the pinhole experiments unchanged.

\subsection{Applications}
\label{ssec:applications}

The trained map also supports the simulation of other camera and LiDAR models and surface extraction.
The field answers arbitrary ray queries, so it can simulate sensors it was never trained on (Sec.~\ref{sssec:alt_sensor_sim}).
Its field can be polygonized into an explicit surface (Sec.~\ref{sssec:surface_recon}).

\subsubsection{Sensor Simulation}
\label{sssec:alt_sensor_sim}

\begin{figure}[t]
  \centering
  \setlength{\subfigcapskip}{-3pt}
  \subfigure[Pinhole]{
    \centering
    \includegraphics[width=0.29\columnwidth]{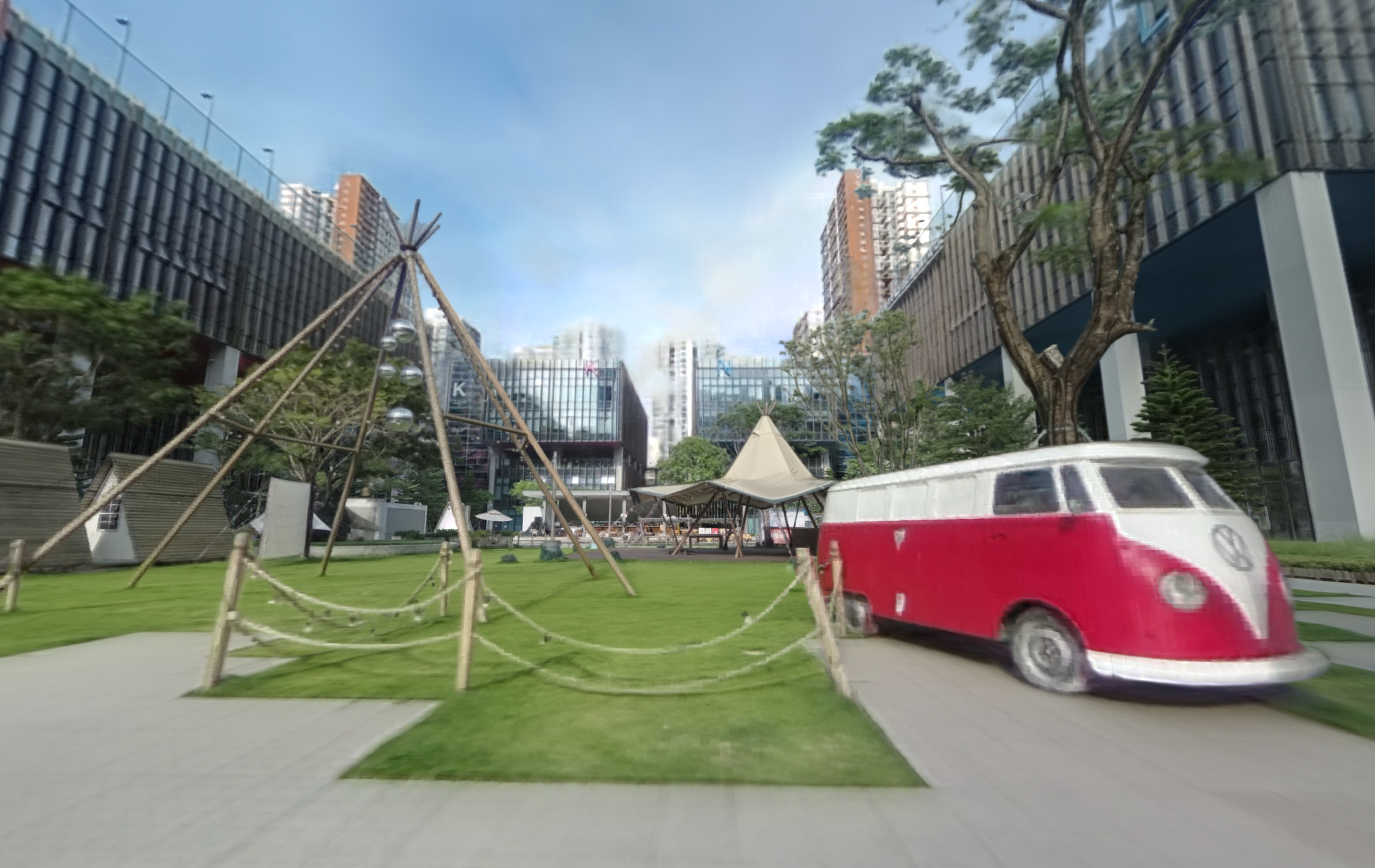}
  }
  \hspace{-4pt}
  \subfigure[Fisheye]{
    \centering
    \includegraphics[width=0.29\columnwidth]{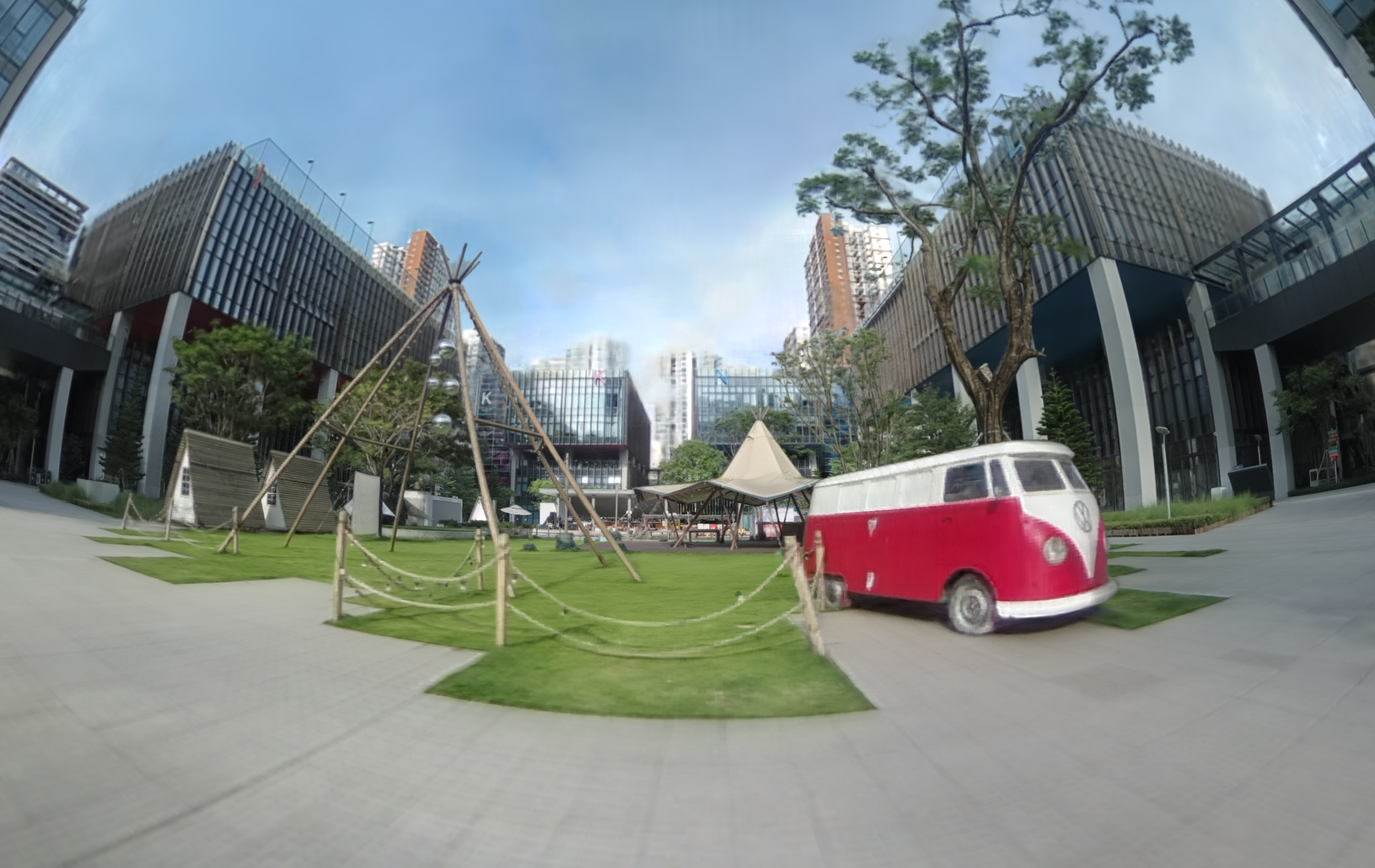}
  }
  \hspace{-4pt}
  \subfigure[Omni]{
    \centering
    \includegraphics[width=0.29\columnwidth]{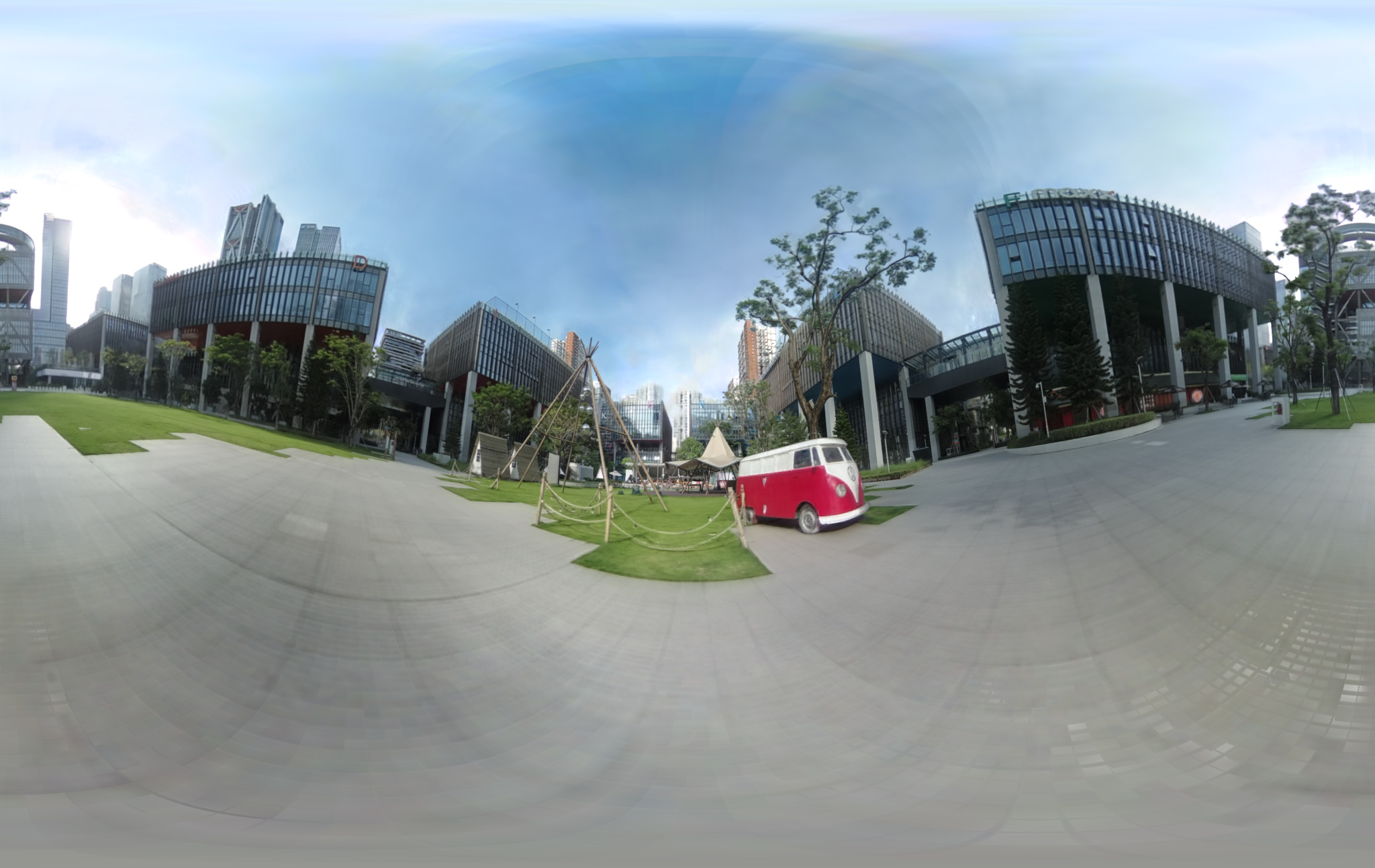}
  }

  \vspace{-6pt}

  \subfigure[AVIA pattern]{
    \centering
    \includegraphics[width=0.29\columnwidth]{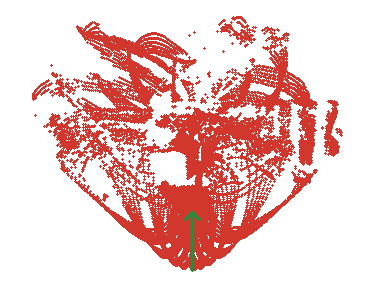}
  }
  \hspace{-4pt}
  \subfigure[MID360 pattern]{
    \centering
    \includegraphics[width=0.29\columnwidth]{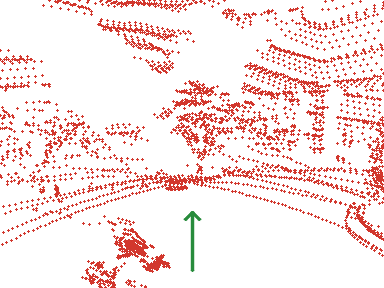}
  }
  \hspace{-4pt}
  \subfigure[OS128 pattern]{
    \centering
    \includegraphics[width=0.29\columnwidth]{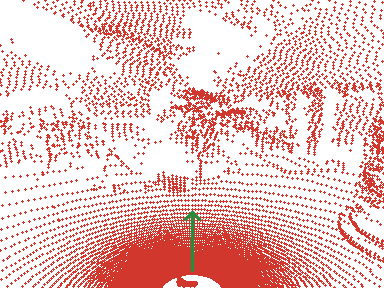}
  }
  \caption{Sensor simulation from one trained map: three camera models (top) and three LiDAR scan patterns (bottom).}
  \label{fig:alternative_sensor_sim}
  \vspace{-10pt}
\end{figure}

After training, SpaRV renders a different sensor by changing only the ray generator; the geometry and appearance stay fixed.
Fig.~\ref{fig:alternative_sensor_sim} shows three camera models (pinhole, fisheye, and omnidirectional) and three LiDAR scan patterns (AVIA, MID360, and OS128), all from one trained map.
The Livox-style AVIA and MID360 patterns use time-ordered scan tables from the official Livox scan-mode assets. OS128 is a wide-field ring-azimuth grid approximation~\cite{kong2023marsim}.
For each preset, SpaRV ray-marches those rays and converts the rendered median depth into per-frame point clouds.

\subsubsection{Surface Extraction}
\label{sssec:surface_recon}

\begin{figure}[h]
  \centering
  \setlength{\subfigtopskip}{0pt}
  \setlength{\subfigbottomskip}{4pt}
  \setlength{\subfigcapskip}{-2pt}
  \subfigure[Street]{\includegraphics[width=\columnwidth]{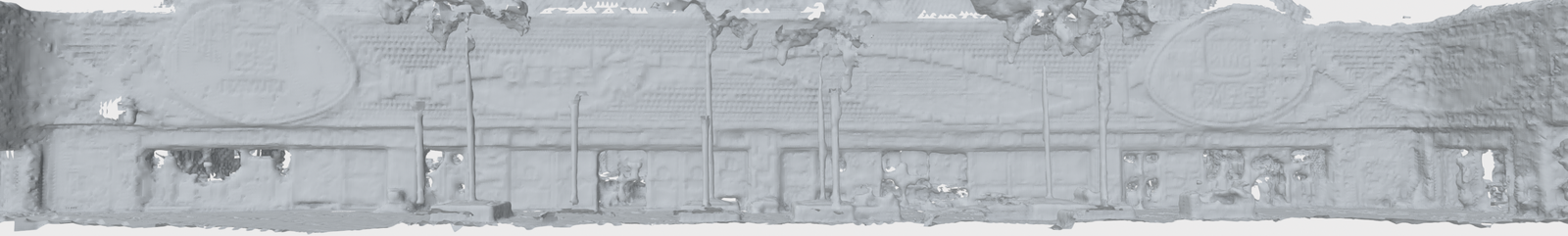}}\par
  \subfigure[CBD]{\includegraphics[width=\columnwidth]{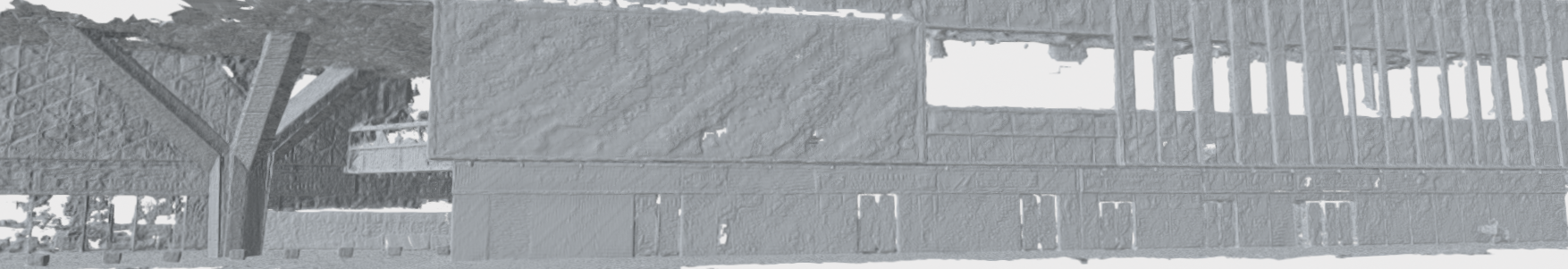}}\par
  \subfigure[Landmark]{\includegraphics[width=0.492\columnwidth]{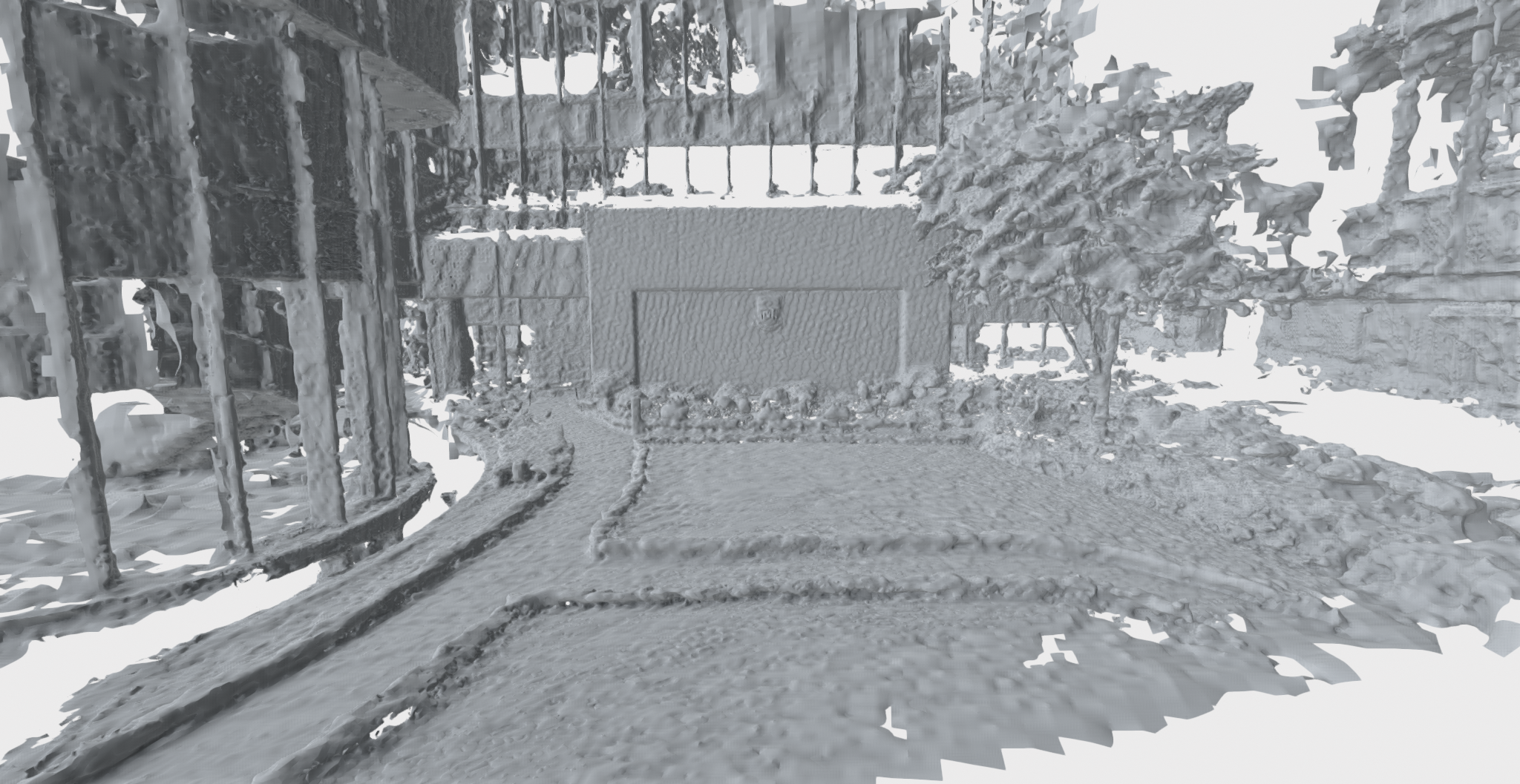}}\hspace{1pt}%
  \subfigure[Sculpture]{\includegraphics[width=0.492\columnwidth]{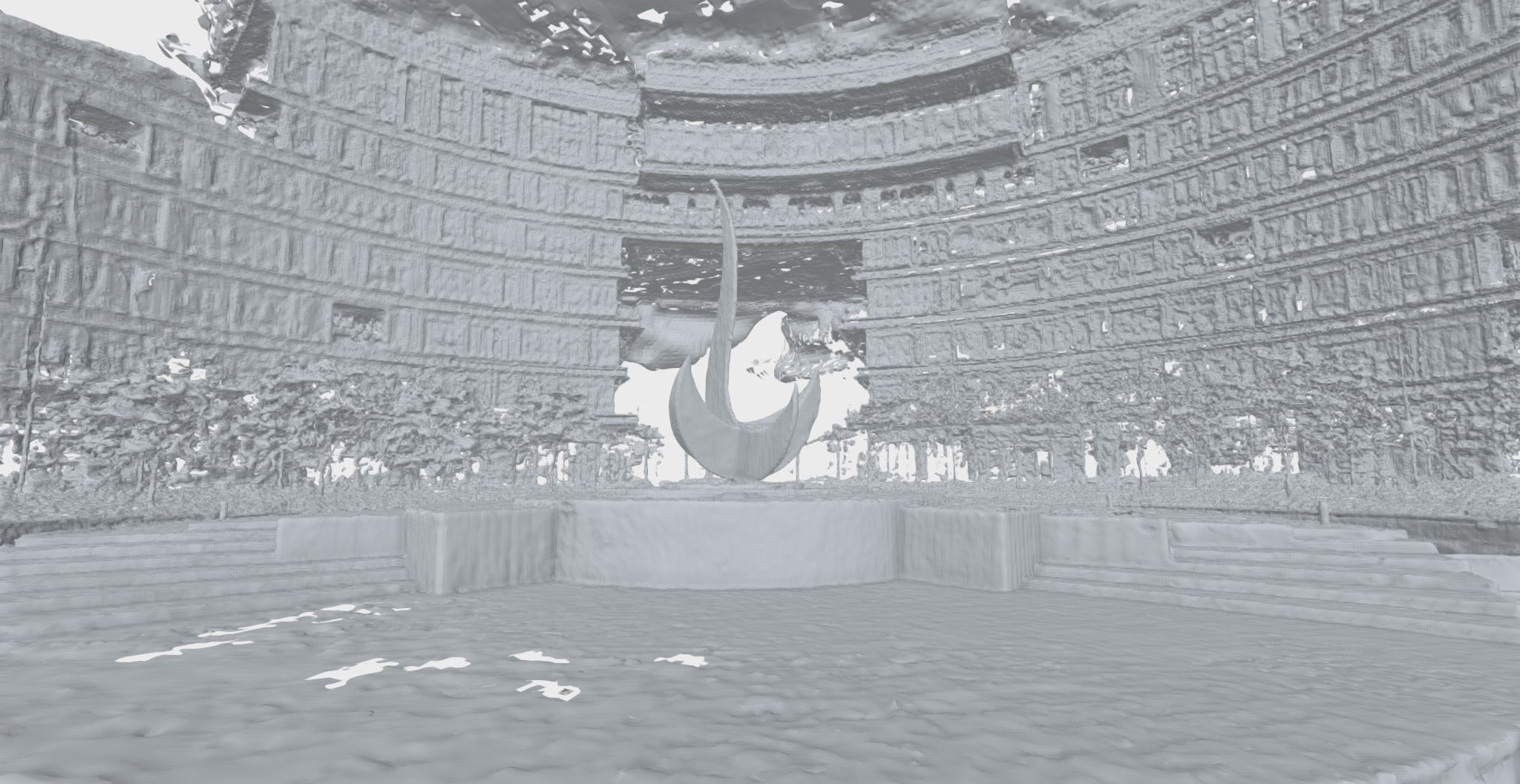}}\par
  \subfigure[Culture]{\includegraphics[width=0.492\columnwidth]{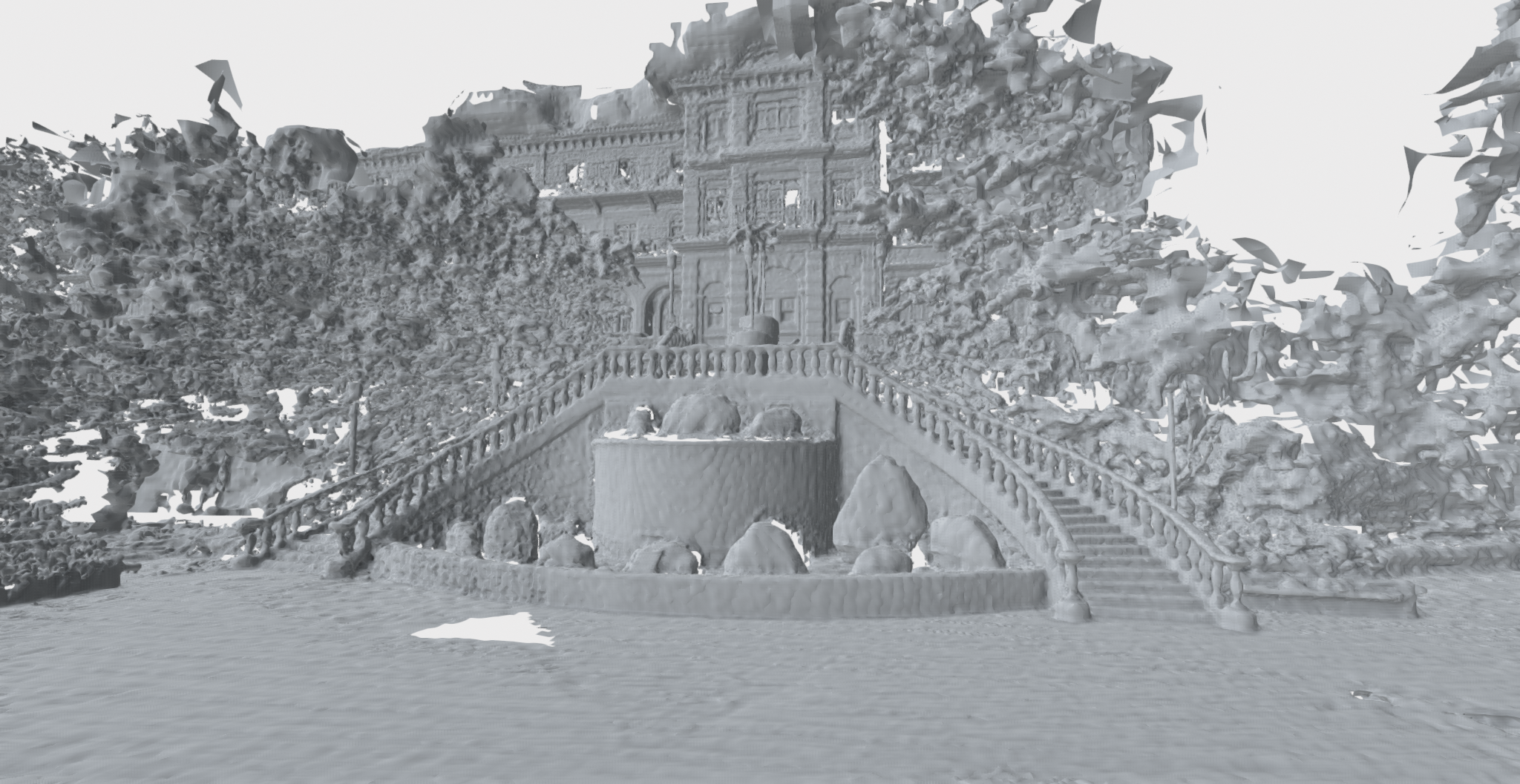}}\hspace{1pt}%
  \subfigure[Drive]{\includegraphics[width=0.492\columnwidth]{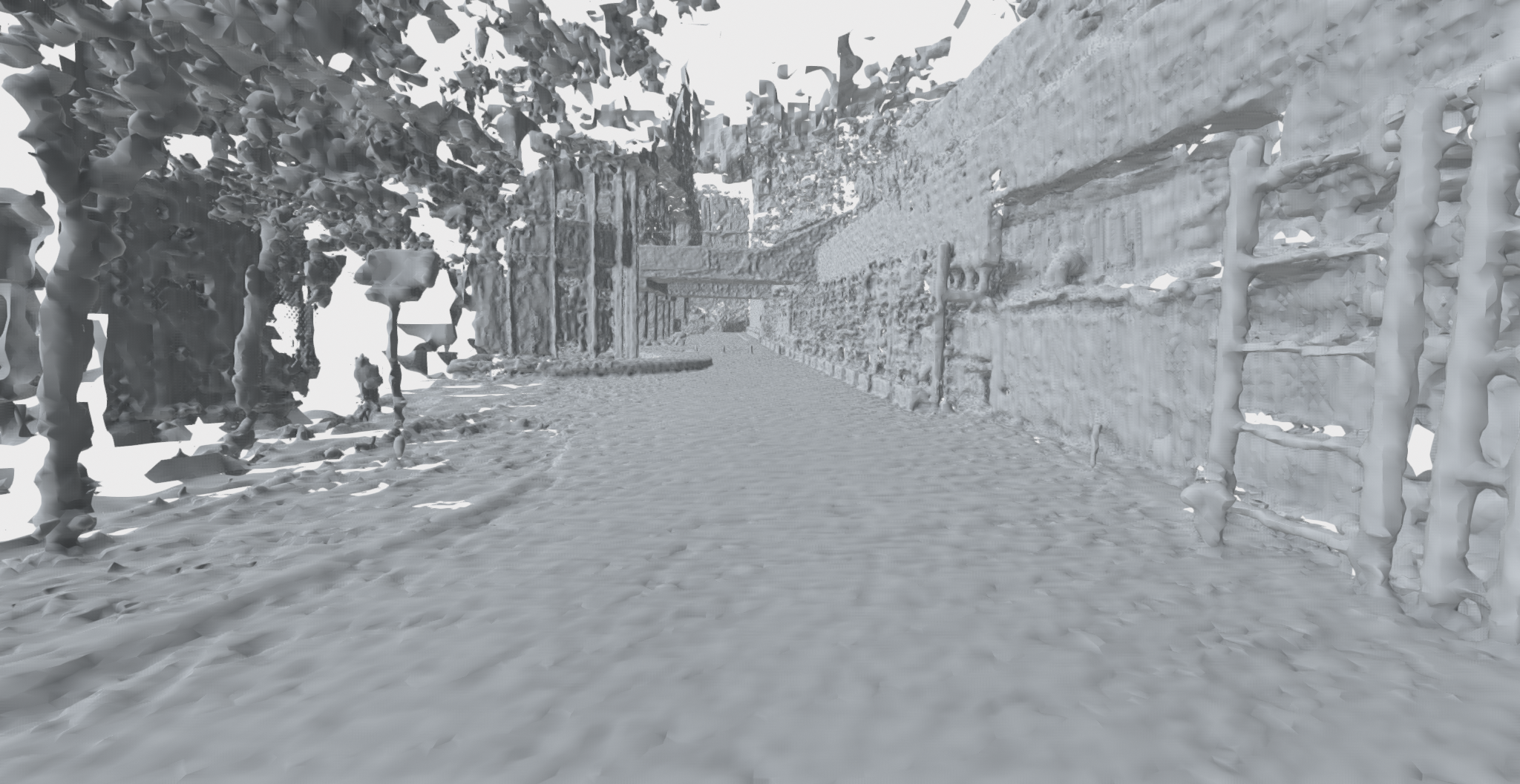}}
  \vspace{-10pt}
  \caption{Meshes extracted from the six FAST-LIVO2 maps of Sec.~\ref{ssec:fast_livo2} by marching cubes.}
  \label{fig:fast_livo2_mesh}
  \vspace{-10pt}
\end{figure}

Collision checking, physics, and simulator assets need an explicit mesh.
Instead of fusing rendered depth maps into a TSDF volume~\cite{huang20242d,liu2024towards}, which smooths away detail the field already encodes, SpaRV maps each trained grid-point density $\sigma$ to a per-vertex opacity through Eq.~\eqref{eq:voxel_alpha}, with the interval set to the local voxel size $\voxsiz$.
It then polygonizes that alpha field by marching cubes~\cite{lorensen1987marching}.
The mesh resolution thus follows the adaptive voxels.
Fig.~\ref{fig:fast_livo2_mesh} shows the meshes extracted from the six FAST-LIVO2 maps of Sec.~\ref{ssec:fast_livo2}.

\section{Conclusion and Future Work}
\label{sec:conclusion}

This paper presented SpaRV, a sparse radiance voxel map that fuses LiDAR and cameras for geometrically consistent mapping and rendering.
A differentiable ray marcher renders the sparse voxels along each ray, so LiDAR rays supervise the map alongside images.
LiDAR depth supervision and occupied voxel guidance ground the geometry, so the map stays consistent at viewpoints away from the captured trajectory.
Map training alternates with photometric optimization of the camera poses and focal lengths, which sharpens the textures of the map.
A granularity-driven cut of the octree bounds GPU memory in training and rendering, so SpaRV reconstructs and renders the kilometer-scale scene on a single 24\,GB GPU.
The trained maps also support surface extraction and the simulation of other camera and LiDAR models.

SpaRV stores geometry and appearance in the same voxels, but texture is often high-frequency where the geometry is not: a flat poster needs small voxels for its texture, not for its shape.
Keeping geometry in the sparse voxels and moving appearance to a more compact representation would let the map resolve finer detail within the same GPU memory.
SpaRV also assumes a static scene and carries no semantic labels.
We leave the decoupled representation, dynamic objects, and semantics to future work.

{
  \balance
  \bibliographystyle{IEEEtran}
  \bibliography{reference}
}

\end{document}